%% file: neurips_2026.tex
\documentclass{article}

\PassOptionsToPackage{numbers, compress}{natbib}
\usepackage{multirow}
\usepackage{url}            % simple URL typesetting
\usepackage{booktabs}       % professional-quality tables
\usepackage{amsfonts}       % blackboard math symbols
\usepackage{nicefrac}       % compact symbols for 1/2, etc.
\usepackage{microtype}      % microtypography
\usepackage{xcolor}         % colors
\usepackage{graphicx}
\usepackage{paralist}
\usepackage{wrapfig}
\usepackage{amssymb}
\usepackage{amsmath}
\usepackage{multirow}
\usepackage{gensymb}
\usepackage{enumitem}
\usepackage{makecell}
\usepackage{dsfont}
\usepackage{upquote}
 \usepackage[preprint]{neurips_2026}

\usepackage[utf8]{inputenc} % allow utf-8 input
\usepackage[T1]{fontenc}    % use 8-bit T1 fonts

\usepackage[
    colorlinks=true,
    urlcolor=RoyalBlue
]{hyperref}
\usepackage{url}            % simple URL typesetting
\usepackage{booktabs}       % professional-quality tables
\usepackage{amsfonts}       % blackboard math symbols
\usepackage{nicefrac}       % compact symbols for 1/2, etc.
\usepackage{microtype}      % microtypography
\usepackage[dvipsnames]{xcolor}        % colors
\usepackage{listings}
\input{docs/listings_styles}
\usepackage{paralist}
\usepackage{amsmath}
\usepackage{graphicx}
\usepackage{makecell}
\usepackage{wasysym}
\usepackage{pifont}
\usepackage{float}
\usepackage{subcaption}
\usepackage{caption} 
\usepackage{enumitem}

\def\name{IDEAL-Bench}

\title{IDEAL-Bench: Indoor Dataset and Evaluation suite for Analyzing 3D Layout reasoning}

\author{
  Yuening Cai \quad Junwei Zhou \quad Youran Qu \quad Yu-Wing Tai$^\dag$ \\[0.6em]
  Department of Computer Science \\[0.3em]
  Dartmouth College \\[0.6em]
\url{https://ideal3d.github.io/}
}
\begin{document}
\maketitle

\renewcommand{\thefootnote}{}
\footnotetext{$^\dag$Corresponding author}

\vspace{-0.05in}

\begin{figure}[h!]
    \centering
    \small
    \includegraphics[width=1\linewidth]{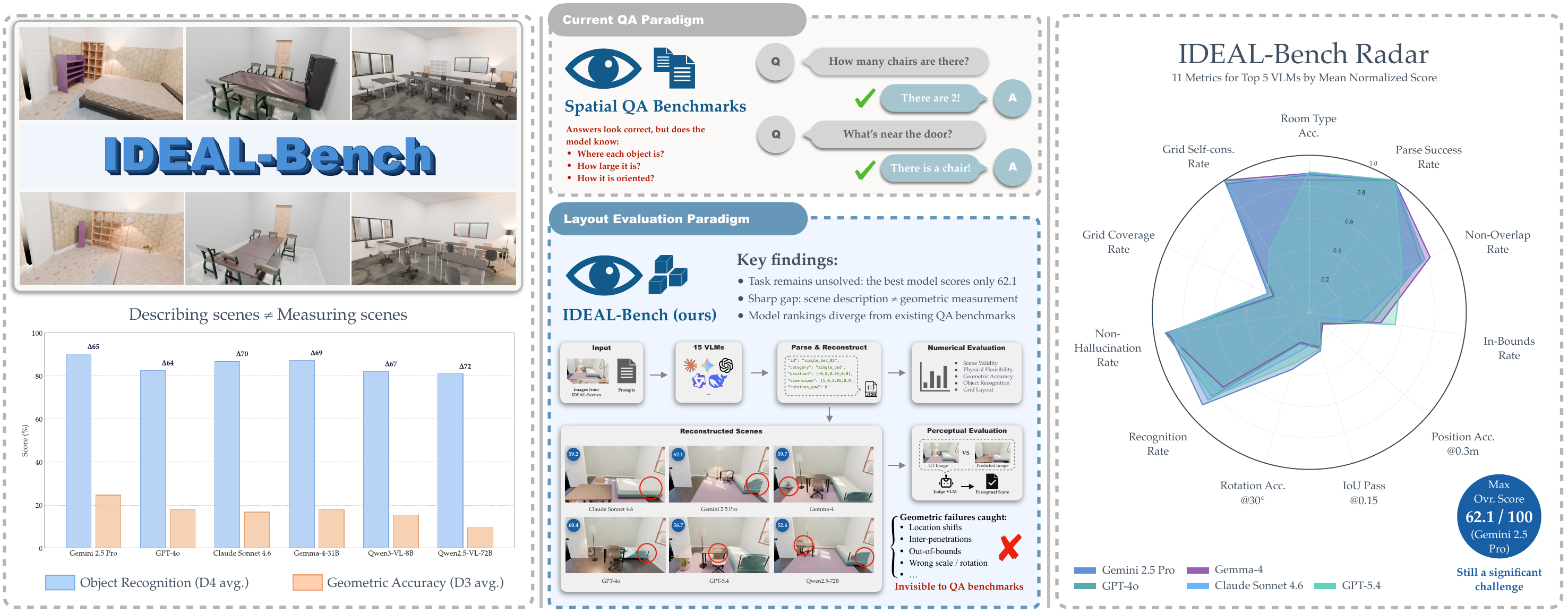}\\
    \vspace{.05in}
    \caption{\textbf{Towards better evaluation of spatial intelligence in VLMs}. While existing 3D QA benchmarks probe relational understanding, IDEAL-Bench requires models to estimate every visible object's 3D pose and extent, evaluated via numerical metrics and a render-and-compare protocol. The task demands genuine spatial reasoning and remains substantially unsolved, with the strongest model reaching only 62.1/100 overall.}
    \label{fig:teaser}
\end{figure}
\vspace{.15in}

\input{sections/abstractv1}

\input{sections/1intro}

\input{sections/2related_work}

\input{sections/3task}

\input{sections/4dataset}

\input{sections/5bench}

\input{sections/6experiments}
\input{sections/7concl}
\input{sections/8discussion}

\newpage
{
    \small
    \bibliographystyle{unsrt}
    \bibliography{main}
}

%%%%%%%%%%%%%%%%%%%%%%%%%%%%%%%%%%%%%%%%%%%%%%%%%%%%%%%%%%%%
\newpage
\appendix

\input{sections/AppA_dataset}
\input{sections/AppB_evaluation}

\input{sections/AppC_setup}
\input{sections/AppD_results}
\input{sections/AppE_discussion}

\input{sections/AppF_broader_impacts}

%%%%%%%%%%%%%%%%%%%%%%%%%%%%%%%%%%%%%%%%%%%%%%%%%%%%%%%%%%%%

% \newpage
% \input{checklist.tex}

\end{document}

%% file: docs/listings_styles.tex
\lstdefinestyle{promptstyle}{
  basicstyle=\ttfamily\footnotesize,
  breaklines=true,
  breakatwhitespace=true,
  columns=fullflexible,
  keepspaces=true,
  showstringspaces=false,
  frame=single,
  framesep=6pt,
  rulecolor=\color{black!30},
  backgroundcolor=\color{black!3},
  xleftmargin=8pt,
  xrightmargin=8pt,
  aboveskip=8pt,
  belowskip=8pt,
  literate=
    {—}{{---}}1
    {–}{{--}}1
    {±}{{$\pm$}}1
    {×}{{$\times$}}1
    {°}{{$^\circ$}}1
    {≤}{{$\leq$}}1
    {≥}{{$\geq$}}1
    {∈}{{$\in$}}1
    {→}{{$\to$}}1
}

\lstdefinelanguage{json}{
  morestring=[b]",
  morestring=[d]',
  literate=
    *{0}{{{\color{black!70}0}}}1
     {1}{{{\color{black!70}1}}}1
     {2}{{{\color{black!70}2}}}1
     {3}{{{\color{black!70}3}}}1
     {4}{{{\color{black!70}4}}}1
     {5}{{{\color{black!70}5}}}1
     {6}{{{\color{black!70}6}}}1
     {7}{{{\color{black!70}7}}}1
     {8}{{{\color{black!70}8}}}1
     {9}{{{\color{black!70}9}}}1
     {:}{{{\color{black!90}{:}}}}1
     {,}{{{\color{black!90}{,}}}}1
     {\{}{{{\color{black}{\{}}}}1
     {\}}{{{\color{black}{\}}}}}1
     {[}{{{\color{black}{[}}}}1
     {]}{{{\color{black}{]}}}}1,
}
\lstdefinestyle{jsonstyle}{
  language=json,
  basicstyle=\ttfamily\footnotesize,
  stringstyle=\color{teal!70!black},
  commentstyle=\color{gray}\itshape,
  morecomment=[l]{//},
  showstringspaces=false,
  breaklines=true,
  breakatwhitespace=false,
  columns=fullflexible,
  keepspaces=true,
  frame=single,
  framesep=6pt,
  rulecolor=\color{black!30},
  backgroundcolor=\color{black!3},
  xleftmargin=8pt,
  xrightmargin=8pt,
  aboveskip=8pt,
  belowskip=8pt,
}

%% file: sections/abstractv1.tex
\begin{abstract}
Spatial question answering is the dominant paradigm for evaluating spatial intelligence in Vision-Language Models (VLMs), but it leaves a complementary axis of spatial competence under-evaluated: holistic 3D layout inference, which predicts every visible object's pose and extent from a single image in a structured form. 
To this end, we introduce IDEAL-Bench, an evaluation suite that requires VLMs to predict structured 3D layouts on photorealistic indoor scenes across 10 room types, scored along five numerical dimensions (scene validity, physical plausibility, geometric accuracy, object recognition, and grid layout) and a perceptual render-and-compare protocol. By operating on semantically realistic scenes with full asset substitution under controlled lighting and viewpoint, IDEAL-Bench moves beyond %CLEVR-style 
simple geometric primitives so that any image-space discrepancy reflects spatial reasoning alone. 
The benchmark is built on IDEAL-Scenes, a procedurally generated dataset of 1,000 re-renderable Blender environments with ground-truth layouts. Evaluating 15 prominent VLMs reveals three findings: 
(i) the task remains substantially unsolved, with the strongest model reaching only 62.1/100 overall; 
(ii) all models exhibit a sharp asymmetry between object recognition and geometric regression, indicating that current VLMs are trained to describe scenes rather than to measure them;
(iii) model rankings partially diverge from those on QA-based and primitive-reconstruction benchmarks: top-tier consensus holds, but mid-tier rankings shift substantially.
Collectively, these findings establish IDEAL-Bench as a diagnostic suite, targeting the geometric and structural competencies that QA-based evaluation cannot surface, and paving the way towards more rigorous evaluation of spatial intelligence in next-generation VLMs.
% Together, these findings position IDEAL-Bench as a principled diagnostic for whether future VLMs achieve genuine spatial understanding rather than linguistic approximations of it.
\end{abstract}

%% file: sections/1intro.tex
\section{Introduction}
\label{sec:intro}

Vision-Language Models (VLMs) have made remarkable strides in visual perception and language understanding, though 3D spatial reasoning remains one of their most persistent and consequential weaknesses \citep{zha, spatialreasoner, ma}. Closing this gap is essential for applications that interact with the physical world, such as augmented reality anchoring and robotic navigation, where models must go beyond describing a scene to accurately localizing themselves and objects within it.

The research community has established a foundation through diverse spatial benchmarks, most of which evaluate competence via question answering or object grounding. These evaluations have been instrumental in measuring relational reasoning, directional understanding, and depth ordering. 
However, success on relational queries does not by itself require a model to maintain a coherent, globally consistent 3D representation of the scene. A complementary axis, namely the ability to infer a structured geometric layout in absolute terms, remains comparatively under-evaluated.

We introduce \textit{holistic 3D layout inference} as a new evaluation protocol: from a single image, a model must produce a globally consistent 3D layout that is jointly verifiable against ground truth, rather than answering isolated relational queries. This framing makes spatial competence directly auditable in a way QA-based protocols cannot.
To instantiate this paradigm, we introduce \textbf{IDEAL-Bench}, which requires VLMs to predict each object's 3D position, orientation, and scale from a single RGB image. Predictions are evaluated along two dimensions: a suite of numerical metrics measuring geometric accuracy against ground-truth annotations, and a perceptual render-and-compare assessment validated against human judgments. This dual verifiability (i.e., numerical and perceptual) makes evaluation falsifiable in ways QA cannot achieve.
Underpinning IDEAL-Bench is \textbf{IDEAL-Scenes}, a dataset of 1,000 indoor scenes across 10 room types built via a scalable generation pipeline; each scene ships with programmatically extracted ground-truth layouts and the original Blender source files, enabling re-rendering and reproducible evaluation. Unlike prior benchmarks that probe spatial competence through QA or single-object grounding, IDEAL-Bench is the first to require holistic, scene-level layout estimation with re-renderable verification (Table~\ref{tab:related_comp}).

Our evaluation of 15 prominent VLMs reveals failure modes invisible to QA-based protocols. While nearly all models produce well-formatted outputs and identify scene objects correctly, spatial-related estimation remains a major issue for all VLMs; even the strongest model achieves only 62.1/100 overall. We further uncover a sharp divergence in how models handle scene complexity: mid-tier VLMs degrade rapidly as object count grows, while frontier models do not degrade monotonically, suggesting they rely on learned structural priors rather than per-object geometric inference.
%while frontier models maintain near-flat performance, suggesting they rely on learned structural priors rather than per-object geometric inference. 
Together, these findings position IDEAL-Bench as a diagnostic complement to existing benchmarks: it surfaces precisely the geometric deficiencies that linguistic spatial reasoning hides.

Our main contributions are threefold:
\begin{itemize}[leftmargin=20pt, itemsep=2pt, topsep=2pt, parsep=0pt, partopsep=0pt]
\item We introduce holistic 3D layout inference as a new evaluation paradigm for VLM spatial competence, requiring structured, geometrically verifiable estimations of semantic-aware object pose and extent from a single image.

\item We release IDEAL-Scenes, a dataset of 1,000 indoor scenes across 10 room types with programmatically extracted ground-truth layouts and re-renderable Blender source files, together with IDEAL-Bench, an evaluation suite combining numerical geometric metrics with perceptual render-and-compare assessment validated against human judgments.

\item Our evaluation of 15 VLMs uncovers a critical gap between linguistic proficiency and geometric precision, identifying coordinate estimation as a primary bottleneck.
These findings demonstrate that IDEAL-Bench serves as a necessary complement to existing benchmarks by surfacing hidden geometric deficiencies in holistic 3D reasoning.
\end{itemize}

%% file: sections/2related_work.tex
\vspace{-0.05in}
\section{Related Work}
\label{sec:related}

\input{tables/relatedworks}

\noindent \textbf{3D Spatial Understanding and Layout Inference.}
Large VLMs~\cite{gemini, gemini2.5pro, gpt4, gpt4o, gpt5, sonnet4.6, gemma, qwen2.5, qwen3, glm4.6, internvl, llava, januspro} have demonstrated strong performance across visual perception and reasoning tasks, and recent work has evaluated their spatial capabilities through 3D visual
question answering, language-guided grounding, 3D captioning, and embodied
navigation~\cite{zha, windecker2025navitrace, liu}. 
Despite promising results, it remains unclear whether these models actually recover 3D scene structure or exploit local visual cues and statistical regularities~\cite{ma}, a gap that current evaluation protocols may systematically underestimate.
Explicit 3D layout provides a structured and measurable interface for probing this question directly. Prior work spans layout-driven generation~\cite{li2024advances, wen20253d, liu2024comprehensivesurvey3dcontent, zhou2026gena, zhou2024layout}, which uses explicit spatial representations to guide controllable 3D synthesis~\cite{chen2025sam,xie2024physgaussian, lin2025pat3d, zhou2025coco}, and layout inference~\cite{wang2025tabletopgen, ling2025scenethesislanguagevisionagentic, zhou2026perceivethenplan}, where
VLMs predict 3D arrangements from language descriptions~\cite{Gu_2025_CVPR, feng2023layoutgpt, Yang_2024_CVPR, liu2025worldcraft}. Despite this progress, recovering complete and coherent room-scale 3D layouts from a \emph{single image} remains underexplored
and lacks systematic evaluation, a gap that IDEAL-Bench is designed to fill.

\noindent \textbf{Evaluating Spatial Reasoning in VLMs.} 
Existing benchmarks for VLM spatial reasoning rely on indirect proxies rather than explicitly measuring global 3D understanding.
Current approaches include question-answering (QA)~\cite{vsibench, spatialvlm, spatialbench, omnispatial, space10} and language-guided 3D grounding~\cite{scanrefer, referit3d}, but these share a key limitation. 
QA-based benchmarks use multiple-choice or numeric answers, where models can rely on local visual cues without building a global 3D representation~\cite{ma}.
Language-guided 3D grounding localizes objects from 3D inputs and text, thus only evaluating spatial understanding indirectly.
Even recent benchmarks that move toward explicit spatial perception, such as Space3D-Bench~\cite{space3d}, SpatialRGPT~\cite{spatialrgpt}, and SPAR-Bench~\cite{sparbench}, still rely on per-question outputs and do not require holistic scene reconstruction~\cite{zhang}. 
The closest baseline, IR3D-Bench~\cite{ir3dbench}, uses a render-and-compare protocol but is confined to simple synthetic primitives, observed from a single camera held fixed across all samples, lacking the object complexity, large viewpoint variation, and realistic scene diversity that genuine global 3D understanding demands.
To address this gap, we introduce IDEAL-Bench, built on IDEAL-Scenes, which evaluates models across 5 dimensions and 11 metrics, with a Judge-VLM protocol for holistic 3D spatial understanding (Section~\ref{sec:bench}).

\noindent \textbf{Indoor Scene Datasets.} While real-world datasets~\cite{scannet, matterport3d, arkit, sun3d, sunrgbd, scenenn} capture authentic spatial layouts, they suffer from noisy geometry, imperfect annotations, and lack re-renderable representations. Synthetic datasets~\cite{scenenetrgbd,
interiornet, structure3d, hypersim, 3dfront, ase, m3dlayout} guarantee exact ground truth but face distinct limitations: 3D-FRONT~\cite{3dfront} lacks pre-rendered images, ASE~\cite{ase} does not include scene assets, and Hypersim~\cite{hypersim} restricts re-rendering due to commercial licenses. Consequently, no existing dataset simultaneously delivers re-renderable scene files, rendered images, consistent assets, programmatic ground truth extraction, and procedural scalability. IDEAL-Scenes bridges this gap by releasing native Blender scene files generated via a procedural pipeline based on InfiniGen~\cite{infinigen}, enabling the holistic 3D spatial understanding task defined in Section~\ref{sec:task}.

%% file: tables/relatedworks.tex
\begin{table}[t!]
\centering
\caption{Comparison of \name{} with representative spatial datasets and benchmarks across five capability dimensions. \checkmark{}~denotes full support, $\LEFTcircle$~partial coverage, and \ding{55}~no support.}
\vspace{.07in}
\resizebox{\textwidth}{!}{
\begin{tabular}{l ccccc}
\toprule
Dataset \& Benchmark & \makecell{Geometry\\Accuracy} & \makecell{Category\\Coverage} & \makecell{Physical\\Plausibility} & \makecell{Scene\\Diversity} & \makecell{Perceptual\\Judge}\\
\midrule
IR3D-Bench~\cite{ir3dbench}     & \checkmark     & $\LEFTcircle$ & \ding{55}     & \ding{55}     & \checkmark\\
VSI-Bench~\cite{vsibench}                 & $\LEFTcircle$  & $\LEFTcircle$ & \ding{55}     & \checkmark    & \ding{55}\\
SPAR-Bench~\cite{sparbench}                & $\LEFTcircle$  & \checkmark    & \ding{55}     & \checkmark    & \ding{55}\\
SpatialRGPT-Bench~\cite{spatialrgpt}         & $\LEFTcircle$  & \checkmark    & \ding{55}     & \checkmark    & \ding{55}\\
OmniSpatial~\cite{omnispatial}               & $\LEFTcircle$  & $\LEFTcircle$ & \ding{55}     & \checkmark    & \ding{55}\\
\midrule
\name{} (ours)                    & \checkmark     & \checkmark    & \checkmark    & \checkmark    & \checkmark\\
\bottomrule
\end{tabular}
}
\vspace{-0.15in}
\label{tab:related_comp}
\end{table}

%% file: sections/3task.tex
\section{Task Formulation}
\label{sec:task}
We formulate the task as a structured 3D layout prediction. Given a single RGB image $\mathbf{I} \in \mathbb{R}^{H \times W \times 3}$, 
a list of candidate object categories $\mathcal{C} = \{c_1, c_2, \ldots, c_m\}$ 
that may appear in the scene (not guaranteed to appear, and possibly with 
multiple instances), and a set of candidate room types $\mathcal{T}$, a model 
must produce a structured prediction 
$\hat{\mathcal{S}} = (\hat{t}, \hat{\mathbf{r}}, \{\hat{s}_1, \hat{s}_2, \ldots, \hat{s}_{\hat{n}}\})$, 
where $\hat{t} \in \mathcal{T}$ is the predicted room type, 
$\hat{\mathbf{r}} = (\hat{W}, \hat{D}, \hat{H}) \in \mathbb{R}^3$ are the predicted 
room dimensions along the lateral (+X), depth (+Y), and vertical (+Z) axes, 
and each object entry $\hat{s}_i$ specifies:

\begin{itemize}[leftmargin=20pt, itemsep=2pt, topsep=2pt, parsep=0pt, partopsep=0pt]
\item $\hat{c}_i \in \mathcal{C}$: the object category, which must be drawn 
exactly from the candidate list.
\item $\hat{\mathbf{p}}_i \in \mathbb{R}^3$: the 3D center coordinates of the object (in meters), in a room-centered coordinate frame with origin at the center of the floor, +Y pointing into the depth of the scene, +X right and +Z up.
\item $\hat{\mathbf{d}}_i \in \mathbb{R}^3$: the dimensions of the bounding-box $(w, d, h)$ in meters, where $(w, d, h)$ denote the extents along the canonical lateral (+X), depth (+Y), and vertical (+Z) axes respectively, defined under canonical orientation.
\item $\hat{\theta}_i \in [0^\circ, 360^\circ)$: the yaw rotation around the
+Z axis, expected to be discretized to one of four cardinal orientations
$\{0^\circ, 90^\circ, 180^\circ, 270^\circ\}$ for the large majority of
objects (aligning with the IDEAL-Scene setting to alleviate the reasoning
burden of VLMs); a small minority are intentionally placed at oblique
angles to probe whether models genuinely ground predictions in the image
rather than defaulting to this prior.
\end{itemize}

\begin{figure}[t!]
    \centering
    \small
    \includegraphics[width=1\linewidth]{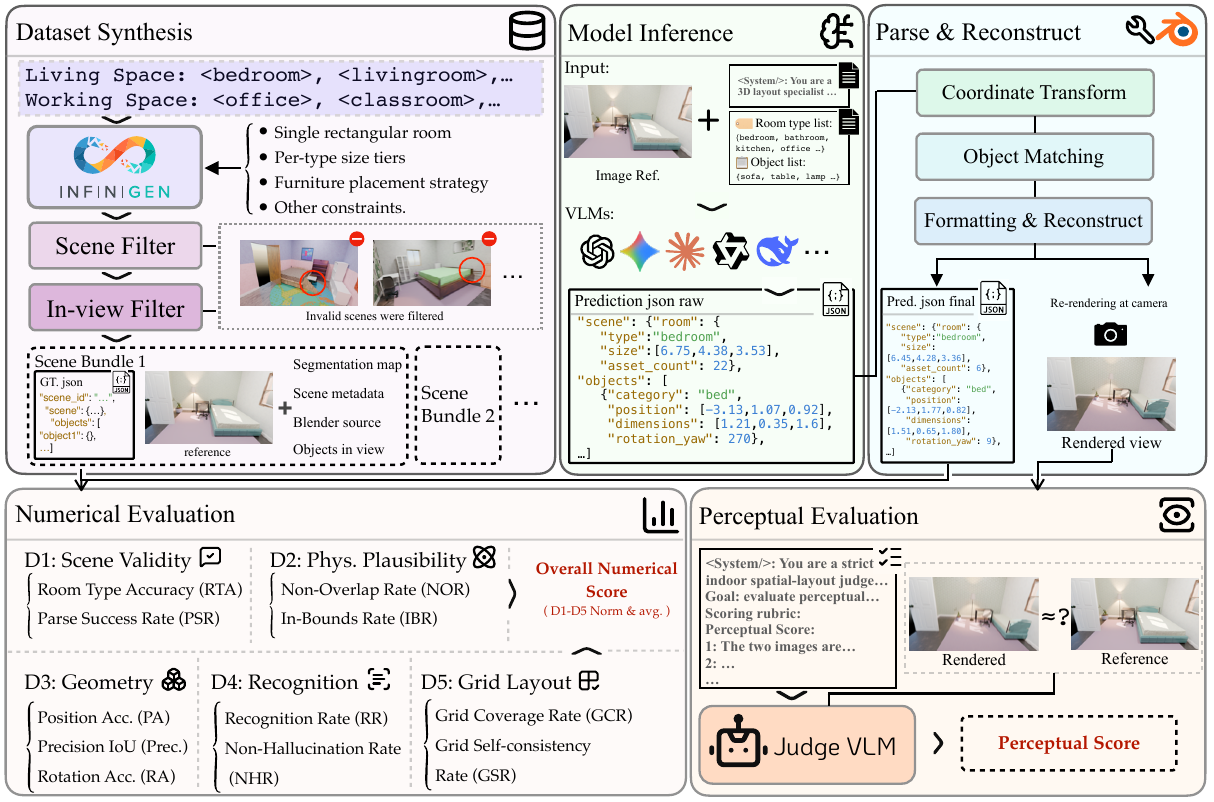}\\
    % \vspace{-0.05in}
    \caption{\textbf{Overview of IDEAL-Bench.} Four stages. (1) \emph{Dataset synthesis}: we modified InfiniGen to produce preprocessed, self-contained scenes across 10 room types.  (2) \emph{Model inference}: a VLM predicts a structured layout from an image, a system prompt and a user prompt containing a room type, and a category list. (3) \emph{Parse \& reconstruct}: predicted pose and position are applied to GT assets and re-rendered from the original camera viewpoint. (4) \emph{Evaluation}: D1–D5 numerical metrics plus a Judge VLM perceptual score to provide a comprehensive evaluation of VLMs.}
    \label{fig:pipeline}
    \vspace{-.25in}
\end{figure}

The task is delivered via a structured prompt: geometric rules, estimation steps, coordinate conventions, and output schema form the system prompt, while the image $\mathbf{I}$, category list $\mathcal{C}$, and room type candidates $\mathcal{T}$ form the user prompt. The model receives a prior on camera height (%1.2m–2.0m above the floor, consistent with typical human eye-level photography
1.2m–2.0m above the floor, varied around typical eye level to introduce viewpoint diversity), but no depth, camera intrinsics or extrinsics, or instance-level annotations are provided. The models must reason about 3D structure purely from monocular appearance. The full system prompt is provided in Appendix~\ref{app_c2}.

The structured output $\hat{\mathcal{S}}$ is a falsifiable geometric claim about the scene that admits two parallel forms of evaluation, together constituting IDEAL-Bench (Sec.~\ref{sec:bench}). First, each predicted attribute can be compared directly against the ground-truth 
layout $\mathcal{S}^*$ through numerical metrics that quantify positional, dimensional, and rotational errors. Second, $\hat{\mathcal{S}}$ can be instantiated by substituting ground-truth 
assets at predicted poses and re-rendering from the original viewpoint, yielding a reconstructed image $\hat{\mathbf{I}}$ for perceptual comparison against the reference $\mathbf{I}$.
Both evaluation pathways impose concrete requirements on the underlying 
dataset: %programmatically exact ground-truth layouts in a machine-readable 
%format, and re-renderable scene files with all asset geometries preserved. 
programmatically exact ground-truth layouts for machine-readable numerical evaluation, and re-renderable scene files that preserve full asset geometry for human-interpretable perceptual inspection.
This motivates the design of IDEAL-Scenes, a synthetic dataset 
purpose-built to support both evaluation modes, described in 
Section~\ref{sec:dataset}.

%% file: sections/4dataset.tex
\section{IDEAL-Scenes Dataset}
\label{sec:dataset} 
IDEAL-Scenes is a synthetic indoor dataset of 1,000 scenes spanning 10 room categories, purpose-built to support our evaluation protocol. Each scene is a bundle comprising a rendered RGB image, per-asset ground-truth annotations \texttt{GT.json}, a Blender source file \texttt{scene.blend}, an instance segmentation map, and scene metadata. Together, these assets support both direct numerical comparison against ground truth and re-rendering of the scene from the original viewpoint after substituting predicted layouts, as well as further downstream analysis (see Appendix~\ref{app_a4}). 
Because the original and reconstructed renders share the same assets, camera, and lighting, any image-space discrepancy reflects spatial reasoning error isolated from confounders.

\noindent \textbf{Synthetic Scene Generation.} Real-world datasets introduce two obstacles: annotation noise from depth sensors and human labeling undermines render-and-compare precision, and original meshes are typically unavailable due to licensing restrictions. Synthetic generation resolves both: ground-truth poses are extracted programmatically with no labeling noise, and full Blender source files are retained. We build on InfiniGen~\cite{infinigen}, which supports constraint-based furniture placement and natively exports per-asset geometry, pose, and camera parameters in Blender.

\noindent \textbf{InfiniGen Engine Modifications.} %InfiniGen's default configuration emphasizes scene diversity through multi-room floor plans and wide variance in lighting and furniture density. However, these properties are detrimental to a benchmark where metric differences should reflect model capability rather than generation randomness. 
InfiniGen's default configuration emphasizes scene diversity through multi-room floor plans and high randomness in lighting and furniture density. However, implausible layouts and inconsistent lighting confound model evaluation, particularly for the perceptual protocol, since metric differences would then reflect generation artifacts rather than genuine differences in model capability.
We introduce three targeted modifications: (i) restricting room geometry to a single rectangular room with category-specific floor dimension ranges; (ii) standardizing lighting and rendering parameters across the dataset; (iii) enforcing furniture placement constraints for physical plausibility and minimum accessibility. Full specifications are in Appendix~\ref{app_a1}.

\noindent \textbf{Preprocessing Scenes.} After generation, each scene passes through a two-stage preprocessing pipeline. Quality filtering discards scenes with no visible furniture (all instance segmentation masks with fewer than 300 pixels of occupied area are treated as not visible) and any scene containing pairwise mesh interpenetration, verified via bounding-box pre-screening followed by Boolean intersection (with a voxel-grid and surface-proximity fallback for non-watertight meshes). In-view object determination then retains objects with sufficient visual evidence using three filters on instance-mask size (the same 300-pixel threshold), aspect ratio, and boundary proximity; full criteria in Appendix~\ref{app_a3}.

\noindent \textbf{Scene Coverage.} IDEAL-Scenes covers 10 room types (bedroom, home studio, living room, kitchen, bathroom, dining room, office, meeting room, classroom, and library), chosen as the primary indoor environments for real-world applications.
Living-space rooms feature asymmetric arrangements, while working-space rooms (e.g., classrooms) exhibit repetitive grid layouts, enabling IDEAL-Bench to probe whether VLMs generalize across both irregular and highly structured configurations. Each type is governed by a distinct constraint set (floor dimensions, furniture count, allowed categories) with 100 scenes per type to ensure balanced evaluation across categories.

%% file: sections/5bench.tex
\section{Evaluation Protocol}
\label{sec:bench}

\subsection{Overview}
IDEAL-Bench evaluates structured layout predictions through two parallel 
assessment pathways. \textbf{Numerical metrics} compare predicted object 
attributes directly against ground-truth layouts $\mathcal{S}^*$, 
quantifying geometric errors at the object level. 
\textbf{Perceptual metrics} assess layout quality at the scene level by comparing the reference image $\mathbf{I}$ against the reconstructed image $\hat{\mathbf{I}}$, capturing scene-level plausibility that object-level attributes cannot fully reflect.

\noindent \textbf{Coordinate Alignment.}
To evaluate spatial understanding across diverse viewpoints, camera positions are initially sampled at random near the four cardinal directions and oriented toward the center of the room, before the generation engine selects the optimal angle for each scene. 
Since the model predicts layouts in the image-view coordinate frame, we first align the prediction to the canonical world frame before comparing it with the ground truth:

\begin{equation}
  \hat{S} = T_{\mathrm{align}} \cdot \hat{S}_{\mathrm{cam}},
\end{equation}

where $T_{\mathrm{align}}$ rotates the predicted positions and yaw angles into the canonical world frame. We approximate $T_{\mathrm{align}}$ in closed form: we round the camera's known yaw to the nearest of the four cardinal directions (ignoring pitch and roll), and use this rounded angle to compute the required rotation. This is consistent with how the model is prompted: it is asked to treat the wall that it is most directly facing as +Y. However, this method introduces a trade-off for scenes viewed near a 
room diagonal, where the model's implicit choice of reference wall could diverge from the discretized ground truth. We find, however, that this divergence does not manifest as a uniform degradation across metrics, suggesting it is not the dominant source of error at diagonal viewpoints (Appendix~\ref{app_viewpoint}).

\noindent \textbf{Vertical Position Convention.}
% Model predictions include a full 3D position $(x, y, z)$ for each object. However, the predicted $z$ is tightly coupled to predicted object dimensions: an overestimated height shifts the geometric center upward, introducing systematic $z$ errors that reflect dimension inaccuracy rather than spatial layout understanding. We therefore adopt a category-aware convention. For floor-standing objects, $z$ is fixed to the ground-truth value in both numerical evaluation and reconstruction, so positional error is computed only on the horizontal plane. For wall-mounted objects (e.g., mirrors, wall art), the predicted $z$ is retained, since mounting height is independently meaningful.
Model predictions include a full 3D position $(x, y, z)$ for each object. However, the predicted $z$ is tightly coupled to predicted object dimensions: an overestimated height shifts the geometric center upward, introducing systematic $z$ errors that reflect dimension inaccuracy rather than spatial layout understanding. In pilot experiments, scoring predicted $z$ directly caused floor-standing objects to float or interpenetrate the ground at rates that made both the geometric metrics and the perceptual reconstruction uninformative. We therefore adopt a category-aware convention: for floor-standing objects, $z$ is fixed to the ground-truth value in both numerical evaluation and reconstruction, and we compute the positional error only for the horizontal plane; for wall-mounted objects (e.g., mirrors, wall art), the $z$ prediction is retained, since mounting height is independently meaningful. Nonetheless, the evaluated models are still required to predict full 3D position at inference, including $z$ parameter (Sec.~\ref{sec:task}); this convention only controls for the axis along which current dimension estimates do not yet provide a reliable basis for comparison.

\noindent \textbf{Object Matching.}
Predicted-to-ground-truth correspondences are required for both numerical 
evaluation and reconstruction. We partition scenes into two categories: 
\textbf{Non-Grid Scenes} (e.g., bedrooms, living rooms) contain few 
same-category instances; we assign correspondences via category-wise Hungarian 
matching, with cost defined as the Euclidean distance between predicted and 
ground-truth centers projected onto the horizontal plane. 
Matching is performed 
without a distance cutoff, so every predicted object is unconditionally assigned 
to its nearest same-category counterpart; this may inflate position error when 
predictions are far off, but leaves recognition metrics ($\text{RR}$, 
$\text{NHR}$) unaffected, as these depend only on object counts. 
\textbf{Grid-structured scenes} (classrooms, libraries, offices) contain dozens 
of same-category instances in regular patterns, where per-instance assignment is 
largely arbitrary and per-instance error fails to capture whether the prediction 
reflects the underlying regularity; these scenes bypass Hungarian matching 
entirely for numerical evaluation and are assessed via the dedicated D5 metric 
(Sec.~\ref{sec:metrics}).

\vspace{-.1in}
\subsection{Metrics Design}
\label{sec:metrics}

\noindent \textbf{Numerical Metrics.} We organize numerical metrics into five dimensions, each targeting a functionally distinct aspect of spatial competence (full table in Appendix~\ref{app_b3} Tab.~\ref{tab:metric_scenetypes}). 
D2 (Physical Plausibility), D3 (Geometric Accuracy), and D5 (Grid Layout) decompose geometric reasoning along independently failable axes: boundary compliance, metric localization, and periodic regularity. D5 in particular targets grid-structured scenes (e.g., classrooms with dozens of same-category instances), where per-instance Hungarian matching is ill-posed and meaningful evaluation must instead measure structural regularity rather than per-object correspondence. D1 (Scene Validity) and D4 (Object Recognition) follow established conventions in scene parsing and object detection.
Each dimension $\textsc{D}_k$ is normalized to $[0, 100]$ by averaging its per-metric means. %(D3--D4 over non-grid scenes; D5 over grid scenes; D1--D2 over all). The per-model \textbf{Overall} score is the unweighted mean of all 11 applicable rate metrics.
Because D3 and D4 rely on object-level correspondence, they are computed over non-grid scenes only; D5 is computed over grid scenes only, where it instead measures structural regularity; D1 and D2 require no correspondence and are computed over all scenes. 
Within each room type, only the dimensions applicable to that type (\{D1, D2, D3, D4\} for non-grid rooms, \{D1, D2, D5\} for grid rooms) are averaged into a per-room-type Overall (full derivation in Appendix~\ref{app_b4}). The aggregate Overall reported in Table~\ref{tab:ideal_leaderboard} (Eq.~\ref{eq:overall} ) then pools across room types, taking the unweighted mean of all 11 applicable rate metrics over the full dataset.
We adopt unweighted averaging to avoid privileging any single dimension a priori; per-dimension scores remain available for fine-grained analysis (Sec.~\ref{sec:exp}):
% \begin{equation}
%   \mathrm{D}_{\text{overall}} = \tfrac{1}{11}
%   \Bigl(
%     \underbrace{\text{PSR} + \text{RTA}}_{\text{D1}}
%     + \underbrace{\text{NOR} + \text{IBR}}_{\text{D2}}
%     + \underbrace{\text{PA} + \text{Prec.} + \text{RA}}_{\text{D3}}
%     + \underbrace{\text{RR} + \text{NHR}}_{\text{D4}}
%     + \underbrace{\text{GCR} + \text{GSR}}_{\text{D5}}
%   \Bigr),
%   \label{eq:overall}
% \end{equation}
\begin{equation}
  \mathrm{D}_{\text{overall}} = mean
  \Bigl(
    \underbrace{\text{PSR} + \text{RTA}}_{\text{D1}}
    + \underbrace{\text{NOR} + \text{IBR}}_{\text{D2}}
    + \underbrace{\text{PA} + \text{Prec.} + \text{RA}}_{\text{D3}}
    + \underbrace{\text{RR} + \text{NHR}}_{\text{D4}}
    + \underbrace{\text{GCR} + \text{GSR}}_{\text{D5}}
  \Bigr),
  \label{eq:overall}
\end{equation}
% Is it better to make it more sturctured, like using the summation notation...?

The Geo. score is the equal-weighted mean of the available PA, Prec., RA (from D3), and GCR, GSR (from D5) rates, as shown in Tab.~\ref{tab:ideal_leaderboard}. Full definitions are in Appendix~\ref{app_b3} and~\ref{app_b4}.

\noindent \textbf{Scene Reconstruction.} Numerical metrics capture local geometric fidelity but cannot reflect whether a reconstructed scene \emph{appears} spatially correct. We therefore evaluate each prediction perceptually by re-rendering it. For every object listed in a scene's \texttt{objects\_in\_view} file, the predicted position, yaw, and dimensions replace the ground-truth values in the original \texttt{scene.blend}; architectural elements, out-of-view objects, camera, and lighting remain unchanged. 
Predicted objects without a ground-truth correspondence (i.e., hallucinations) are rendered as red spheres to make false positives visually salient, while unmatched ground-truth objects are hidden. This visualization makes hallucination explicit to the \textbf{Judge VLM} rather than silently degrading reconstruction quality.
The scene is then re-rendered to produce $\hat{\mathbf{I}}$. 
Full reconstruction details are in Appendix~\ref{app_b5}.

\noindent \textbf{Perceptual metrics.} To complement the numerical metrics with a perceptual measure of room-level spatial plausibility, we use Claude Opus 4.7 as a \textbf{Judge VLM} to compare each rendering of reconstruction against its ground-truth image. We restrict this evaluation to the six representative models on 200 scenes (20 per room type), 
%as weaker models frequently produce reconstructions with missing assets or failed composition that are not meaningful to score. 
as weaker models frequently produce estimated reconstructions with objects scattered out of bounds or overlapping with each other, leaving sparse, incoherent scenes that are not meaningful to score.
For each scene, a single API call presents the ground truth alongside all six reconstructions in randomized order, and the judge assigns each a 1--5 Likert score focused on spatial similarity rather than photorealism, together with a strict total ranking that breaks ties among equally-scored reconstructions. We repeat the judging five times and report averaged scores. The full protocol details are in Appendix~\ref{app_b6} and human-perceptual justifications for the Judge VLM are shown in Sec.~\ref{subsec:perceptual}.

%In summary, numerical metrics quantify geometric correctness while Judge-VLM scores capture perceptual spatial plausibility, providing complementary windows into VLM 3D understanding.

%% file: sections/6experiments.tex
\section{Experiments}
\label{sec:exp}
\subsection{Experimental Setup}
We benchmark 15 VLMs spanning proprietary and open-weight families, with parameter scales ranging from 7B to 235B. %and architectural variants including dense, mixture-of-experts, and unified vision-language backbones. 
The proprietary set comprises Gemini 2.5 Pro, GPT-4o, Claude Sonnet 4.6, and GPT-5.4; open-source models include Gemma-4-31B-IT, GLM-4.6V (and -Flash), Qwen2.5-VL-72B, Qwen3-VL at 8B/30B/235B, LLaVA-1.6-Mistral-7B, InternVL3.5 at 8B/30B-A3B, and Janus-Pro-7B. Each model is queried once per scene with a single forward pass on monocular RGB across all 1,000 scenes. 
%without chain-of-thought, depth input, camera parameters, or instance masks. 
Inference protocol, system prompts, and parse-recovery rules are detailed in Appendix~\ref{app_c}.

%########################################################################################################################################################################################################################################################################################################################

\input{tables/mainresult}
\subsection{Numerical Results}
\label{sec:numerical_results}

\noindent\textbf{Statistical robustness.}
All results in Table~\ref{tab:ideal_leaderboard} are point estimates over the 1,000-scene benchmark from a single forward pass per model. To assess how much of the reported variation reflects genuine differences in model capability versus sampling noise over our scene set, we compute 95\% bootstrap confidence intervals (2,000 resamples) for every metric; the full results and their implications for specific comparisons are reported in Appendix~\ref{app_d6}.

\noindent\textbf{Strong recognition, weak measurement.}
The leaderboard (Tab.~\ref{tab:ideal_leaderboard}) reveals a paradigm-level asymmetry that is consistent across all evaluated VLMs, regardless of family or scale. 
On one hand, D1 (Scene Validity) is near-saturated (PSR $\ge$ 99\% for 8 of 13 parsing-successful models) and D4 (Object Recognition) is uniformly strong (RR $\ge$ 70\% except for the smallest model). 
On the other hand, performance collapses on D3 (Geometric Accuracy): even the leader, Gemini 2.5 Pro, achieves only 25.5\% Prec.@IoU=0.15, with Position Accuracy (PA@0.3m) dropping to 10.8\%. Notably, PA shows no correlation with overall ranking: Claude Sonnet 4.6 (rank 4 overall) scores only 4.9\%, below several lower-ranked models. 
This geometric deficiency is shared across models and persists even for those with high overall scores, indicating a limitation independent of general capability. We thus describe this as VLMs having \emph{describing eyes} but not \emph{measuring eyes}; prior analyses attribute such deficits in part to vision tokenizers that trade fine-grained spatial detail for semantic generalization~\cite{wang, liu}. This is a gap that existing QA-based metrics are unlikely to surface.

\noindent\textbf{Is Scaling All You Need?}
Stronger performance does not consistently correspond to larger parameter counts or newer models.
Within the Qwen3-VL family, the 8B model (55.1) outperforms both the 30B (50.8) and the 235B-A22B MoE variant (51.8), while InternVL3.5-30B-A3B fails to parse whereas its 8B counterpart succeeds. Gemma-4-31B-IT (59.7) ranks third overall, outperforming Claude Sonnet 4.6 and GPT-5.4 despite being substantially smaller than the proprietary models it outperforms. Newer models are not always better: GPT-4o (60.4) outperforms the newer GPT-5.4 (56.7), and Gemini 2.5 Pro, a 2025-generation model already superseded by newer releases, still leads the leaderboard (62.1) ahead of both Claude Sonnet 4.6 and GPT-5.4, both released in 2026. These inversions suggest that metric 3D understanding depends on factors beyond raw scale or recency. One likely contributor is architectural: current VLMs generate predictions as discrete token sequences, while the spatial signal is continuous; how to bridge this mismatch remains an open problem. Differences in training-data composition and structured-output supervision across model families could also play a role~\cite{liu}. Besides, our protocol's suppression of chain-of-thought reasoning is a further possible confound, which we discuss in Appendix~\ref{app_e2}.

\begin{wrapfigure}{rb}{0.5\textwidth}
\vspace{-.16in}
  \begin{centering}
\includegraphics[width=0.5\textwidth]{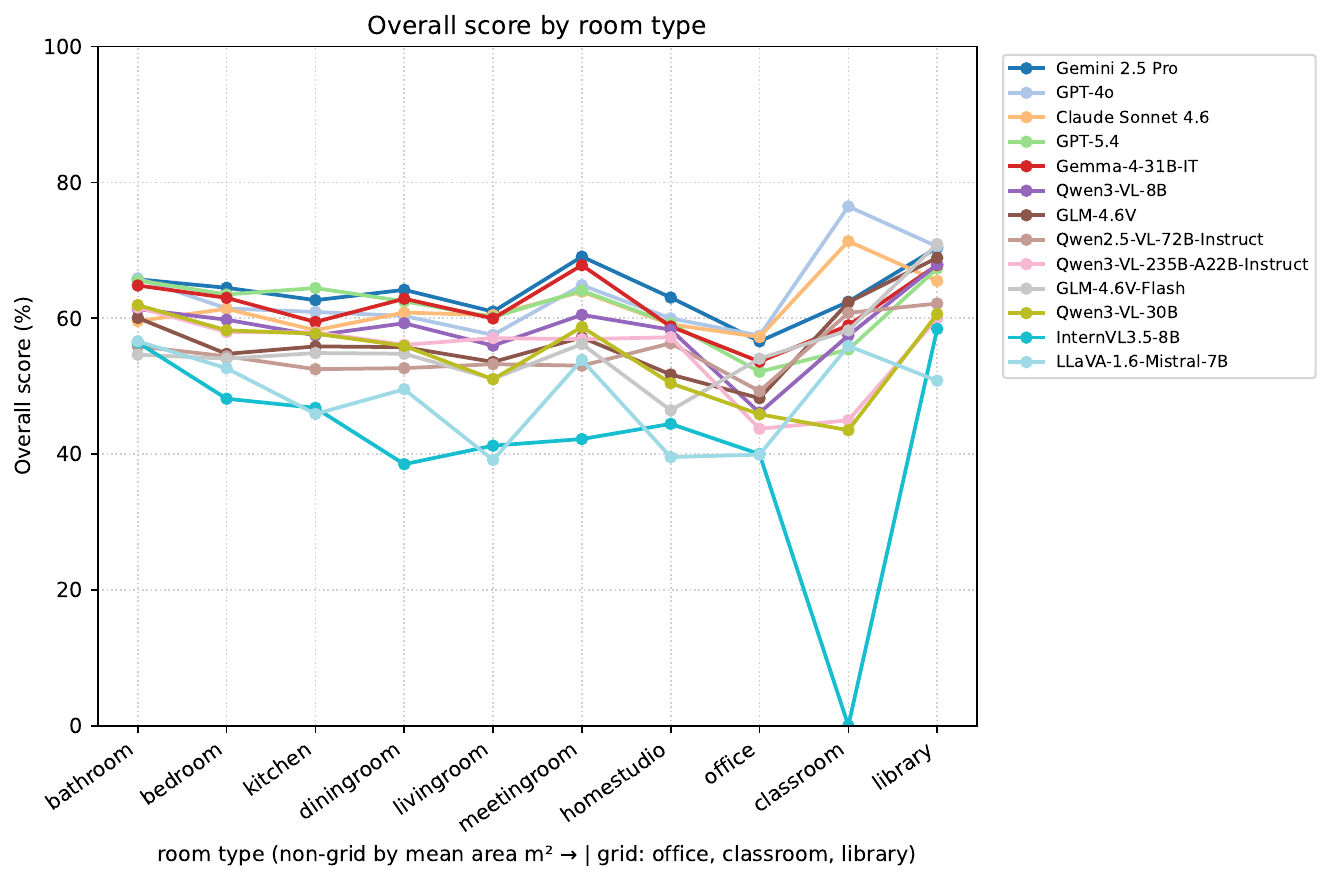}
  \end{centering}
  \vspace{-.15in}
  \caption{\small Overall score by room type.}
  \label{fig:roomtype}
  \vspace{-.05in}
\end{wrapfigure}

\noindent\textbf{The grid illusion: high scores, weak anchoring.}
We partition the 1,000 scenes into five density bins by in-view object count (Tab.~\ref{tab:complexity_d_overall}), and reveal that complexity does not degrade the overall score monotonically: most models drop from Bin 1 to Bin 5, but GPT-4o and Claude Sonnet 4.6 improve on dense scenes. The same pattern shows up by room type (Fig.~\ref{fig:roomtype}): scores spread out further apart in grid-structured rooms, especially classrooms, than in non-grid rooms. Both observations trace back to the same cause. Dense bins and grid room types largely overlap, since offices, classrooms, and libraries account for most of the high-object-count scenes, and these are exactly the room types governed by the D5 metric. Decomposing D5 into its two sub-metrics shows what the Overall score alone does not: GSR is high across most models, meaning they can generate internal regularity, but GCR stays below 30\% even for the strongest proprietary models. In other words, \emph{models reproduce the visual regularity without anchoring it to the correct location in the world frame} (Fig.~\ref{tab:grid_scenes}). Reading D5 therefore requires looking at GSR and GCR together rather than at the combined score. Section~\ref{subsec:perceptual} complements this numerical result with perceptual evidence, where reconstructed renderings allow the same misalignment to be assessed by direct visual comparison.

\input{tables/complexity}

%########################################################################################################
%########################################################################################################
%########################################################################################################

\subsection{Perceptual Results}
\label{subsec:perceptual}

For the perceptual protocol, we evaluate six representative models (four closed-source and two open-source). These models are selected primarily based on their numerical Overall scores. Full inference setup, prompts, parse-recovery pipeline, and judge configuration are in Appendix~\ref{app_b6}.

\input{tables/judge_vlm}

The perceptual ranking in Table~\ref{tab:perception_overall} largely
mirrors the numerical leaderboard: Gemini-2.5-Pro and GPT-5.4 lead
clearly, Gemma-4-31B-IT and Claude Sonnet~4.6 form a tight middle
tier, and Qwen2.5-VL-72B-Instruct sits at the bottom. Crucially, GPT-4o ranks high numerically but drops perceptually, suggesting its advantage relies on metrics favoring local plausibility over global coherence. Conversely, GPT-5.4 climbs from fifth numerically to second perceptually, indicating that its holistic layout quality is preferred by humans yet under-rewarded by automated metrics.
s.

The broader takeaway is that the 
two protocols converge: scores derived from object-level numerical
metrics and judgments derived from rendered reconstructions agree
on the ordering of model capability. This convergence matters
because the reconstruction gives a more direct
read on whether a predicted layout actually \emph{resembles} the
scene than any numerical metric can, and its agreement with the
numerical leaderboard supports the validity of both.

\begin{figure}[t!]
    \centering
    \small
    \includegraphics[width=1\linewidth]{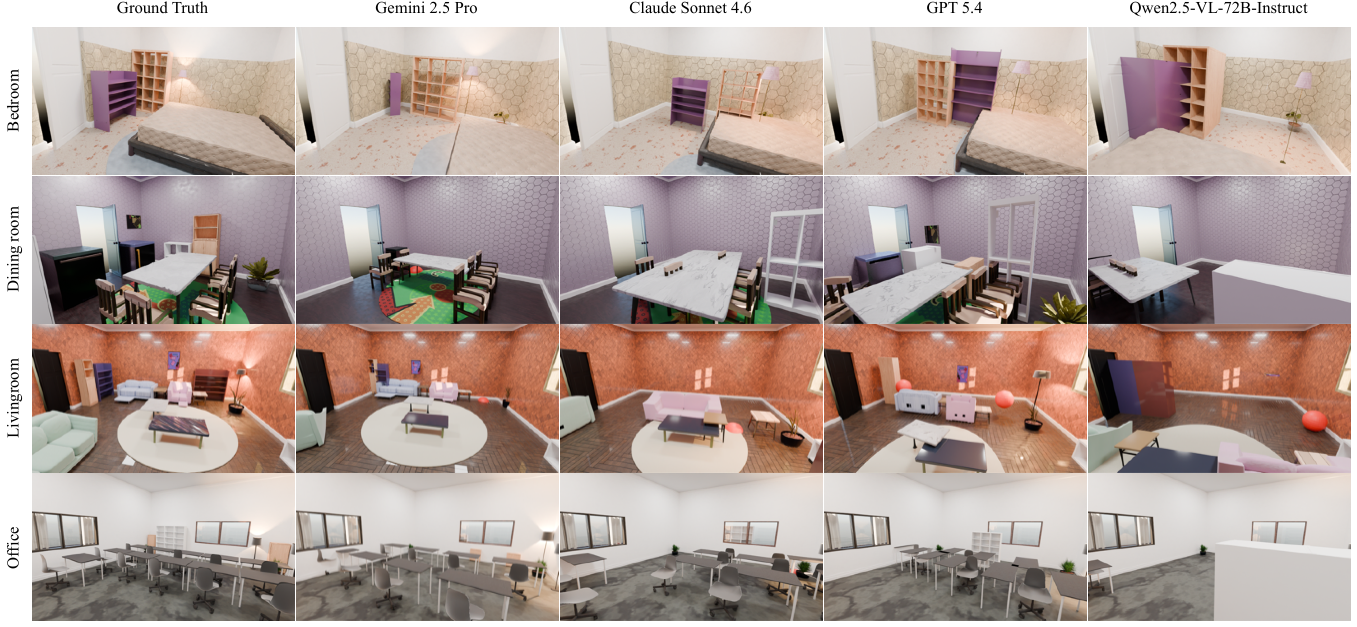}\\
    \caption{Scene reconstruction visualizations for four representative models (Gemini 2.5 Pro, Claude Sonnet 4.6, GPT-5.4, and Qwen2.5-VL-72B-Instruct), shown for four room types: bedroom, dining room, living room, and office.} 
    \label{fig:reconstruction}
    \vspace{-.15in}
\end{figure}

\noindent \textbf{Human Correlation Study.} To validate whether Judge VLM captures human perceptual preference, we conduct a human-correlation study under the same rubric and scene presentation protocol. Human ratings are aggregated as Mean Opinion Score (MOS), i.e., the average Likert score across annotators for each scene and perceptual dimension. We then compute \textbf{scene-wise} agreement by averaging model scores within each scene and measuring Spearman correlation between Human MOS and Judge-VLM scores across scenes. Judge VLM shows strong alignment with human perception: $\mathbf{\rho=0.79}$. These results support the \textbf{validity} of Judge VLM as a scalable perceptual evaluator. This validation matters because judging whether a reconstruction preserves the same spatial layout as the reference requires reasoning jointly about visual-level structure and semantic content, a combination that is difficult to capture with a single algorithmic metric. Using a VLM as judge provides a practical way to scale human-like evaluation across many models and scenes. Comprehensive details regarding the configuration of the Human Correlation Study are provided in the Appendix \ref{app_d4}.

%########################################################################################################
%########################################################################################################
%########################################################################################################

\subsection{Cross-Bench Analysis}
\label{sec:diagnostic}

We compare model rankings across IDEAL-Bench and five established spatial benchmarks; full per-benchmark details are in Appendix~\ref{app_d5}. Given the small per-benchmark intersection of jointly-evaluated models (n=3-4), this comparison should be read as suggestive. At the top, IDEAL-Bench agrees with prior benchmarks: Gemini-2.5-Pro and GPT-4o lead consistently. Where rankings diverge, IDEAL-Bench surfaces two blind spots of existing paradigms. Models strong on QA-style benchmarks drop on IDEAL-Bench, since local cues and language priors can substitute for global 3D representation in question answering but not in metric layout output~\cite{ma, zhang}. 
Models that perform well on simple geometric primitive reconstruction also degrade, suggesting that pose estimation for isolated shapes does not directly transfer to indoor layout inference, where objects are larger, asymmetric, and observed from unconstrained in-scene viewpoints.
\textbf{In short: Spatial QA tests whether a model \emph{sees} space; IDEAL-Bench tests whether it \emph{understands} the space.}

%% file: tables/mainresult.tex
% --- IDEAL main table (tabular): generated by benchmarks/analysis/main_table/main_table.py ---
% Requires: \usepackage{booktabs,multirow}
\begin{table}[t]
\centering
\caption{\textbf{IDEAL-Bench leaderboard.} 
Overall: Ovr.\ score in $[0,100]$ ($\uparrow$) is the unweighted mean of all 11 applicable rate metrics; Rank (↓) is the ranking according to Ovr. score across all models; Geo ($\uparrow$) is the unweighted mean of the five underlying D3 and D5 metrics (PA, Prec., RA, GCR, GSR), reflecting pure geometric reasoning. Other ratio metrics are in $[0,100]$ ($\uparrow$), full definitions of all metrics are in Appendix~\ref{app_b3} and ~\ref{app_b4}. $\times$~\textit{Task Failed}: see Appendix~\ref{app_d3} for failure analysis. Best per-metric value is in \textbf{bold}, second-best is \underline{underlined}.}
\vspace{.14in}
\resizebox{\textwidth}{!}{%
\small
\setlength{\tabcolsep}{4pt}
\renewcommand{\arraystretch}{0.95}
\begin{tabular}{lcccccccccccccc}
\toprule
\multirow{2}{*}{Model} & \multicolumn{3}{c}{Overall} & \multicolumn{2}{c}{D1: Scene Val.} & \multicolumn{2}{c}{D2: Phys. Plaus.} & \multicolumn{3}{c}{D3: Geo. Acc.} & \multicolumn{2}{c}{D4: Obj. Recog.} & \multicolumn{2}{c}{D5: Grid Layout} \\
\cmidrule(lr){2-4} \cmidrule(lr){5-6} \cmidrule(lr){7-8} \cmidrule(lr){9-11} \cmidrule(lr){12-13} \cmidrule(lr){14-15}
 & Ovr.$\uparrow$ & Rank$\downarrow$ & Geo.$\uparrow$ & RTA$\uparrow$ & PSR$\uparrow$ & NOR$\uparrow$ & IBR$\uparrow$ & PA$\uparrow$ & Prec.$\uparrow$ & RA$\uparrow$ & RR$\uparrow$ & NHR$\uparrow$ & GCR$\uparrow$ & GSR$\uparrow$ \\
\midrule
\multicolumn{15}{l}{\textit{Proprietary Models}} \\
Gemini 2.5 Pro & \textbf{62.1} & \textbf{1} & \textbf{40.1} & 87.5 & \textbf{100.0} & \underline{80.9} & 34.3 & \underline{10.8} & \textbf{25.5} & \textbf{37.7} & \textbf{89.9} & 90.3 & \textbf{29.1} & 97.3 \\
GPT-4o & \underline{60.4} & \underline{2} & \underline{35.9} & 88.5 & \textbf{100.0} & \textbf{84.6} & 46.1 & \textbf{11.5} & 23.1 & 20.0 & 72.8 & \underline{92.2} & 25.2 & \textbf{100.0} \\
Claude Sonnet 4.6 & 59.2 & 4 & 34.7 & 88.5 & \textbf{100.0} & 80.6 & 34.6 & 4.9 & 18.6 & 26.8 & \underline{85.4} & 88.1 & 27.8 & 95.7 \\
GPT-5.4 & 56.7 & 5 & 28.4 & \underline{89.3} & \textbf{100.0} & 65.0 & 55.3 & 8.5 & \underline{24.3} & \underline{33.7} & 82.2 & 89.9 & \underline{28.6} & 46.7 \\
\midrule
\multicolumn{15}{l}{\textit{Open-source Models}} \\
Gemma-4-31B-IT & 59.7 & 3 & 35.7 & 84.4 & \underline{99.9} & 78.3 & 40.9 & 9.8 & 24.1 & 20.8 & 81.8 & \textbf{92.6} & 24.1 & \underline{99.9} \\
Qwen3-VL-8B-Instruct & 55.1 & 6 & 28.6 & 84.1 & \textbf{100.0} & 62.2 & 52.5 & 8.2 & 18.5 & 19.6 & 73.0 & 91.1 & 14.4 & 82.2 \\
GLM-4.6V & 54.4 & 7 & 30.7 & 87.0 & \textbf{100.0} & 60.9 & 29.6 & 4.4 & 10.1 & 26.4 & 76.8 & 91.2 & 13.3 & 99.3 \\
Qwen2.5-VL-72B-Instruct & 52.6 & 8 & 27.9 & 85.4 & \textbf{100.0} & 64.8 & 26.7 & 2.1 & 7.7 & 18.5 & 70.6 & 91.4 & 11.9 & 99.3 \\
Qwen3-VL-235B-A22B-Instruct & 51.8 & 9 & 24.3 & \textbf{90.3} & 76.0 & 58.5 & \underline{56.1} & 7.2 & 21.2 & 22.8 & 76.7 & 90.4 & 16.1 & 54.3 \\
GLM-4.6V-Flash & 51.5 & 10 & 24.0 & 86.2 & 92.5 & 53.6 & 49.0 & 4.7 & 11.8 & 19.2 & 73.7 & 91.2 & 7.5 & 76.9 \\
Qwen3-VL-30B-A3B-Instruct & 50.8 & 11 & 22.6 & 84.0 & 92.0 & 56.8 & 44.2 & 7.5 & 14.6 & 21.6 & 79.9 & 89.2 & 22.9 & 46.5 \\
InternVL3.5-8B & 45.9 & 12 & 27.0 & 80.8 & 43.7 & 66.9 & 23.0 & 2.6 & 4.9 & 18.4 & 72.0 & 84.2 & 8.9 & \textbf{100.0} \\
LLaVA-1.6-Mistral-7B & 44.5 & 13 & 17.5 & 64.8 & 98.1 & 16.3 & \textbf{70.6} & 2.4 & 4.6 & 19.2 & 69.2 & 82.8 & 8.0 & 53.3 \\
InternVL3.5-30B-A3B & \multicolumn{14}{c}{$\times$~\textit{Task Failed}} \\
Janus-Pro-7B & \multicolumn{14}{c}{$\times$~\textit{Task Failed}} \\
\bottomrule
\end{tabular}%
}
\vspace{-.15in}
\label{tab:ideal_leaderboard}
\end{table}

%% file: tables/complexity.tex
% IDEAL complexity — preamble: \usepackage{booktabs,graphicx}
\begin{table}[t!]
  \centering
  \small
  \caption{\textbf{Performance breakdown by scene complexity.} We report the mean overall score (Ov.) and Parse-Success Rate (PSR, \% of validly parsed layouts) across five density bins defined by in-view object counts (see Appendix~\ref{app_density} for bin boundaries). Italicized values denote total scenes per bin. Ov. is reported as N/A when PSR = 0\%.}
  \vspace{.07in}
  \label{tab:complexity_d_overall}
  \small
  \setlength{\tabcolsep}{2pt}
  \renewcommand{\arraystretch}{1}
  \resizebox{\linewidth}{!}{%
  \begin{tabular}{lcccccccccc}
    \toprule
    \textbf{Model} & \multicolumn{2}{c}{\shortstack[c]{\textbf{Bin 1: 1--6}\\\scriptsize \textit{298\,scenes}}} & \multicolumn{2}{c}{\shortstack[c]{\textbf{Bin 2: 7--9}\\\scriptsize \textit{231\,scenes}}} & \multicolumn{2}{c}{\shortstack[c]{\textbf{Bin 3: 10--14}\\\scriptsize \textit{253\,scenes}}} & \multicolumn{2}{c}{\shortstack[c]{\textbf{Bin 4: 15--24}\\\scriptsize \textit{122\,scenes}}} & \multicolumn{2}{c}{\shortstack[c]{\textbf{Bin 5: 25--49}\\\scriptsize \textit{96\,scenes}}} \\
    % \cmidrule(lr){2-3} \cmidrule(lr){4-5} \cmidrule(lr){6-7} \cmidrule(lr){8-9} \cmidrule(lr){10-11} \\
    & Ov. & PSR & Ov. & PSR & Ov. & PSR & Ov. & PSR & Ov. & PSR \\
    \midrule
    Gemini 2.5 Pro & 53.3\% & 100.0\% & \textbf{55.3\%} & 100.0\% & \textbf{54.5\%} & 100.0\% & 48.5\% & 100.0\% & 38.8\% & 100.0\% \\
    GPT-4o & 52.9\% & 100.0\% & 51.0\% & 100.0\% & 50.4\% & 100.0\% & \textbf{53.7\%} & 100.0\% & \textbf{61.1\%} & 100.0\% \\
    Claude Sonnet 4.6 & 46.6\% & 100.0\% & 48.1\% & 100.0\% & 50.7\% & 100.0\% & 49.6\% & 100.0\% & 54.5\% & 100.0\% \\
    GPT-5.4 & \textbf{54.6\%} & 100.0\% & 52.2\% & 100.0\% & 50.0\% & 100.0\% & 41.6\% & 100.0\% & 32.3\% & 100.0\% \\
    Gemma-4-31B-IT & 51.9\% & 100.0\% & 52.4\% & 100.0\% & 51.2\% & 99.6\% & 46.5\% & 100.0\% & 42.0\% & 100.0\% \\
    Qwen3-VL-8B & 48.8\% & 100.0\% & 48.3\% & 100.0\% & 47.6\% & 100.0\% & 43.1\% & 100.0\% & 31.9\% & 100.0\% \\
    GLM-4.6V & 45.4\% & 100.0\% & 42.7\% & 100.0\% & 42.7\% & 100.0\% & 40.8\% & 100.0\% & 34.6\% & 100.0\% \\
    Qwen2.5-VL-72B-Instruct & 40.9\% & 100.0\% & 40.2\% & 100.0\% & 41.9\% & 100.0\% & 43.5\% & 100.0\% & 39.6\% & 100.0\% \\
    Qwen3-VL-235B-A22B-Instruct & 49.7\% & 73.5\% & 47.2\% & 73.2\% & 46.3\% & 76.3\% & 36.7\% & 75.4\% & 16.9\% & 90.6\% \\
    GLM-4.6V-Flash & 42.2\% & 93.0\% & 42.9\% & 90.9\% & 44.5\% & 94.5\% & 41.2\% & 92.6\% & 34.9\% & 89.6\% \\
    Qwen3-VL-30B & 47.2\% & 98.0\% & 44.5\% & 93.1\% & 44.3\% & 94.1\% & 38.1\% & 97.5\% & 24.4\% & 58.3\% \\
    InternVL3.5-8B & 40.3\% & 98.0\% & 36.9\% & 55.8\% & 38.4\% & 5.1\% & 39.7\% & 2.5\% & N/A & 0.0\% \\
    LLaVA-1.6-Mistral-7B & 39.7\% & 100.0\% & 34.7\% & 98.7\% & 35.0\% & 98.0\% & 32.9\% & 93.4\% & 33.9\% & 96.9\% \\
    \bottomrule
  \end{tabular}%
  }
  \vspace{-.15in}
\end{table}

%% file: tables/judge_vlm.tex
\begin{table}[t]
\centering
\footnotesize
\caption{We perform perceptual evaluation among the 6 representative models, reporting perceptual score, model rank, and top-1 score.}
\vspace{.07in}
\label{tab:perception_overall}
\begin{tabular}{lccc}
\toprule
Model & Mean Perceptual $\uparrow$ & Mean Rank $\downarrow$ & Top-1 (\%)\\
\midrule
Gemini-2.5-pro & \textbf{3.56} & \textbf{2.40} & \textbf{41.2} \\
GPT-5.4 & 3.49 & 2.45 & 23.5 \\
Gemma-4-31B-it & 3.22 & 3.26 & 11.8 \\
Claude-Sonnet-4-6 & 3.06 & 3.39 & 17.6 \\
GPT-4o & 2.78 & 4.15 & 2.9 \\
Qwen2.5-VL-72b-instruct & 1.92 & 5.39 & 2.9 \\
\bottomrule
\end{tabular}
\end{table}

%% file: sections/7concl.tex
\section{Conclusion}
\label{sec:conclusion}
We introduced IDEAL-Bench, a benchmark for holistic 3D layout inference from single images. 
By requiring models to produce structured predictions that are both numerically evaluable and
visually re-renderable, IDEAL-Bench reveals geometric reasoning failures that QA-based evaluations cannot surface. 
Evaluating 15 VLMs shows that the task remains substantially unsolved (best
62.1/100), that geometric regression is a universal bottleneck decoupled from object recognition, and that model rankings diverge from those on QA and primitive-reconstruction benchmarks. 
These findings position IDEAL-Bench as a necessary complement to existing
spatial evaluation suites.

%\noindent \textbf{Limitations and Scope.} IDEAL-Bench is scoped for diagnostic clarity at the current VLM capability level: single-room rectangular layouts, canonical orientations, and a closed in-view category list to isolate geometric reasoning from open-vocabulary detection. The released procedural pipeline supports expansion along any of these axes; See Appendix~\ref{app_e} for scoping choices.

% \noindent \textbf{Limitations and Scope.} IDEAL-Bench is intentionally 
% scoped to maximize diagnostic clarity at the current capability level of 
% VLMs, and several classes of simplifications should be made explicit. 
% (i) Scene geometry is restricted to single-room, axis-aligned rectangular 
% layouts; (ii) object yaw is mainly discretized to the four cardinal 
% orientations, reflecting predominantly wall-parallel placement; 
% (iii) only layout-determining large furniture is included, omitting 
% small objects and clutter; (iv) vertical position $z$ for floor-standing 
% objects uses ground truth to avoid conflating asset-pivot conventions 
% with prediction error; (v) the in-view category list is provided in 
% the prompt, isolating geometric reasoning from open-vocabulary detection. 
% The 1{,}000-scene scale (100 per room type) provides stable per-type 
% estimates while remaining tractable for the reconstruct-and-render 
% pathway; the released procedural pipeline scales linearly. Appendix 
% \ref{app_e} expands on the rationale behind each choice.

%% file: sections/8discussion.tex
\section{Discussion}
\label{sec:discussion}

\noindent \textbf{Limitations and Scope.} IDEAL-Bench is intentionally scoped to maximize diagnostic clarity at the current capability level of VLMs, and several classes of simplifications should be made explicit. Scenes are restricted to single-room rectangular layouts, with the orientation of most objects discretized to four cardinal directions. The in-view category list is provided in the prompt, isolating geometric reasoning from open-vocabulary detection. The coordinate alignment can introduce discrepancy for scenes where the camera faces close to a room diagonal. The 1,000-scene dataset is limited in scale but supports expansion along any of these axes; see Appendix~\ref{app_e1} for the reasoning behind each scope choice.

\noindent \textbf{Future Work.} The most direct extensions follow the axes this benchmark currently fixes: real-captured indoor scenes, multi-room and irregular floor plans, explicit lighting variability, and small objects or clutter with unrestricted rotation. Another direction is to vary scene difficulty one axis at a time within a fixed room category, for example by sweeping lighting or occlusion levels across matched layouts, to see whether model performance degrades gradually or fails sharply past some threshold. Our finding that geometric reasoning is a bottleneck across all 15 evaluated VLMs is based on comparing models that differ in scale, training recipe, and architecture all at once. To isolate which of these factors actually drives the bottleneck, future work would need to fix the prompt and parsing strategy and evaluate a single model family across scales. See Appendix~\ref{app_e2} for further discussion.

%% file: sections/AppA_dataset.tex
% =====================================================================
% Appendix A. Dataset Details
% =====================================================================

\section{Dataset Details}
\label{app_a}

% =====================================================================
% A1
% =====================================================================
\subsection{Scene Generation}
\label{app_a1}
IDEAL-Scenes is built on InfiniGen Indoors~\cite{infinigen}, a Blender-based procedural scene-generation engine that combines a procedural asset library with a constraint-based arrangement solver. We synthesized the scenes with 8 * RTX6000 GPUs, which took about one week to complete. The open-sourced models are all deployed on the same GPU devices.
We retain its core generation and constraint-solving framework, but introduce targeted modifications that make the resulting scenes suitable as a controlled benchmark for single-image layout estimation. InfiniGen's default configuration targets maximal scene diversity: multi-room floor plans, wide variance in lighting and furniture density, occasional degenerate elongated geometries, and arbitrary mixing of small decorative items. This diversity suits data augmentation but inflates inter-sample variance in ways that confound model evaluation.
Our modifications fall into three categories: \textit{room geometry}, \textit{global scene controls}, and \textit{per-room-type specification}.

\paragraph{Room geometry.}
\label{para:room_geometry}
Each scene is restricted to a single rectangular room, with multi-room floor plans and unrelated generation sub-stages disabled. Wall thickness is fixed at $0.25\,\text{m}$ across all room types, while wall heights are drawn per scene from discrete per-tier ranges (see Table~\ref{tab:room_specs}). Floor-plan dimensions are likewise drawn from per-room-type discrete
size tiers, eliminating extreme aspect ratios that occur under unconstrained rectangular sampling.

\paragraph{Global scene controls.}
All scenes share fixed rendering and lighting parameters: a uniform interior light-energy multiplier of $5.0$. Cycles denoising is explicitly enabled, with a Cycles sample count of 128 per pixel, and an output resolution of $1280\times720$.
Each scene is captured by a single pinhole camera with a fixed $15\,\text{mm}$ focal length on a $32\,\text{mm}$ sensor, yielding a wide horizontal field of view that fits a full room layout into a single frame. The camera is placed inside the room at a height sampled uniformly from $1.2$--$2.0\,\text{m}$ above the floor and oriented toward the room's geometric center; 
per-scene camera, lighting, environment, and image-quality statistics are recorded in \texttt{metadata.json}.

\paragraph{Per-room-type specification.} IDEAL-Scenes covers 10 room types drawn from two constraint modules: \textit{Living Space} (bedroom, homestudio, livingroom, kitchen, bathroom, diningroom) and \textit{Working Space} (office, meetingroom, classroom, library). For each room type we override some aspects of generation, summarized in Table~\ref{tab:room_specs}.

\textit{Semantic asset whitelist.} For each room type, we restrict the set of factory categories eligible for placement to a curated whitelist of layout-relevant furniture, removing small decorative items and disallowing categories incompatible with the room's semantics. This produces scenes whose object inventories match common-sense expectations and ensures that ground-truth layouts contain only objects large enough to be reliably evaluated.

\input{tables/roomtype_statistics}

\paragraph{Scene bundles.}
Each of the $1{,}000$ published scenes is stored as a self-contained folder named \texttt{\{room\_type\}\_s\{seed\}} containing the following files:
\input{tables/scene_bundle}
The first five files are produced by the generation pipeline; \texttt{objects\_in\_view.json}, the index of in-view objects, is appended by Stage~2 preprocessing (Appendix~\ref{app_a3}).
\texttt{metadata.json} contains additional information such as camera parameters, environment metadata, and image-quality statistics for downstream computation and analysis; the embedded mapping from \texttt{instance\_segmentation.png} pass indices to asset names is crucial for filtering in-view objects.
All spatial fields in \texttt{GT.json} and \texttt{metadata.json} follow the coordinate convention defined in Appendix~\ref{app_a2}.

% =====================================================================
% A2
% =====================================================================

\subsection{Geometric Conventions}
\label{app_a2}

All scenes share a unified geometric convention applied consistently across generation, preprocessing, evaluation and reconstruction.

\paragraph{World coordinate frame.}
We adopt a Z-up right-handed coordinate system: $+X$ points right, $+Y$ points forward (into the scene), and $+Z$ points up. The room floor geometric center is the world origin, with $Z=0$ defined as ground level. Units are metres, with precision to 0.01\,m.

\subsubsection{IDEAL-Scenes object representation.}
Each object in \texttt{GT.json} is described by three spatial fields: \texttt{position}, \texttt{dimensions}, and \texttt{rotation\_yaw}, together with an \texttt{id} and a \texttt{category} string. Their semantics match the predicted 
fields $(\hat{p}_i, \hat{d}_i, \hat{\theta}_i)$ defined in Section~\ref{sec:task}. 
Two details specific to GT extraction:

First, \texttt{position} is the center of the asset's tight 3D bounding box, 
computed independently of where the asset's Blender origin sits inside its mesh; 
this independence is why Appendix~\ref{app_a2_2} needs a separate procedure to 
locate the center rather than reading it off directly. Second, although 
$[W,D,H]$ are defined on the asset's own canonical axes as in Section~\ref{sec:task}, 
the mapping from these canonical axes to Blender's local axes is not fixed across 
asset factories; at canonical orientation 
(\texttt{rotation\_yaw}~$=0$) $W$ runs along world $+X$ and $D$ along world $+Y$. 
Categories without a meaningful front direction (e.g.\ \texttt{floor\_lamp}, 
\texttt{plant}, \texttt{rug}) are exported with \texttt{rotation\_yaw}~$=$~\texttt{null}, 
and downstream metrics skip yaw evaluation for these categories.

\subsubsection{From Blender asset to IDEAL-Scenes representation.}
\label{app_a2_2}
Mapping an InfiniGen-generated Blender asset to the representation above is non-trivial, for three reasons that the constraint solver is unaware of: (i) an asset's Blender origin can be placed inside (not necessarily geometric center of AABB) or outside its mesh by the asset's procedural generator, so it is not a reliable proxy for the geometric center; (ii) the asset's local axes are not standardized across factories: most assets have their semantic ``front'' along local $+X$, some along $+Y$; (iii) the asset's $[W, D, H]$ semantics depend on which way the asset is facing, so the mapping from Blender axes to canonical $[W, D, H]$ depends on a per-asset semantic forward direction. We resolve all three while extracting \texttt{GT.json}.

\paragraph{Position via root-local AABB center.}
For each asset, let $o$ denote its Blender root Object and $M^w_o = \texttt{o.matrix\_world}$ the corresponding Blender world matrix, set by InfiniGen's procedural generator. The IDEAL-Scenes position is defined as the geometric center of the asset's tight axis-aligned bounding box, computed in the root's local frame and then mapped to world coordinates:
\begin{equation}
\mathbf{p} = M^w_o \cdot \tfrac{1}{2}(\mathbf{c}_{\text{lo}} + \mathbf{c}_{\text{hi}}),
\end{equation}
where $\mathbf{c}_{\text{lo}}, \mathbf{c}_{\text{hi}} \in \mathbb{R}^3$ are the component-wise minimum and maximum of all evaluated mesh vertices in the asset hierarchy, expressed in the root's local frame.
That is, $\mathbf{c}_{\text{lo}} = (\min x, \min y, \min z)$ and $\mathbf{c}_{\text{hi}} = (\max x, \max y, \max z)$ are the two opposite corners of the root-local AABB.
The result is rounded to $0.01\,\text{m}$ before being written to \texttt{GT.json}.

This definition is independent of whether the Blender origin sits inside (or outside) the asset. The non-trivial offset between Blender origin and IDEAL center is therefore needed at evaluation and reconstruction as well, when a predicted IDEAL position must be converted back to a Blender origin for re-rendering.

Computing the AABB in the root-local frame, rather than directly in world space, also makes the result invariant to the asset's rotation: a yawed asset and the same asset at yaw $= 0$ yield identical $[\mathbf{c}_{\text{lo}}, \mathbf{c}_{\text{hi}}]$, and their world centers differ only through $M^w_o$.

\paragraph{Semantic forward direction.} The asset's canonical front direction is recovered from face-level semantic tags inherited from InfiniGen, which mark the mesh faces forming the asset's front surface.
When such tags are present, the forward vector $\mathbf{f}$ is the sum of the world-space normals of all tagged faces, projected onto the floor plane; when they are absent, we fall back to a per-category default.
The exported \texttt{rotation\_yaw} is the world angle of $\mathbf{f}$ measured counter-clockwise from world $+Y$, snapped to an integer degree, with a fixed correction applied to a small family of dining-style tables whose local axis convention differs from the rest of the asset library.

\paragraph{Canonical $[W, D, H]$.}
The bounding-box extents $(\Delta_x, \Delta_y, \Delta_z)$ from the root-local AABB are mapped to canonical $[W, D, H]$ by inspecting the semantic forward direction in the asset's local frame. If the local forward vector lies primarily along the asset's local $X$ axis, $W = \Delta_y$ and $D = \Delta_x$ (the asset's depth is along its forward axis, so it occupies local $X$); otherwise $W = \Delta_x$ and $D = \Delta_y$.
$H = \Delta_z$ in both cases.
For yaw-exempt categories, where there is no canonical front, dimensions are reported directly as the world-axis AABB extents instead.
Square side tables (where $|\Delta_x - \Delta_y| < 0.05\,\text{m}$) are forced to yaw $= 0$ to avoid arbitrary axis assignments.

% =====================================================================
% A3
% =====================================================================

\subsection{Preprocessing Pipeline}
\label{app_a3}

After generation, each scene passes through a two-stage preprocessing pipeline before entering the dataset. Stage 1 applies scene-level validity checks and discards scenes that fail any criterion; Stage 2 determines, for retained scenes, the subset of objects considered evaluation-eligible.

\paragraph{Stage 1: scene-level quality.}
A scene is retained if and only if all of the following hold.

\textit{(C1) Data completeness}: the full five-file bundle is present.

\textit{(C2) Non-empty field of view}: at least one furniture instance must 
(i)~occupy at least $300$ pixels in the instance segmentation map (the same threshold is reused at object level in Stage~2) and (ii)~be resolvable to a matching ground-truth object id via the metadata-to-GT alignment.

\textit{(C3) No furniture interpenetration}: overlap is checked via a broad-phase 
world-space AABB test (conservative, since it over-approximates yawed assets) 
followed by a narrow-phase Blender Boolean (\textsc{exact} solver) on surviving 
pairs, falling back to voxel- or proximity-based tests on mesh failures. A pair 
counts as overlapping when its estimated intersection volume exceeds 
$10^{-6}\,\text{m}^3$; any scene with such a pair is rejected. Bed frames and 
their mattresses are treated as a single unit and excluded from this check.

\textit{(C4) Within-room geometry}: scenes containing furniture whose 
ground-truth bounding box extends outside the room footprint are discarded 
\textit{(occasional degenerate procedural meshes produce parts extending 
beyond the walls despite the solver's placement constraints)}.

The released dataset is overlap-free and within-room by construction; in 
practice C3 dominates rejections, with C1, C2, and C4 each accounting for at 
most a handful of scenes per batch.

%--------------------------------------------------------------------------------

\paragraph{Stage 2: in-view object determination.} We define evaluation-eligible 
objects via three filters of increasing cost; an object is included in 
\texttt{objects\_in\_view.json} only if it passes all of F1--F3.

\textit{(F1) Visibility.} The instance's pixel count in the segmentation map is 
at least $300$.

\textit{(F2) Aspect ratio.} The instance is rejected if its segmentation 
bounding box has
\begin{equation}
\textsc{ratio} = \max\!\left(\frac{\Delta x}{\Delta y}, \frac{\Delta y}{\Delta x}\right) > 12,
\end{equation}
which catches pathological perspective distortion (e.g.\ a bookshelf at extreme 
grazing incidence).

\textit{(F3) Truncation.} An object whose visible footprint is misleadingly 
small due to image-boundary clipping is rejected iff all three sub-conditions 
hold jointly: (a) the pixel-mass centroid lies within the outermost $12.5\%$ of 
the image width or height \textit{(using the centroid rather than the bbox 
corner, since large in-frame objects such as rugs or beds often touch the 
border while their mass stays central)}; (b) the truncation ratio
\begin{equation}
\texttt{truncation\_ratio} = \frac{\texttt{actual\_pixels}}{\texttt{projected\_pixels}},
\end{equation}
where \texttt{actual\_pixels} is the segmented pixel count and 
\texttt{projected\_pixels} is the unclipped silhouette area from rasterizing 
every mesh triangle without clipping; and (c) visible pixels are below $3\%$ of 
the image area, which acts as a rescue clause overriding (a)--(b) for objects 
very close to the camera, whose inflated \texttt{projected\_pixels} would 
otherwise spuriously trigger (b). For bed instances, \texttt{actual\_pixels} and 
\texttt{projected\_pixels} are computed jointly over \texttt{SingleBedFactory} 
and \texttt{MattressFactory} as a single semantic unit.

% =====================================================================
% A4
% =====================================================================

\subsection{Dataset Statistics}
\label{app_a4}

\begin{figure}[t!]
    \centering
    \includegraphics[width=\linewidth]{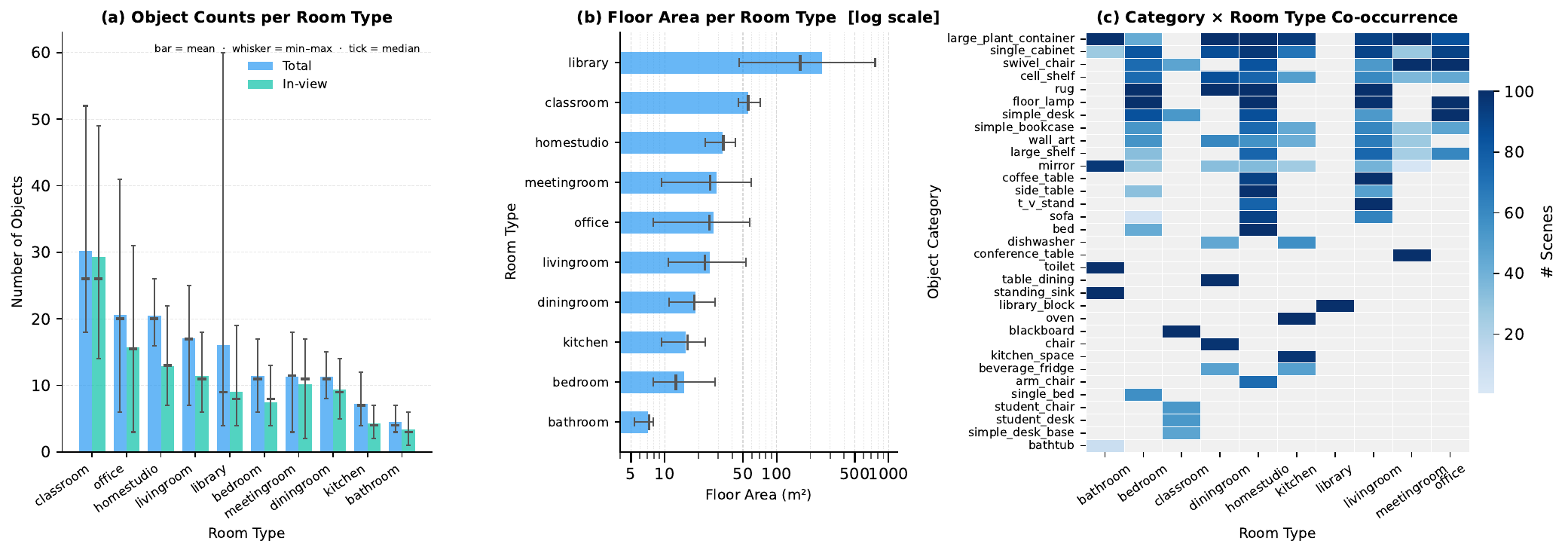}
    \vspace{-.2in}
    \caption{
    \textbf{IDEAL-Scenes statistics.}
    \textbf{(a)} Total and in-view object counts per room type
    (bar = mean, whiskers = min--max, tick = median); classrooms are the 
    densest in-view ($29.3$ avg.) and bathrooms the sparsest ($3.4$), with the 
    largest total–in-view gap in homestudio and library, where peripheral 
    objects fall outside the camera view.
    \textbf{(b)} Floor area per room type, log scale, from 
    $\sim\!6\,\text{m}^2$ (bathroom) to a median of $162\,\text{m}^2$ for 
    libraries (up to $\sim\!760\,\text{m}^2$).
    \textbf{(c)} Category--room-type co-occurrence.
}
    \label{fig:dataset_stats}
    \vspace{-.2in}
\end{figure}

We report dataset-level statistics on the released $1{,}000$ scenes to verify 
that the design choices in Appendices~\ref{app_a1}--\ref{app_a3} produce a 
balanced and well-conditioned benchmark. Figure~\ref{fig:dataset_stats} 
summarizes object counts per room type, floor-area distributions, and 
category--room-type co-occurrence.

\begin{wrapfigure}{r}{0.42\textwidth}
\vspace{-.1in}
  \begin{centering}
\includegraphics[width=0.42\textwidth]{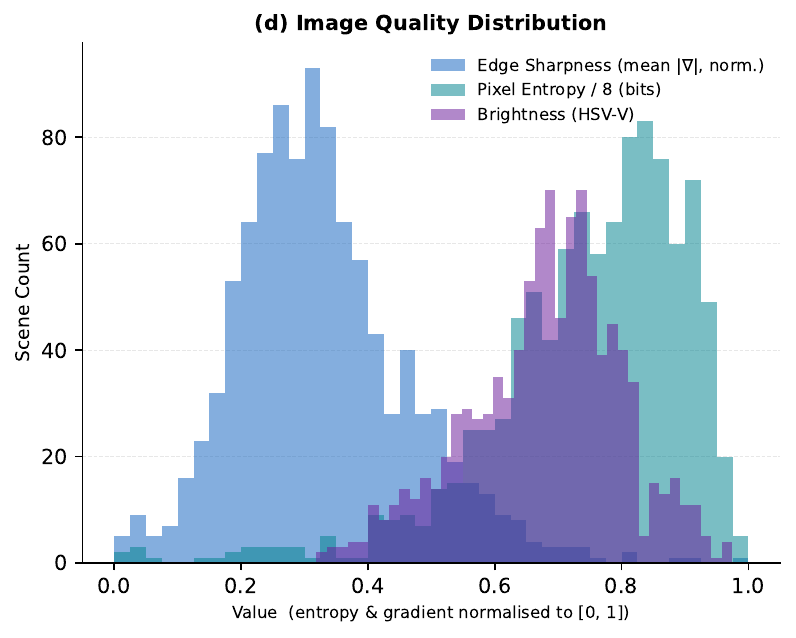}
  \end{centering}
  \vspace{-.15in}
  \caption{\small Image quality distribution across all scenes.}
  \label{fig:app_imgquality}
  \vspace{-.05in}
\end{wrapfigure}

\paragraph{Object counts.}
The mean number of furniture instances per scene is $15.0$ in the full scene 
graph and $11.3$ after in-view filtering. Counts vary substantially across room 
types (Figure~\ref{fig:dataset_stats}a): classrooms have the highest mean 
in-view count ($29.3\pm9.2$), bathrooms the fewest ($3.4\pm1.2$). The gap 
between full-graph and in-view counts is largest for homestudio and library, 
where the F3 truncation filter removes peripheral and wall-adjacent objects.

\paragraph{Floor area.}
Floor areas span over two orders of magnitude across room types 
(Figure~\ref{fig:dataset_stats}b, log scale), from $5$--$8\,\text{m}^2$ for 
bathrooms to a median of $162\,\text{m}^2$ for libraries (up to 
$\sim\!760\,\text{m}^2$); the wide library range is intentionally retained to 
stress-test model performance at extreme scene scale. Area variation within 
each type reflects the discrete size tiers of Appendix~\ref{app_a1}.

\paragraph{Object category diversity.}
Figure~\ref{fig:dataset_stats}c shows category-room-type co-occurrence: 
room-specific categories (\texttt{blackboard}, \texttt{conference\_table}, 
\texttt{toilet}, \texttt{library\_block}) appear exclusively in their 
designated room types, while general categories (\texttt{large\_plant\_container}, 
\texttt{single\_cabinet}, \texttt{swivel\_chair}, \texttt{cell\_shelf}) recur 
across multiple types.

\paragraph{Image quality.}

Per-scene entropy, edge sharpness, and brightness (HSV-V) are recorded in 
\texttt{metadata.json}; Figure~\ref{fig:app_imgquality} shows their distribution 
across all $1{,}000$ scenes. Entropy and sharpness are unimodal and concentrated 
in the mid-to-high range, consistent with the fixed rendering parameters. 
Brightness is bimodal, with a secondary peak near $0.85$ from scenes whose 
larger window apertures admit more ambient HDRI light.

%% file: tables/roomtype_statistics.tex
\begin{table}
\caption{Per-room-type specifications. Floor-plan ranges give \textit{width}
$\times$ \textit{depth} in metres for each tier. Wall heights are sampled
uniformly within the listed range per tier. Sampling column gives the per-scene
tier sampling rule. Strategy: S\,=\,constraint solver, P\,=\,procedural grid,
P\,+\,S\,=\,grid then solver for secondary items.}
\label{tab:room_specs}
\centering
\small
\setlength{\tabcolsep}{4pt}
\begin{tabular}{lllllp{4.4cm}}
\toprule
\textbf{Room type} & \textbf{Tier} & \textbf{Floor plan (m)} & \textbf{Wall ht (m)} & \textbf{Strategy} & \textbf{Representative objects} \\
\midrule
\multicolumn{6}{l}{\textit{Living Space (residential)} --- bedroom and livingroom: $\{\textsc{small}, \textsc{medium}\}$ uniform per seed; others fixed.} \\
Bedroom        & sm. & 3.0--3.9 $\times$ 3.0--3.9 & 2.8--3.2 & S & Single bed, nightstand, storage \\
Bedroom        & md. & 4.0--5.5 $\times$ 4.0--5.5 & 3.0--4.0 & S & Double bed, nightstand, desk, sofa \\
Homestudio     & --- & 5.0--7.0 $\times$ 5.0--7.0 & 4.0--6.0 & S & Double bed, wall sofa, TV stand, floor lamp \\
Livingroom     & sm. & 3.0--5.0 $\times$ 3.0--5.0 & 2.8--3.2 & S & Sofa, coffee table, TV stand, rug \\
Livingroom     & md. & 4.0--8.0 $\times$ 4.0--8.0 & 3.0--4.0 & S & Sofa set, coffee table, TV stand, storage \\
Kitchen        & --- & 3.2--5.0 $\times$ 3.2--5.0 & 3.0--4.0 & S & Kitchen counter, island, oven, sink \\
Diningroom     & --- & 3.5--5.5 $\times$ 3.5--5.5 & 3.0--4.0 & S & Dining table, chairs, storage, rug \\
Bathroom       & --- & 2.5--3.2 $\times$ 2.5--3.2 & 3.0--4.0 & S & Toilet, sink, mirror, bathtub \\
\midrule
\multicolumn{6}{l}{\textit{Working Space (office \& educational)} --- all four types: $\{\textsc{small}, \textsc{medium}, \textsc{large}\}$ uniform per seed.} \\
Office         & sm. & 3.0--4.0 $\times$ 3.0--4.0  & 2.8--3.2 & P\,+\,S & Desks, swivel chairs, storage, plants \\
Office         & md. & 4.0--6.0 $\times$ 4.0--6.0  & 3.0--4.0 & P\,+\,S & Desks, swivel chairs, storage, plants \\
Office         & lg. & 6.0--8.0 $\times$ 6.0--8.0  & 4.0--6.0 & P\,+\,S & Desks, swivel chairs, storage, plants \\
Meetingroom    & sm. & 3.5--5.0 $\times$ 3.0--4.5  & 2.8--3.2 & S & Conference table, chairs, storage \\
Meetingroom    & md. & 5.0--7.0 $\times$ 4.0--5.5  & 3.0--4.0 & S & Conference table, chairs, storage \\
Meetingroom    & lg. & 7.0--10.0 $\times$ 5.5--7.0 & 4.0--6.0 & S & Conference table, chairs, storage \\
Classroom      & sm. & 6.0--6.5 $\times$ 8.0--8.5  & 2.8--3.2 & P & Student desks, chairs, blackboard \\
Classroom      & md. & 6.0--7.0 $\times$ 8.5--9.5  & 3.0--4.0 & P & Student desks, chairs, blackboard \\
Classroom      & lg. & 6.5--7.5 $\times$ 9.0--10.0 & 4.0--6.0 & P & Student desks, chairs, blackboard \\
Library        & sm. & 7.5--10.0 $\times$ 6.0--9.5   & 2.8--3.2 & P & Double-faced bookshelves (grid) \\
Library        & md. & 10.0--15.0 $\times$ 9.5--16.0 & 3.0--4.0 & P & Double-faced bookshelves (grid) \\
Library        & lg. & 16.0--30.0 $\times$ 18.0--36.0 & 4.0--6.0 & P & Double-faced bookshelves (grid) \\
\bottomrule
\end{tabular}
\end{table}

%% file: tables/scene_bundle.tex
\begin{center}
\begin{tabular}{ll}
\toprule
\textbf{File} & \textbf{Contents} \\
\midrule
\texttt{scene.blend}              & Blender scene file (used for re-rendering during evaluation) \\
\texttt{color.png}                & RGB render, $1280 \times 720$, 8-bit \\
\texttt{instance\_segmentation.png} & Per-object index map, $1280 \times 720$, 16-bit grayscale \\
\texttt{metadata.json}            & Camera, environment, render statistics, and instance-index lookup \\
\texttt{GT.json}                  & Ground-truth layout: room dimensions and per-object pose \\
\texttt{objects\_in\_view.json}   & Filtered in-view object list (Stage 2 preprocessing output) \\
\bottomrule
\end{tabular}
\end{center}

%% file: sections/AppB_evaluation.tex
% =====================================================================
% Appendix B. Evaluation Protocol Details
% =====================================================================

\section{Evaluation Protocol Details}
\label{app_b}

% =====================================================================
% B1
% =====================================================================

\subsection{Coordinate Alignment}
\label{app_b1}

\paragraph{Frame transform.} The model emits each predicted object's center and yaw in the 
\textbf{model frame} $\mathcal{M}$: a Z-up right-handed frame with 
$+Y$ along the camera's horizontal forward direction and $+X$ to its 
right. Since $\mathcal{M}$ and the world frame $\mathcal{W}$ 
(Appendix~\ref{app_a2}) share the same origin and $+Z$ axis, they 
differ only by a rotation about $Z$ (unlike the per-asset matrix 
$M^w_o$ of Appendix~\ref{app_a2}, this $\mathcal{M}\!\to\!\mathcal{W}$ 
map is a single scene-level rigid transform shared by all predicted 
objects).

\paragraph{Camera heading.}
We read \texttt{camera.rotation\_euler} from \texttt{metadata.json}: 
in Blender's convention a camera at rest 
(\texttt{rotation\_euler}$=[0,0,0]$) looks along world $+Y$, so 
applying the recorded $R_z(r_z)$ to the rest forward vector $(0,1,0)$ 
yields the floor-plane forward direction $(-\sin r_z,\,\cos r_z)$. Its 
angle from world $+Y$, measured CCW per Appendix~\ref{app_a2}'s yaw 
convention, is
\begin{equation}
\theta_{\text{cam}}
\;=\; \mathrm{atan2}\!\bigl(-\sin r_z,\;\cos r_z\bigr)
\;\in\; (-180^\circ,\, 180^\circ].
\end{equation}
Tilt and roll ($r_x, r_y$) are discarded, since the floor-plane 
projection depends only on $r_z$.

\paragraph{Snap and per-object transform.}
Since every camera is placed within $\pm45^\circ$ of a cardinal 
direction, we snap to the nearest multiple of $90^\circ$ and apply the 
inverse rotation,
\begin{equation}
\Delta\theta \;=\; -90^\circ \cdot \mathrm{round}\!\left(\theta_{\text{cam}} / 90^\circ\right)
\;\in\; \{-180^\circ, -90^\circ, 0^\circ, 90^\circ\, 180^\circ\}.
\end{equation}
For each predicted object with center $\mathbf{p}=[x,y,z]$ and yaw 
$\theta$ in $\mathcal{M}$, the world-frame counterpart is
\begin{equation}
\begin{bmatrix} x_w \\ y_w \end{bmatrix}
= R(\Delta\theta)\!\begin{bmatrix} x \\ y \end{bmatrix},
\qquad
z_w = z,
\qquad
\theta_w = (\theta + \Delta\theta) \bmod 360^\circ.
\label{eq:cam2world}
\end{equation}

This transform is applied to every prediction before any downstream 
use.

% =====================================================================
% B2
% =====================================================================

\subsection{Object Matching}
\label{app_b2}

\paragraph{Non-grid scenes.}
Non-grid scenes (bedroom, living room, kitchen, bathroom, dining
room, home studio, meeting room) are matched per category rather 
than globally, so that a category confusion (e.g.\ a sofa 
misidentified as a bookcase) is recorded as a missed detection plus 
a hallucination rather than being absorbed into a single distance-based 
pairing. Let $\mathcal{O}_c$ and $\mathcal{P}_c$ denote the sets of 
in-view GT and predicted objects of category $c$ (lower-cased before 
grouping). Matching proceeds in two stages:
\begin{enumerate}
    \item \textbf{Category gating.} Categories present on only one 
    side are routed directly to the unmatched pool: GT-only 
    categories contribute missed detections (RR, Appendix~\ref{app_b3}), 
    prediction-only categories contribute hallucinations (NHR, 
    Appendix~\ref{app_b3}).
    \item \textbf{Per-category assignment.} For each category present 
    on both sides, we build the cost matrix
    $C^{(c)}_{ij} = \|\mathbf{p}_i^{(g)} - \mathbf{p}_j^{(p)}\|_2$
    from world-frame XY centers and solve the Hungarian assignment. 
    When $|\mathcal{O}_c| \neq |\mathcal{P}_c|$, the surplus side is 
    routed to the same unmatched pool as stage~1.
\end{enumerate}

\paragraph{Grid scenes.}
Classroom, library, and open-office scenes contain dominant 
categories (student desks, library blocks, cell shelves) in 
axis-aligned arrays, where per-instance Hungarian assignment is 
largely arbitrary when predictions scatter without grid structure. 
We therefore drop object-level matching for these scenes and evaluate 
grid-level structure instead: GCR measures whether predicted objects 
fall into GT grid cells, GSR measures whether they form a consistent 
grid of their own (Appendix~\ref{app_b3}).

Hungarian matching is still applied to grid scenes during 
reconstruction (Appendix~\ref{app_b5}) purely as a visualization 
device: to tie each predicted pose to a specific GT asset for 
rendering, and plays no role in the reported metrics. It is restricted 
to the grid target groups: $\{$\texttt{simple\_desk}, 
\texttt{simple\_desk\_base}, \texttt{student\_desk}$\}$ against 
$\{$\texttt{swivel\_chair}, \texttt{student\_chair}$\}$ in classrooms; 
$\{$\texttt{simple\_desk}, \texttt{simple\_desk\_base}$\}$ against 
$\{$\texttt{swivel\_chair}$\}$ in offices; and a single 
$\{$\texttt{library\_block}$\}$ group in libraries. Non-target 
categories fall back to the per-category Hungarian above.

\paragraph{Yaw exemption rules.}
Categories with undefined or symmetric yaw are exempted from the 
rotation error (RE) metric, either excluded entirely (full case) or 
folded into a reduced angular range before averaging (symmetric 
cases), per Eq.~\ref{eq:ra}. Defined in \texttt{scene\_config.py} and 
applied uniformly across all models:
\begin{itemize}
    \item \textit{No defined yaw} (RE excluded entirely): \texttt{plant},
    \texttt{large\_plant\_container}, \texttt{floor\_lamp},
    \texttt{ceiling\_lamp}, \texttt{rug}
    \item \textit{Two-fold symmetry} ($\theta = \theta + 180^\circ$): \texttt{simple\_desk},
    \texttt{simple\_desk\_base}, \texttt{blackboard}, \texttt{cell\_shelf},
    \texttt{library\_block}, \texttt{coffee\_table}
    \item \textit{Four-fold symmetry} ($\theta = \theta + 90k^\circ$): \texttt{side\_table}
\end{itemize}
This list is a superset of categories appearing in IDEAL-Scenes 
(e.g.\ \texttt{plant} is configured but unused) and does not affect 
any reported metric.

% =====================================================================
% B3
% =====================================================================

\subsection{Metric Definitions}
\label{app_b3}

We organise the evaluation along five dimensions, D1--D5,
each targeting a distinct failure mode of indoor-layout prediction.
Non-grid rooms are scored on D1--D4; grid rooms
(classroom, library, office) are scored on D1, D2, and
D5, since per-object matching is ill-posed under the
near-identical, repeated furniture of regular layouts.
 
\paragraph{Notation.}
For a single scene let $\mathcal{O}$ be the full set of ground-truth
objects and $\mathcal{O}_{\mathrm{iv}} \subseteq \mathcal{O}$ the
\emph{in-view} subset, i.e.\ those whose center lies in the camera
frustum used to render the input image. Let $\mathcal{P}$ be the
predicted objects, and write
$N = |\mathcal{O}|$, $K = |\mathcal{O}_{\mathrm{iv}}|$, $P = |\mathcal{P}|$.
A matching
$\mathcal{M} \subseteq \mathcal{O}_{\mathrm{iv}} \times \mathcal{P}$
is obtained by category-wise Hungarian assignment on XY-plane Euclidean
distance (Appendix~\ref{app_b2}); we write $K_m = |\mathcal{M}|$ and
$H = P - K_m$ for the number of unmatched predictions.

Each object carries a position $\mathbf{p} = (x, y, z) \in \mathbb{R}^3$,
local axis-aligned dimensions $(W, D, H)$, and a yaw
$\psi \in [0, 360^\circ)$. Throughout this section, $\mathrm{AABB}(o)$
denotes the world-axis bounding box obtained by rotating the local
footprint by $\psi$, taking its axis-aligned hull on the XY plane, and
centring vertically on $\mathbf{p}_z$.

\input{tables/metrics_and_scenetypes.tex}

\subsubsection{D1 — Scene Validity}
\label{app:d1}

D1 certifies structural output admissibility before any geometry is scored: the output must be parsed into the required schema, and the predicted room type must match the ground truth; without this gate, a model could trivially evade D2 - D5   by emitting malformed scenes.
 
\paragraph{Parse Success Rate (PSR).}
PSR measures the fraction of scenes for which the model emits a JSON
output that satisfies our minimal schema (top-level \texttt{scene} object
with a typed \texttt{room} and a list of objects, each carrying an \texttt{id},
\texttt{category}, 3-vector \texttt{position} and \texttt{dimensions}, and an
optional numeric \texttt{rotation\_yaw}). Per-scene PSR is the indicator
$\mathds{1}[\text{output parses and validates}]\in\{0,1\}$; the dataset-level
score averages over \emph{all} evaluated scenes:
\begin{equation}
\mathrm{PSR} \;=\; \frac{1}{|\mathcal{S}|}\sum_{s\in\mathcal{S}}
\mathds{1}\!\left[\,s \text{ parses and passes schema validation}\,\right].
\end{equation}
Parse-failed scenes contribute $0$ to the numerator but remain in the
denominator, so failures genuinely penalise the model.
 
\paragraph{Room Type Accuracy (RTA).}
RTA checks whether the predicted \texttt{scene.room.type} string matches
the ground-truth room type. The dataset-level score
conditions on parse success, so a single failure mode is not double-counted
between PSR and RTA:
\begin{equation}
\mathrm{RTA} \;=\;
\frac{\sum_{s\in\mathcal{S}}\mathds{1}[\text{parse success}]\cdot
      \mathds{1}[\hat{t}_s = t_s]}
     {\sum_{s\in\mathcal{S}}\mathds{1}[\text{parse success}]},
\end{equation}
where $t_s$ and $\hat{t}_s$ are the ground-truth and predicted room types.

% ---------------------------------------------------------------------

\subsubsection{D2 — Physical Plausibility}

D2 measures whether the predicted scene respects two basic physical constraints \emph{independently of the ground truth}: no interpenetration and all furniture within room bounds. Both metrics use $P$ (total predictions) as the denominator.

\paragraph{Non-Overlap Rate (NOR).}
NOR is the fraction of predicted objects that are \emph{not} involved in any
pairwise collision. Two predictions $i,j$ are considered as colliding if
their 3D AABBs satisfy
$\mathrm{IoU}_{3\mathrm{D}}\!\bigl(\mathrm{AABB}(o_i),\mathrm{AABB}(o_j)\bigr)
\;\geq\; \tau_{\mathrm{col}}$, with $\tau_{\mathrm{col}}=0.03$ chosen to
ignore the tiny grazing contacts that arise from finite numerical precision.
Let $Q=|\{i:\exists j\neq i,\,\mathrm{IoU}_{3\mathrm{D}}(o_i,o_j)\geq \tau_{\mathrm{col}}\}|$.
Then
\begin{equation}
\mathrm{NOR} \;=\; 1 - \frac{Q}{P}, \qquad P>0.
\end{equation}
 
\paragraph{In-Bounds Rate (IBR).}
IBR is the fraction of predicted objects whose AABB lies entirely within the
room volume. Given room dimensions $(W_r,D_r,H_r)$ and a room frame centerd
on the floor, the room volume is
$[-W_r/2, W_r/2]\times[-D_r/2, D_r/2]\times[0, H_r]$. Let $B$ count the
predictions whose AABB violates any of these bounds. Then
\begin{equation}
\mathrm{IBR} \;=\; 1 - \frac{B}{P}, \qquad P>0.
\end{equation}

% ---------------------------------------------------------------------

\subsubsection{D3 — Geometric Accuracy (Non-Grid Rooms)}
 
D3 scores how accurately matched predictions reproduce ground-truth geometry, decomposed into position, scale (via 3D IoU), and orientation. Each metric is reported as a \emph{pass rate} at a fixed threshold and a supplementary continuous mean; all three are computed over $\mathcal{M}$ with $K_m$ as the denominator.

\paragraph{Vertical-axis convention.} For each matched pair whose
ground-truth category is floor-standing (i.e.\ not wall-mounted), we
overwrite the predicted $z$-coordinate with the ground-truth $z$ before
evaluating D3. This matches the reconstruction convention used elsewhere in the paper. Wall-mounted predictions retain their predicted $z$.
 
\paragraph{Position Accuracy (PA).}
For each matched pair $(g,p)\in\mathcal{M}$ define the XY-plane error
$\mathrm{PE}(g,p)=\sqrt{(g_x-p_x)^2 + (g_y-p_y)^2}$.
The primary metric is the pass rate at $\tau_{\mathrm{PE}}=0.3\,\mathrm{m}$,
\begin{equation}
\mathrm{PA} \;=\; \frac{1}{K_m}\sum_{(g,p)\in\mathcal{M}}
\mathds{1}\!\left[\mathrm{PE}(g,p)\leq \tau_{\mathrm{PE}}\right].
\end{equation}
The supplementary continuous mean is
$\overline{\mathrm{PE}} = K_m^{-1}\sum_{(g,p)\in\mathcal{M}} \mathrm{PE}(g,p)$
(metres). The threshold is chosen to be coarser than typical furniture
half-extents, so PA reflects \emph{slot-level} placement rather than
sub-centimeter alignment.
 
\paragraph{Precision IoU (Prec.).}
Geometric agreement is measured by the 3D IoU of world-axis AABBs:
\begin{equation}
\mathrm{IoU}_{3\mathrm{D}}(g,p)
\;=\;
\frac{\mathrm{vol}\bigl(\mathrm{AABB}(g)\cap\mathrm{AABB}(p)\bigr)}
     {\mathrm{vol}\bigl(\mathrm{AABB}(g)\cup\mathrm{AABB}(p)\bigr)}.
\end{equation}
The primary metric is the pass rate at $\tau_{\mathrm{IoU}}=0.15$,
\begin{equation}
\mathrm{Prec.} \;=\; \frac{1}{K_m}\sum_{(g,p)\in\mathcal{M}}
\mathds{1}\!\left[\mathrm{IoU}_{3\mathrm{D}}(g,p)\geq \tau_{\mathrm{IoU}}\right],
\end{equation}
with the supplementary continuous mean $\overline{\mathrm{IoU}}$. The
permissive threshold of $0.15$ is calibrated against typical furniture
aspect ratios: it credits a prediction that recovers the correct slot and
gross scale, without demanding sub-centimeter alignment.

\paragraph{Rotation Accuracy (RA).}
Rotation error is computed with category-aware symmetry folding, so that an
object indistinguishable under a $90^\circ$ or $180^\circ$ rotation is not
penalised for a symmetric flip. More details in Appendix~\ref{app_b2}. Given a category $c$, define
\begin{equation}
\label{eq:ra}
\mathrm{RE}(g,p)\;=\;
\begin{cases}
\mathrm{fold}_{45^\circ}\!\big((g_\psi-p_\psi)\bmod 90^\circ\big), & c\in\mathcal{C}_{90},\\[4pt]
\mathrm{fold}_{90^\circ}\!\Big(\mathrm{fold}_{180^\circ}\!\big((g_\psi-p_\psi)\bmod 360^\circ\big)\Big), & c\in\mathcal{C}_{180},\\[4pt]
\mathrm{fold}_{180^\circ}\!\big((g_\psi-p_\psi)\bmod 360^\circ\big), & \text{otherwise},
\end{cases}
\end{equation}
where $\mathrm{fold}_{\theta}(x):=\min(|x|,2\theta-|x|)$ folds an angle into
$[0,\theta]$, and $\mathcal{C}_{90},\mathcal{C}_{180}$ are the
$90^\circ$- and $180^\circ$-symmetric category sets. Yaw-isotropic
categories (e.g.\ round stools) are excluded from the average. Let $K_r$
denote the count of matched pairs evaluated under this rule. Then
\begin{equation}
\mathrm{RA} \;=\; \frac{1}{K_r}\sum_{(g,p)} \mathds{1}\!\left[\mathrm{RE}(g,p)\leq \tau_{\mathrm{RE}}\right],
\qquad \tau_{\mathrm{RE}}=30^\circ,
\end{equation}
with supplementary mean $\overline{\mathrm{RE}}$ (degrees).

% ---------------------------------------------------------------------
\subsubsection{D4 — Object Recognition (Non-Grid Rooms)}

D4 measures whether the model recovers the correct \emph{set} of objects independently of placement, reporting under- and over-prediction separately via two complementary rates.

\paragraph{Recognition Rate (RR).}
RR is the fraction of in-view ground-truth objects that are matched by
some prediction:
\begin{equation}
\mathrm{RR} \;=\; \frac{K_m}{K}, \qquad K>0.
\end{equation}
Restricting the denominator to $K$ rather than $N$ avoids penalising the
model for objects outside the camera frustum, which the input image does
not contain.
 
\paragraph{Non-Hallucination Rate (NHR).}
NHR is the fraction of predictions that are matched (i.e.\ not
hallucinated):
\begin{equation}
\mathrm{NHR} \;=\; 1 - \frac{H}{P} \;=\; \frac{K_m}{P}, \qquad P>0.
\end{equation}
Reporting $\mathrm{NHR}$ rather than the raw hallucination count keeps the
metric on $[0,1]$ and comparable across scenes of different sizes.

% ---------------------------------------------------------------------
\subsubsection{D5 — Grid Layout}

Grid rooms (classroom, library, office) contain near-identical repeated furniture, making per-object Hungarian matching degenerate. D5 replaces matching with \emph{layout-level} evaluation: GCR measures whether predicted objects fall into the correct GT grid cells, and GSR measures whether the predicted arrangement is itself self-consistently grid-shaped; both are computed per target group and pooled per scene.

\paragraph{Grid cell extraction.}
D5 compares the prediction against a ground-truth lattice
recovered from the GT centers $\{(x_i,y_i)\}$ of the target category:
\begin{enumerate}
    \item Round each $x_i$ to two decimals, deduplicate, and sort to obtain
    column coordinates $\mathsf{col\_xs}$ ($C$ values); analogously for
    $y_i$ to obtain row coordinates $\mathsf{row\_ys}$ ($R$ values).
    \item Estimate the row pitch as the median of consecutive differences
    in $\mathsf{row\_ys}$, and the column pitch analogously. A singleton
    axis ($R=1$ or $C=1$) borrows the other axis's pitch.
    \item Form the $R\times C$ lattice; each cell is the axis-aligned
    rectangle with width equal to the column pitch and height equal to
    the row pitch, centerd on its lattice point and abutting its
    neighbours with no gap.
\end{enumerate}
This procedure assumes axis-aligned arrangements parallel to the world
$X$/$Y$ axes, which the dataset generator guarantees for the supported
room types. A cell is \emph{valid} for D5 if it contains at
least one GT center; let $G$ denote the count of valid cells. Predicted
target objects are assigned to whichever cell contains their XY position;
when several predictions fall into the same cell, only the one nearest the
GT center is retained for the cell-coverage tally. The others remain in
the prediction pool for D2 but contribute
nothing to D5.
 
\paragraph{Grid Coverage Rate (GCR).}
Let the ground-truth grid induce a partition of the floor into rectangular
cells $\{c_k\}_{k=1}^{G}$, where $G$ is the number of cells that contain at
least one ground-truth object of the target category. A cell is \emph{hit}
if at least one prediction of the same category falls inside it:
\begin{equation}
\mathrm{GCR}\;=\;\frac{1}{G}\sum_{k=1}^{G}
\mathds{1}\!\left[\,\exists\, p\in\mathcal{P}_{\mathrm{cat}}:\; (p_x,p_y)\in c_k\right].
\end{equation}
GCR is bounded in $[0,1]$ and rewards correct \emph{positional coverage}
without penalising minor scale or orientation errors that D3
already addresses.
 
\paragraph{Grid Self-consistency Rate (GSR).}
GSR measures, independently of the ground truth, how grid-shaped the
prediction is by itself. Let $\{(p_x^{(i)}, p_y^{(i)})\}_{i=1}^{N_p}$ be
the XY positions of predicted objects in the target category. We cluster
the $x$-coordinates and the $y$-coordinates separately by $1$-D
single-linkage with tolerance $\tau_{\mathrm{cl}}=0.3\,\mathrm{m}$, yielding
$n_c$ column centers $\{\bar{x}_a\}$ and $n_r$ row centers $\{\bar{y}_b\}$.
The \emph{self-fitted lattice} is
$\Lambda=\{(\bar{x}_a,\bar{y}_b):a=1,\ldots,n_c;\,b=1,\ldots,n_r\}$, and
\begin{equation}
\mathrm{GSR}\;=\;\frac{1}{N_p}\sum_{i=1}^{N_p}
\mathds{1}\!\left[\,\min_{q\in\Lambda}\bigl\lVert (p_x^{(i)},p_y^{(i)}) - q\bigr\rVert_2
                  \;\leq\; \tau_{\mathrm{hit}}\right],
\end{equation}
with hit tolerance $\tau_{\mathrm{hit}}=0.1\,\mathrm{m}$.
GSR is undefined (and excluded from aggregation) in three degenerate
regimes that would otherwise yield trivially perfect scores: (i) $N_p<3$,
(ii) all predictions collapse to a single $1{\times}1$ cluster, and
(iii) the fill ratio $N_p/(n_r n_c)\leq 0.5$, indicating a sparse scatter
in which each point becomes its own row and column.

% =====================================================================
% B4
% =====================================================================

\subsection{Aggregate Score}
\label{app_b4}

The five dimensions are designed to be reported side-by-side, since a model strong on RR but weak on Prec. fails differently from one with the reverse profile. For ranking purposes we additionally report a single scalar \textsc{Overall} score at three levels of granularity, summarized in Table~\ref{tab:overall-levels} below.

\input{tables/score.tex}

D1 (PSR, RTA) is a scene-level binary flag and is not meaningful to average
within a single scene; it therefore enters only once scenes are pooled
across a room type or the full dataset.

\paragraph{Per-scene overall.}
For each parse-success scene $s$, let $\mathcal{R}_s$ be the set of
D2-D5 rate metrics applicable to its room type
(Table~\ref{tab:metric_scenetypes}):
\begin{equation}
\mathcal{R}_s \;=\;
\begin{cases}
\{\mathrm{NOR},\mathrm{IBR},\mathrm{PA},\mathrm{Prec.},\mathrm{RA},\mathrm{RR},\mathrm{NHR}\}, & s \text{ non-grid},\\[2pt]
\{\mathrm{NOR},\mathrm{IBR},\mathrm{GCR},\mathrm{GSR}\}, & s \text{ grid}.
\end{cases}
\end{equation}
Writing $m^{(s)}$ for the value of metric $m$ on scene $s$, we define
\begin{equation}
\textsc{Overall}_s \;=\;
\frac{1}{|\mathcal{R}_s'|}\sum_{m\in\mathcal{R}_s'} m^{(s)},
\qquad
\mathcal{R}_s' \;=\; \bigl\{\,m\in\mathcal{R}_s : m^{(s)}\neq\textsc{n/a}\,\bigr\},
\label{eq:overall_scene}
\end{equation}
where $\mathcal{R}_s'$ excludes any metric undefined for this scene
(e.g.\ all D3 metrics when $K_m=0$). Parse-failed scenes
are excluded from this stage. D1 does not enter the per-scene overall: PSR and RTA are
scene-level $\{0,1\}$ flags whose information is absorbed into the
parse-success gate and the aggregate overall below. All rate metrics
already follow the higher-is-better convention, so no sign inversion is
applied. For grid rooms with two target groups (e.g.\ desks and chairs),
GCR and GSR enter $\textsc{Overall}_s$ pooled as the unweighted mean
across groups present in the scene.

\paragraph{Per-room-type overall.}
Moving from a single scene to a room type $t$, we compute \textsc{Overall} by a
two-stage mean: first averaging each applicable rate metric across all scenes
of that room type, then averaging those per-metric means across metrics. For
each rate metric $m \in \{\text{PSR}, \text{RTA}\} \cup \bigcup_{s \in t} \mathcal{R}_s$,
let $\mathcal{S}_m^{(t)}$ be the set of scenes of room type $t$ on which $m$ is
both applicable and defined:
\begin{equation}
\begin{aligned}
\bar{m}^{(t)} &\;=\; \frac{1}{|\mathcal{S}_m^{(t)}|}\sum_{s \in \mathcal{S}_m^{(t)}} m^{(s)},
\qquad
\textsc{Overall}^{(t)} \;=\; \frac{1}{|\mathcal{M}^{\star (t)}|}\sum_{m \in \mathcal{M}^{\star (t)}} \bar{m}^{(t)}, \\[4pt]
\mathcal{M}^{\star (t)} &= \big\{\, m : |\mathcal{S}_m^{(t)}| \geq 1 \,\big\}.
\end{aligned}
\label{eq:per-room-type-overall}
\end{equation}
Since each room type is either non-grid or grid, $\mathcal{R}_s$ reduces to the
corresponding fixed set throughout: $\mathcal{M}^{\star (t)} = \{\text{PSR},
\text{RTA}\} \cup \{\text{NOR}, \text{IBR}, \text{PA}, \text{Prec.}, \text{RA},
\text{RR}, \text{NHR}\}$ for non-grid room types, and $\{\text{PSR},
\text{RTA}\} \cup \{\text{NOR}, \text{IBR}, \text{GCR}, \text{GSR}\}$ for grid
room types. Unlike the per-scene overall, D1 enters here: once pooled across
multiple scenes, PSR and RTA become meaningful proportions rather than single
binary flags. We report this breakdown for all ten room types in the
supplementary material.

\paragraph{Aggregate overall.}
Across the $S$ evaluated scenes we compute the leaderboard score
(Table~\ref{tab:ideal_leaderboard}) by a second mean. For each rate
metric $m\in\{\mathrm{PSR},\mathrm{RTA}\}\cup\bigcup_s\mathcal{R}_s$,
let $\mathcal{S}_m$ be the set of scenes on which $m$ is both
applicable and defined, and define
\begin{equation}
\bar m \;=\; \frac{1}{|\mathcal{S}_m|}\sum_{s\in\mathcal{S}_m} m^{(s)}.
\end{equation}
The denominator follows the conditioning rules of \ref{app:d1} while $\mathcal{S}_{\mathrm{RTA}}$ and all
D2--D5 sets are restricted to parse-success
scenes. The aggregate overall is the unweighted mean of these
per-metric means,
\begin{equation}
\textsc{Overall} \;=\;
\frac{1}{|\mathcal{M}^\star|}\sum_{m\in\mathcal{M}^\star} \bar m,
\qquad
\mathcal{M}^\star \;=\; \bigl\{\,m : |\mathcal{S}_m|\geq 1\,\bigr\},
\end{equation}
Every D1--D5 rate metric with at least one
contributing scene enters the aggregate with equal weight. The
continuous diagnostics
$\overline{\mathrm{PE}}$, $\overline{\mathrm{IoU}}$, and
$\overline{\mathrm{RE}}$ from D3 are reported separately and
do not enter \textsc{Overall}.

% =====================================================================
% B5
% =====================================================================

\subsection{Reconstruction Pipeline}
\label{app_b5}

The reconstruction pipeline turns a model's predicted JSON into a 
rendered image $I_{\text{pred}}$, reusing the original 
\texttt{scene.blend} and overriding only the per-object poses of 
matched assets, hiding unmatched GT assets, and marking hallucinated 
predictions—so that the only image-level difference between 
$I_{\text{gt}}$ and $I_{\text{pred}}$ is attributable to the prediction 
itself.

\paragraph{Pipeline steps.}
\begin{enumerate}
    \item \textbf{In-view filter.} Intersect \texttt{GT.json} object 
    ids with \texttt{objects\_in\_view.json} to obtain 
    $\mathcal{O}_{\mathrm{iv}}$, as used by the numerical metrics.
    \item \textbf{Frame transform.} Apply Eq.~\ref{eq:cam2world} to 
    bring the prediction into $\mathcal{W}$.
    \item \textbf{Match.} Run the matcher of \S~\ref{app_b2} to obtain 
    $\mathcal{M}$.
    \item \textbf{Read original asset state.} For each matched GT 
    object, record its \texttt{obj.location} ($\mathbf{o}_{\text{orig}}$) 
    and \texttt{obj.dimensions} ($(W^o,D^o,H^o)$) via \texttt{bpy}.
    \item \textbf{Inverse pose.} Solve for the Blender pose realising 
    the predicted center, orientation, and dimensions (below).
    \item \textbf{Compose render JSON.} Per object: (a) an overridden 
    pose for matched GT, (b) a hide flag for unmatched in-view GT, or 
    (c) a hallucination marker (solid red sphere) for unmatched 
    predictions.
    \item \textbf{Re-render.} Apply the overrides and re-render from 
    the original camera pose, yielding $I_{\text{pred}}$.
\end{enumerate}

\begin{figure}[h]
    \centering
    \includegraphics[width=\linewidth]{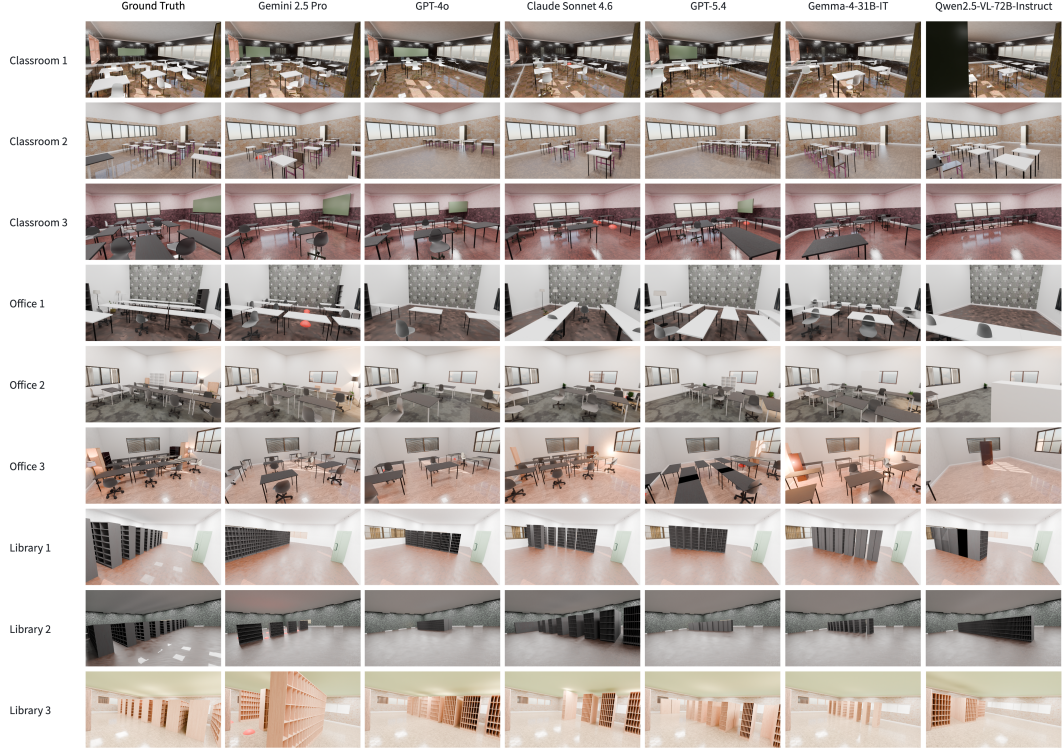}
    \vspace{0in}
    \vspace{-.2in}
    \caption{Grid layout scenes perceptual reconstruction.}
    \label{tab:grid_scenes}
\end{figure}

\paragraph{Inverse pose.}
The Blender state per-object comprises \texttt{obj.location} (root-mesh origin, generally \emph{not} the geometric center), \texttt{obj.rotation\_euler.z} ($\theta_b$) and \texttt{obj.dimensions} in local axes.  Given the prediction's world-frame center $\mathbf{p}_{\text{pred}}$, canonical $(W_p,D_p,H_p)$, and yaw $\theta_w^{(p)}$, yaw and dimensions are converted via Eq.~\ref{eq:blender_yaw}, which applies a $90^\circ$ offset and $W\!\leftrightarrow\!D$ swap for most categories, with exceptions for yaw-ignore categories 
$\mathcal{Y}_{\text{ignore}}$ (\S~\ref{app_b2}) and the table-dining 
family $\mathcal{Y}_{\text{td}} = \{\texttt{coffee\_table}, 
\texttt{conference\_table}, \texttt{table\_dining}\}$.

\begin{equation}
\theta_b =
\begin{cases}
0, & c \in \mathcal{Y}_{\text{ignore}}, \\
\theta_w^{(p)}, & c \in \mathcal{Y}_{\text{td}}, \\
(\theta_w^{(p)} + 90^\circ) \bmod 360^\circ, & \text{otherwise},
\end{cases}
\qquad
(W^p_b, D^p_b, H^p_b) =
\begin{cases}
(W_p, D_p, H_p), & c \in \mathcal{Y}_{\text{ignore}} \cup \mathcal{Y}_{\text{td}}, \\
(D_p, W_p, H_p), & \text{otherwise}.
\end{cases}
\label{eq:blender_yaw}
\end{equation}

The root-to-center offset is recovered from the GT pose: $\mathbf{u}_w = \mathbf{o}_{\text{orig}} - \mathbf{p}_{\text{gt}}$ is un-rotated by the GT's Blender yaw 
$\theta_b^{\,\text{orig}}$, scaled by the prediction-to-original dimension 
ratio $(W^p_b/W^o_b, D^p_b/D^o_b, H^p_b/H^o_b)$, and re-rotated by 
$\theta_b$; adding the result to $\mathbf{p}_{\text{pred}}$ yields 
$\mathbf{o}_{\text{final}}$. The write-back assigns 
$\texttt{obj.location} = \mathbf{o}_{\text{final}}$, 
$\texttt{obj.rotation\_euler.z} = \theta_b$, and 
$\texttt{obj.dimensions} = (W^p_b, D^p_b, H^p_b)$.

\paragraph{Hallucination markers.}
Each unmatched prediction is rendered as a solid red sphere at its 
predicted position, with radius
\begin{equation}
r = \max\bigl(0.02,\ \min(\bar d_p / 2,\ 0.3)\bigr)\ \text{m},
\end{equation}
where $\bar d_p$ is the mean of the predicted $(W_p,D_p,H_p)$ (capped 
at $0.3$\,m to avoid dominating the frame, floored at $0.02$\,m to 
remain visible).

% =====================================================================
% B6
% =====================================================================

\subsection{Judge VLM Prompt and Perceptual Score}
\label{app_b6}

\paragraph{Judge Design.}We initially attempted to re-render scene images from every model's 
prediction, but most reconstructions were not meaningful for scoring: 
assets were missing or the scene failed to compose altogether. We 
therefore restrict the perceptual evaluation to six representative 
models: the four top-performing proprietary models (Gemini 2.5 Pro, 
GPT-4o, Claude Sonnet 4.6, GPT-5.4), the strongest open-source model 
(Gemma-4-31B-IT), and Qwen2.5-VL-72B-Instruct for cross-comparison with 
other benchmarks, over 20 scenes per room type (200 total).

We use Claude Opus 4.7 as the judge. For each scene, a single API call 
presents the ground-truth image alongside all six reconstructions 
(order randomly shuffled to reduce position bias), and the judge 
assigns each a 1--5 Likert score for overall spatial similarity plus a 
strict total ranking that resolves ties. This single-call multi-image 
setting provides a shared comparison context and enforces consistent 
within-scene ranking. The judge is instructed to focus on spatial 
understanding rather than photorealism and to return JSON only. We 
repeat the judging five times and report averaged results. The full 
system prompt is reproduced below verbatim.

\input{docs/judge_vlm_prompt}

%% file: tables/metrics_and_scenetypes.tex
\begin{table}[t]
\centering
\small
\caption{Metric organization across the five evaluation dimensions
and their applicability to scene types. Full metric definitions are
provided in Section~5.2; \checkmark: applicable; $\times$: not applicable.}
\begin{tabular}{llcc}
\toprule
Dimension & Metrics & Unstructured & Grid-structured \\
\midrule
D1: Scene Validity        & RTA, PSR              & \checkmark & \checkmark \\
D2: Physical Plausibility & NOR, IBR              & \checkmark & \checkmark \\
D3: Geometric Accuracy    & PA, Prec., RA           & \checkmark & $\times$ \\
D4: Object Recognition    & RR, NHR                & \checkmark & $\times$ \\
D5: Grid Layout& GCR, GSR & $\times$ & \checkmark \\
\bottomrule
\end{tabular}
\vspace{-0.2in}
\label{tab:metric_scenetypes}
\end{table}

%% file: tables/score.tex
\begin{table}[h]
  \centering
  \caption{The three levels of \textsc{Overall}, computed at increasing scope. D1 (PSR, RTA) is a scene-level binary flag and is not meaningful to average within a single scene; it therefore enters only once scenes are pooled across a room type or the full dataset.}
  \label{tab:overall-levels}
  \begin{tabular}{lcc}
    \toprule
    Name & Dimensions included & Scope \\
    \midrule
    Per-scene overall     & D2, D3, D4 (non-grid); D2, D5 (grid)     & Single scene \\
    Per-room-type overall & D1, D2, D3, D4 (non-grid); D1, D2, D5 (grid) & All scenes of one room type \\
    Aggregate overall     & D1, D2, D3, D4, D5                        & All evaluated scenes \\
    \bottomrule
  \end{tabular}
  \vspace{0.5em}
\end{table}

%% file: docs/judge_vlm_prompt.tex
\begin{lstlisting}[style=promptstyle]

You are a strict indoor spatial-layout judge. You will compare one ground-truth reference image (GT) with multiple reconstructed images from different models for the same scene.
Task: Please score every candidate image against GT.png according to their spatial similarity impression, and return a JSON array only, following the required schema.

Output schema (JSON array)

[
  {
    "image_id": "string",
    "overall_spatial_impression_similarity": 1,
    "rank": 1
  }
]

Important constraints
- Judge spatial understanding, not photorealism.
- Do not over-penalize texture/material/style differences.
- Ignore minor lighting/color differences unless they clearly break spatial perception.
- Evaluate all candidate models in the same response for consistency.
- Output valid JSON only, no extra text.

Scoring rubric: overall_spatial_impression_similarity
- 1: clearly different room impression or function mismatch
- 2: major layout mismatch
- 3: broadly similar but key spatial organization differs
- 4: largely consistent layout with local mismatches
- 5: highly consistent overall spatial impression

Ranking rule
- Assign the rubric score (1-5) first; ties on the score are allowed.
- The final `rank` must still be strictly total: every candidate gets a unique rank from 1..N, no ties.
- To break ties, decide which candidate is closer to GT using your own judgment — e.g. finer-grained reasoning within the rubric, or your own spatial-semantic priors. The score field stays unchanged; only `rank` reflects the tie-break.

## User Prompt Template

You are given one scene with:
- Reference image: GT image
- Candidate reconstructions (same scene), each with a model name:
  - model_name_1: image_1
  - model_name_2: image_2
  - ...

\end{lstlisting}

%% file: sections/AppC_setup.tex
% =====================================================================
% Appendix C. Experimental Setup
% =====================================================================

\section{Experimental Setup}
\label{app_c}

% =====================================================================
% C1
% =====================================================================

\subsection{Model List, Access, and Inference Protocol}
\label{app_c1}

\paragraph{Models.} Table~\ref{tab:model_list} lists the 15 evaluated 
models with parameter counts, access modality, and parsed-scene counts. Open-weight models are run locally; proprietary models are accessed via their official APIs.

\input{tables/model_list}

\paragraph{Inference protocol.} For every model, we issue a single 
forward pass per scene, with no self-consistency or best-of-N sampling 
and no in-context examples from IDEAL-Scenes. The user message contains 
the reference image, room-type candidates $\mathcal{T}$, and category list 
$\mathcal{C}$; the system message specifies coordinate conventions, 
the estimation procedure, and the output schema (Appendix~\ref{app_c2}). 
No depth, camera intrinsics and extrinsics, or instance masks are provided.

We adopt a low-entropy decoding regime (temperature=$0.1$, top-$p$ 
typically $0.95$) across all models for consistent comparison, with 
output length controlled via each backend's native mechanism.

% =====================================================================
% C2
% =====================================================================

\subsection{System prompt and output schema}
\label{app_c2}

\paragraph{Prompt design.}
We use a single shared system prompt across all evaluated models, 
fixing the room-centered coordinate convention and the output schema, 
with one schema-level example. The prompt specifies a fixed estimation 
order (canonical dimensions $\to$ yaw $\to$ position) to structure how 
each object is estimated, but does not elicit or score any explicit 
intermediate reasoning text: models are instructed to return JSON only, 
with no chain-of-thought or self-correction directives, since 
prompt-side scaffolds risk measuring prompt sensitivity rather than 
spatial reasoning. The full system prompt is reproduced below, with 
minor whitespace and section-divider formatting removed for layout.
\input{docs/system_prompt}

% =====================================================================
% C3
% =====================================================================

\subsection{Parsing rules}
\label{app_c3}

\paragraph{Parse outcomes.}
The inference wrapper attempts to recover a valid JSON object from 
each model response, yielding a binary \texttt{parse\_ok}/%
\texttt{parse\_failed} status stored in the \texttt{\_parse\_status} 
field; \texttt{parse\_failed} predictions contribute only to metric D1-PSR (see Appendix~\ref{app_b3}).

\paragraph{Recovery steps.}
Recovery is attempted in order, stopping at the first success: 
(S1) Trim whitespace and parse directly; (S2) Strip \verb|```| or \verb|```json| code fences; (S3) Extract the first 
balanced \texttt{\{...\}} span from surrounding prose; (S4) Close 
unbalanced opening braces from truncated output. We do not attempt 
general-purpose repair (e.g.\ correcting escape or quoting errors) 
that could alter the model's intended content.

\paragraph{Failure modes.}
A prediction is marked \texttt{parse\_failed} when recovery yields no 
valid JSON, the response is empty or refusal-only, the output is 
truncated before any top-level \texttt{\{}, or the result lacks the 
required \texttt{scene}/\texttt{room}/\texttt{objects} structure. 
Per-model \texttt{parse\_ok} counts on the $S=1{,}000$ scene benchmark 
are listed in the Parsed column of Table~\ref{tab:model_list}.

%% file: tables/model_list.tex
\begin{table}[h]
\centering
\caption{Evaluated models. \emph{Parsed} is the number of scenes (out of 1{,}000)
on which a schema-valid JSON was recovered.}
\label{tab:model_list}
\small
\begin{tabular}{lllrr}
\toprule
Model & Family & Access & Params & Parsed \\
\midrule
Gemini 2.5 Pro          & Proprietary & API   & --              & 1000 \\
GPT-4o                  & Proprietary & API   & --              & 1000 \\
GPT-5.4                 & Proprietary & API   & --              & 1000 \\
Claude Sonnet 4.6       & Proprietary & API   & --              & 1000 \\
\midrule
Gemma-4-31B-IT          & Open        & local & 31B             & 999  \\
GLM-4.6V-Flash          & Open        & local & 9B              & 925  \\
GLM-4.6V                & Open        & local & 106B            & 1000 \\
Qwen2.5-VL-72B-Instruct & Open        & local & 72B             & 1000 \\
Qwen3-VL-8B-Instruct    & Open        & local & 8B              & 1000 \\
Qwen3-VL-30B-A3B-Instruct        & Open   & local & 30B (3B act.)   & 920  \\
Qwen3-VL-235B-A22B-Instruct      & Open   & local & 235B (22B act.) & 760  \\
LLaVA-1.6-Mistral-7B    & Open        & local & 7B              & 981  \\
InternVL3.5-8B          & Open        & local & 8B              & 437  \\
InternVL3.5-30B-A3B     & Open   & local & 30B (3B act.)   & 195  \\
Janus-Pro-7B            & Open        & local & 7B              & 1000 \\
\bottomrule
\end{tabular}
\end{table}

%% file: docs/system_prompt.tex
% Styles defined in docs/listings_styles.tex (loaded in preamble).
\label{sp}
\begin{lstlisting}[style=promptstyle]
# Task
You are a spatial scene analyst with excellent spatial understanding capability. You are given one indoor image and a user message containing the possible assets list in this scene and the possible roomtype of this scene. Your task is to recognize visible objects in the scene, estimate their layout, and describe the scene as accurately as possible in a structured JSON format.

# User Message Hint
Each API request's user message includes:
- assets_list: snake_case identifiers for object categories that may appear
  (e.g. bed, floor_lamp, side_table). Your output category must use
  these strings exactly.
- allowed_room_types: the only allowed values for scene.room.type. Choose
  exactly one string that best matches the room visible in the image. Do not
  infer room.type from folder names, file paths, or any metadata.

## Assets Recognition
1. Use the image to decide what is actually visible.
2. The assets_list does not guarantee that every listed item appears in the
   scene, nor that each item appears only once.
3. Only include objects you see. Do not output categories outside
   assets_list, and do not invent instances you do not see.
4. Output one instance per visible object. A scene may contain multiple
   instances of the same category; distinguish them with different id
   values (e.g. floor_lamp_01, floor_lamp_02).
5. When multiple instances of the same category are arranged in a regular
   pattern (e.g., rows or a grid), reflect this regularity in their
   positions, and account for perspective distortion.

### Special Assets and Edge Cases
If you see a piece of furniture comprising multiple components that serve a
single function, please treat them as a single integrated object with
coordinates referenced to the ensemble's geometric center. Examples include:
- A mattress + a bed frame = bed
- A combination of upper (wall-mounted) and lower cabinets = kitchen_space
  (treat as a single integrated unit)
- Two back-to-back shelf grids sharing one base = library_block (a
  double-sided bookshelf unit; its depth spans both shelf faces even when
  only one side is visible in the image)

## Room type Recognition
Identify the room type from the semantic characteristics of the image (using
your prior knowledge about indoor spaces), and select the best-matching
option from the allowed_room_types list.
The room type below might have some special features:
- homestudio: open-plan residential space combining bedroom and living
  area in one room
- meetingroom: workplace meeting space (not a residential dining room)
- diningroom: residential dining space, may include kitchen appliances
  nearby

# Coordinate System (Room-Centered)
## Core Axioms --- Manhattan World
1. Orthogonal grid. Assume the room is a perfectly orthogonal 3D box. All
   walls meet at 90 degrees angles.
2. Discrete rotation. Most objects in the scene are aligned with this grid.
   Unless there is overwhelming visual evidence of diagonal placement, every
   object's rotation_yaw must be exactly one of [0, 90, 180, 270]. Snap all
   orientations to the nearest 90 degree axis.

## Axes Definition
- Origin (0, 0, 0): Geometric center of the room floor; Z = 0 is ground level.
- +Y (Depth axis): The ground-plane axis that runs parallel to the side walls
  (and perpendicular to the back wall), pointing away from the camera into
  the depth of the scene --- i.e., the direction along which objects recede
  into the distance. In a corner view, pick whichever of the two visible
  walls runs more parallel to the camera line of sight; +Y runs along that
  wall, pointing into the far end.
- +X: Points right (from the camera's perspective).
- +Z: Points up.
- Room boundary: X in [-W/2, W/2], Y in [-D/2, D/2]. All objects must stay
  strictly within these bounds.

## Camera Priors
The camera is placed inside the room, typically between 1.2 m and 2.0 m
above the floor, and oriented toward the room center with a 15 mm
ultra-wide lens (large horizontal field of view).

# Estimation Steps
All positions and dimensions MUST be in meters. The accuracy can be up to
0.01 meter.
When you estimate the layout of an object, follow this order for every
object:

Step 1 --- Canonical dimensions [W, D, H]
Estimate dimensions assuming the object is at yaw = 0 (standard
orientation). Standard orientation means the object's front faces +Y.
For examples:
- Bed: headboard at -Y, foot toward +Y
- Desk / table: long side perpendicular to Y-axis, user-side faces +Y
- Sofa / chair: backrest at -Y, seat faces +Y
Once you finish estimating the canonical dimensions, do not change them
regardless of how the object's yaw or position changes.

Step 2 --- Rotation yaw
rotation_yaw is the counterclockwise rotation angle (in degrees) about
the +Z axis, measured from the canonical orientation (front facing +Y).
For example, yaw=90 rotates the object's front from +Y toward -X.

  yaw | Meaning
  ----+----------------------------------------------
   0  | Front faces the back wall (away from camera)
   90 | Front faces the left wall
  180 | Front faces the camera
  270 | Front faces the right wall

Step 3 --- Position
Determine the position of the object relative to the room center (0, 0).
To determine the room center, carefully observe the intersection of the
floor and walls to estimate the scene boundaries.
The position must represent the exact coordinates of the object's
geometric center (x, y, z). For wall-mounted assets, ensure their
Z-coordinate accurately reflects their mounting height above the floor.

# Output format
1. Output valid JSON only. No explanation, no markdown fences.
2. You should follow strictly the json schema. Here is an example to
   describe a bedroom with one floor lamp:
   {
     "scene": {
       "room": {
         "type": "bedroom", "size": [W, D, H], "asset_count": 1
       },
       "objects": [
         {
           "id": "floor_lamp_01", "category": "floor_lamp",
           "position": [x, y, z], "dimensions": [W, D, H],
           "rotation_yaw": null
         }
       ]
     }
   }
3. room.asset_count must equal the length of the scene.objects array.
   Count after completing the array.
4. room.type must be exactly one string from allowed_room_types.
5. category must be exactly one string from assets_list.
6. Object id format: snake_case category + two-digit index, e.g.
   dining_table_01, floor_lamp_02.
7. Use rotation_yaw: null only for rotationally symmetric objects with
   no meaningful facing direction (e.g. floor_lamp, table_lamp,
   potted_plant, rug).
\end{lstlisting}

%% file: sections/AppD_results.tex
% =====================================================================
% Appendix D. Experimental Result
% =====================================================================

\section{Results and Analysis}

% =====================================================================
% D1
% ===================================================================== 
\subsection{Per-Room-Type Breakdown}
\label{app_d1}

We report per-room-type Overall scores in Table~\ref{tab:roomtype_ov}. Difficulty is relatively flat across the seven non-grid room types (spread under 10 points for the strongest models), with bathroom consistently easiest and meeting room highest for several leaders. The three grid rooms diverge markedly: library is easiest (single target category in a strict regular pattern), office is hardest (heterogeneous layout requiring correct world-frame anchoring), and classroom shows the widest cross-model spread, making it the most discriminative room type for grid-level reasoning. Weaker models (LLaVA-1.6-Mistral-7B, InternVL3.5-8B) stay close to the pack where category-room semantic priors dominate, but collapse sharply in larger, lower-density rooms (living room, home studio) where such priors are weak. Most visibly InternVL3.5-8B's drop to 0.0\% on classroom (see failure analysis in Appendix~\ref{app_d3}), consistent with the dense-scene collapse reported in Appendix~\ref{app_d2}.

\input{tables/overall_score_per_roomtype}

% =====================================================================
% D2
% =====================================================================

\subsection{Scene Difficulty Axes: Density and Camera Viewpoint}
\label{app_d2}

\subsubsection{Object Density}
\label{app_density}

We bin the 1{,}000 scenes by in-view object count $k_{\text{inview}}$ into five density bins: Bin~1 ($k_{\text{inview}}\in[1,6]$, 298 scenes), Bin~2 ($[7,9]$,
231 scenes), Bin~3 ($[10,14]$, 253 scenes), Bin~4 ($[15,24]$, 122
scenes), and Bin~5 ($[25,49]$, 96 scenes), reporting the unweighted mean per-scene Overall score over successfully parsed scenes. Parse stability and spatial-reasoning capacity are largely independent: most models maintain flat PSR while Overall declines with density, whereas Qwen3-VL-235B-A22B-Instruct shows the reverse: PSR rising from 73.5\% to 90.6\% as Overall collapses from 49.7\% to 16.9\%, meaning it becomes \emph{more} likely to emit parseable output as quality deteriorates.

\subsubsection{Viewpoint Difficulty}
\label{app_viewpoint}
\input{tables/camera_viewpoint}

\begin{figure}[t]
  \centering
  \includegraphics[width=\linewidth]{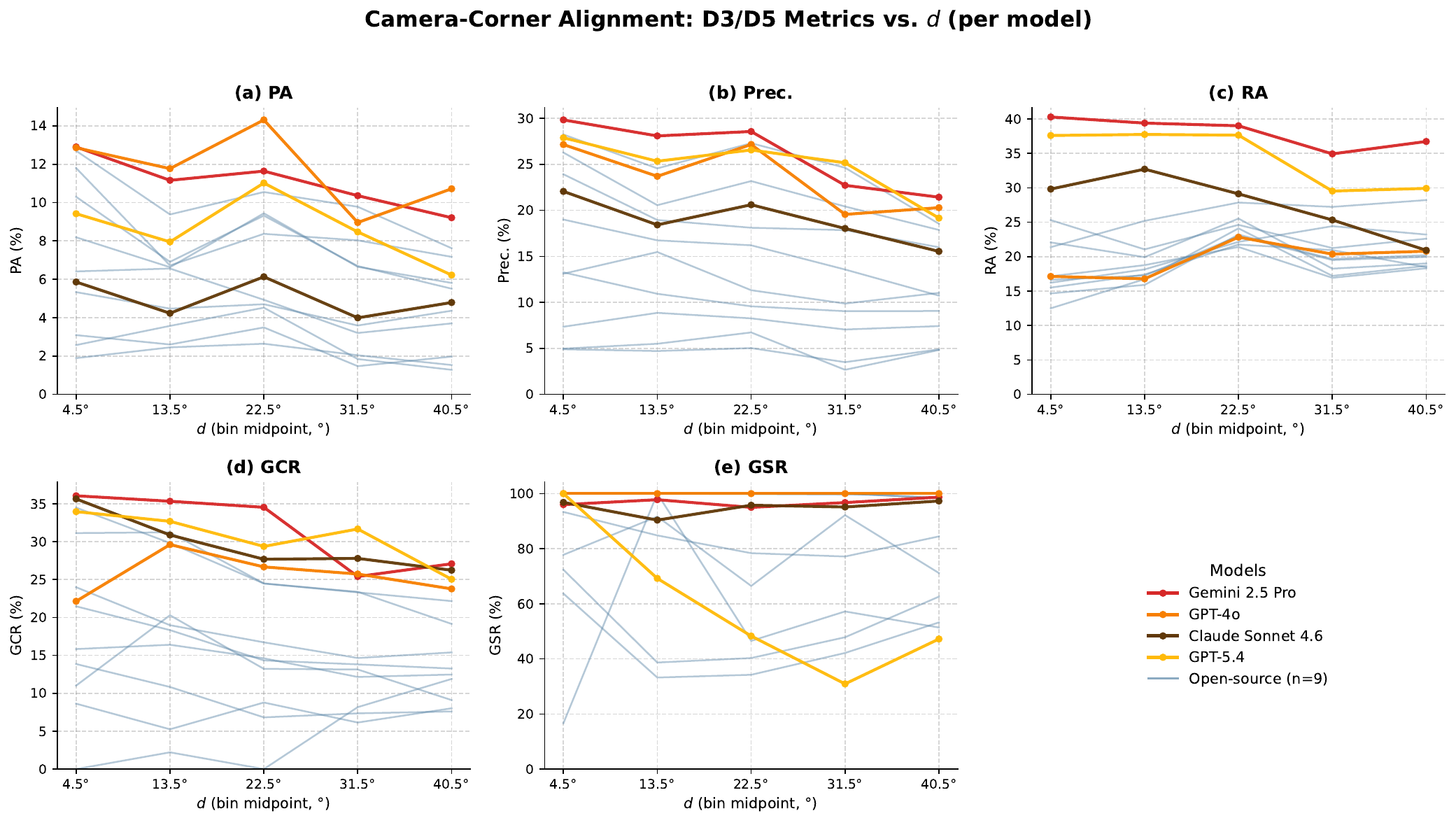}
  \caption{\textbf{D3/D5 metrics across camera viewpoints, per model.} Each line shows a model's absolute score (PA, Prec., RA, GCR, or GSR; \%) as a function of $d$, the angular distance between camera yaw and the nearest cardinal direction. Proprietary models are individually colored; the 9 open-source models are shown in light blue (several overlap near the ceiling in (e) GSR).}
  \label{fig:camera_viewpoint}
\end{figure}

Because $T_{\mathrm{align}}$ rounds the true camera heading to the nearest cardinal direction, we examine whether this introduces a difficulty gradient as the camera approaches a room diagonal. We partition the dataset into 5 bins over $d \in [0^\circ, 45^\circ]$ in steps of $9^\circ$, where $d$ is the angular distance to the nearest cardinal direction, and analyze D3 and D5 submetrics as a function of $d$ (Fig.~\ref{fig:camera_viewpoint}, Tab.~\ref{tab:camera_viewpoint}).

{Prec. and GCR degrade consistently as $d$ increases across nearly all models, while PA, RA, and GSR do not track this decline: PA fluctuates non-monotonically; across 13 models, RA increases for 9, and GSR stays within $\pm$10\% for 8. One exception is GPT-5.4, whose GSR drops sharply from 98\% to 30\% at high $d$: qualitative inspection reveals predicted positions following the 2D perspective layout rather than a reconstructed world-frame grid, consistent with directly mapping image coordinates to world coordinates rather than inverting the projection.}

{The asymmetry suggests reference-wall mismatch is not the dominant driver: if it were, position and rotation should degrade jointly since $T_{\mathrm{align}}$ applies the same rotation to both. Object density is also an incomplete explanation: in-view count peaks in the $[27^\circ,36^\circ)$ bin, but Prec. and GCR degrade most in $[36^\circ,45^\circ]$ where object count has already declined. We therefore read $T_{\mathrm{align}}$ discretization as a minor contributor at most; disentangling density from $d$ and marking the reference direction more explicitly during inference are left to future work.}

% =====================================================================
% D3
% =====================================================================

\subsection{Failure cases}
\label{app_d3}

Three models exhibit degenerate behavior that renders their metrics uninterpretable. \textbf{Janus-Pro-7B} does not engage with the task: it copies the JSON example from the system prompt verbatim, producing a single-object scene for nearly every input, and should be treated as a degenerate baseline. \textbf{InternVL3.5-30B-A3B} fails along two axes: a large fraction of outputs do not pass schema validation (truncated JSON, malformed wrappers), and among parse-success scenes, \texttt{asset\_count} is reliably inflated while the \texttt{objects} array contains only one or two entries: making downstream metrics uninterpretable. \textbf{InternVL3.5-8B} reports a high GSR, but this is built on a heavily filtered subset: the model fails to parse most scenes overall (overall PSR of 43.7\%) and is evaluated almost exclusively on library scenes, where collinear object rows trivially form a self-consistent line; its spatial metrics sit near the bottom of the table, revealing that the high GSR reflects internal regularity without true layout accuracy.

% =====================================================================
% D4
% =====================================================================

\subsection{Human-Correlation Validation}
\label{app_d4}

\input{tables/judge_vlm_full}

We evaluate whether the Judge VLM aligns with human perceptual judgment by uniformly sampling 200 scenes from IDEAL-Scenes (20 per room type). We recruit 15 annotators with computer-vision background, blinded to model identities and leaderboard statistics; residual preference bias may still exist. Scene assignments are randomized across room types, and each annotator scores all model reconstructions on a 1-5 Likert scale for \emph{overall spatial impression similarity} in a single-scene, multi-model view. A minimum observation time of 20\,s per scene is enforced; per-scene ratings are aggregated by mean (MOS), with samples flagged for quality review when (max--min $\ge 3$).

\begin{figure*}[t]
\centering
\includegraphics[width=\textwidth]{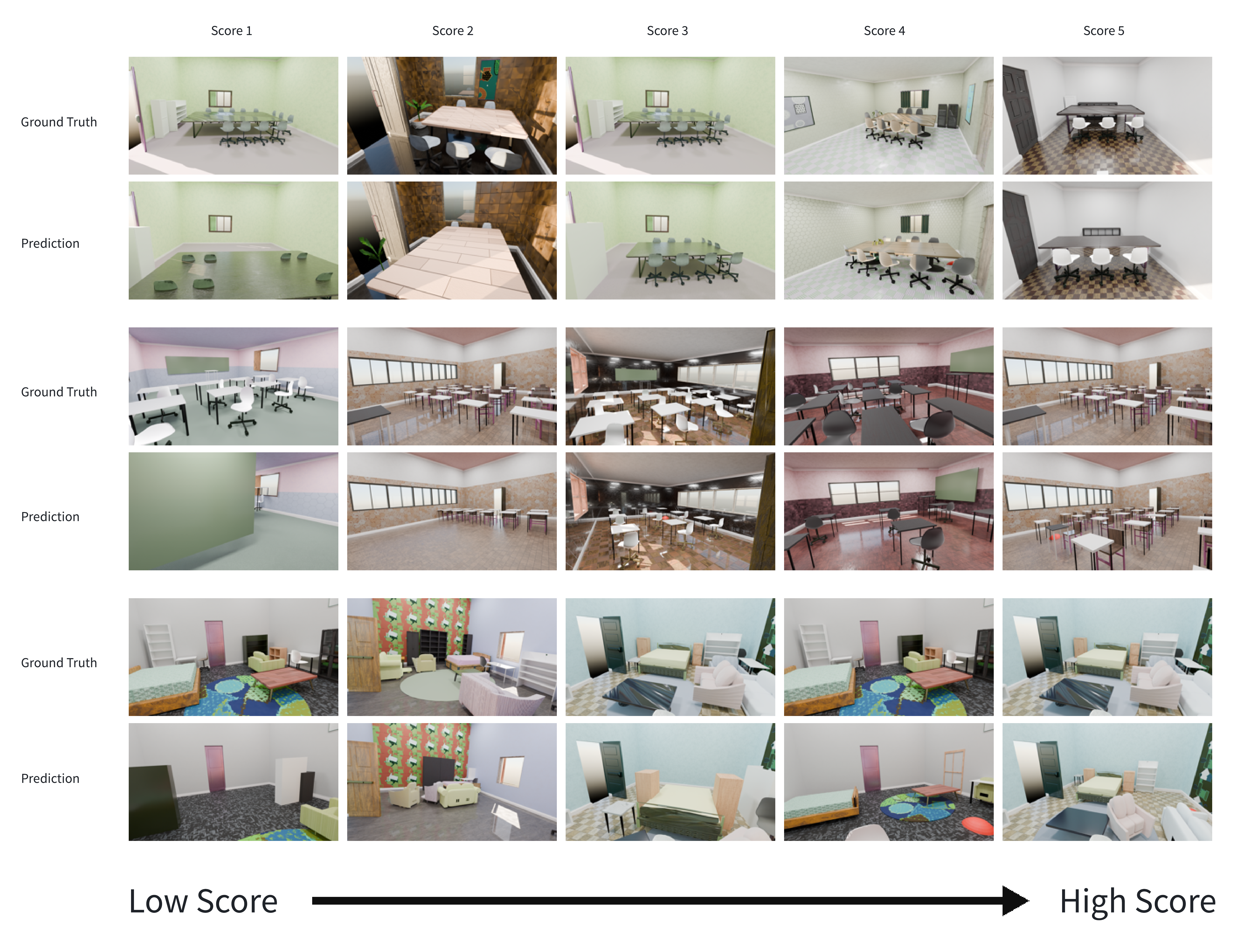}
\caption{
Qualitative examples of Judge-VLM perceptual scoring across three room categories.
Each row shows one room category, and each column corresponds to a target score from 1 to 5 (left to right).
For every cell, the top image is the ground-truth view (GT) and the bottom image is the model prediction (Pred).
The bottom arrow indicates increasing perceptual quality from LOW to HIGH.
}
\label{fig:judge-vlm-score-examples-3cats}
\end{figure*}

\paragraph{Human rubric}
Human annotation rubric for overall spatial impression similarity (1--5 Likert).
\begin{itemize}
    \item 1: clearly different room impression or function mismatch;
    \item 2: major layout mismatch;
    \item 3: broadly similar, but key spatial organization differs;
    \item 4: largely consistent layout with local mismatches;
    \item 5: highly consistent overall spatial impression.
\end{itemize}

\paragraph{MOS and scene-wise correlation.}
Per scene and model, human Mean Opinion Scores are:
\[
\mathrm{MOS}_{s,m} \;=\;
\frac{1}{|\mathcal{A}_{s,m}|}
\sum_{a\in\mathcal{A}_{s,m}} y_{a,s,m}.
\]
To match our evaluation granularity, we report scene-wise agreement:
for each scene we average MOS and Judge VLM scores across models, then compute
Spearman rank correlation across scenes.
We obtain $\rho = 0.79$ on overall spatial impression similarity.

% =====================================================================
% D5
% =====================================================================

\subsection{Cross-bench Analysis}
\label{app_d5}

\input{tables/cross_bench}

Tab.~\ref{tab:cross_bench_rank} re-ranks each model within the intersection of models jointly evaluated on IDEAL-Bench and each external benchmark. Caveats apply: intersections are small  ($n{=}3$ or $n{=}4$), benchmarks use incompatible score scales (so we compare ordinal ranks only), and empty cells reflect models not evaluated by the source benchmark (we did not re-run to fill them). The rank shifts are asymmetric: models consistently rank higher on QA-style and primitive-reconstruction benchmarks than on IDEAL-Bench, but not the reverse. This direction is consistent with the framing in Sec.~\ref{sec:diagnostic}: layout reconstruction is a harder ordinal test than the alternatives we compared against, since it requires every dimension to be approximately correct simultaneously rather than probing one at a time. The aggregate pattern therefore supports the same conclusion as the model-level cases discussed in the main text: IDEAL-Bench complements existing suites by exposing failure modes that QA-style and primitive-reconstruction evaluations cannot.

% =====================================================================
% D6
% =====================================================================

\subsection{Bootstrap Confidence Intervals}
\label{app_d6}

\input{tables/bootsrap}

{To quantify the uncertainty in Table~\ref{tab:ideal_leaderboard} arising from our fixed 1,000-scene sample, we compute 95\% confidence intervals via bootstrap resampling: for each model and metric, we resample scenes with replacement (sample size equal to the original number of applicable scenes) 2,000 times, recompute the metric on each resample, and report the 2.5th and 97.5th percentiles as the interval bounds. Table ~\ref{tab:bootstrap} reports point estimates alongside these intervals for all models and metrics.}

{Among the 12 adjacent rank pairs in the 13-model leaderboard, the overall ordering is well-supported at both ends: Gemini 2.5 Pro leads with a non-overlapping margin over GPT-4o (62.1 [61.6, 62.7] vs. 60.4 [59.9, 60.9]), and the bottom of the ranked field is similarly well-separated. Most rank transitions across tiers are confirmed by non-overlapping intervals. The exceptions are a small number of adjacent pairs within the same tier whose intervals overlap, indicating their relative order is not distinguishable at the 1,000-scene sample size; we flag these explicitly rather than over-interpreting the point-estimate ranking.}

{Two findings emerge that are not visible from point estimates alone. First, the GCR-GSR split is not an artifact of sampling noise: across all 13 models with a defined GSR, the GCR and GSR intervals are non-overlapping, so models consistently populate grid scenes sparsely (low GCR) while keeping the placed objects in a consistent regularity (high GSR). The same "grid illusion" pattern discussed in Sec.~\ref{sec:numerical_results} is now confirmed to hold beyond sampling noise. Second, the low PA scores reflect a genuine model-level ceiling rather than estimation noise: across all 13 models, the upper bound of the PA confidence interval never exceeds 12.8\% (best case: GPT-4o, [10.3, 12.8]). This strengthens our reading of geometric position estimation as a genuine and consistent bottleneck across evaluated VLMs.}

%% file: tables/overall_score_per_roomtype.tex
% IDEAL room-type Ov. — preamble: \usepackage{booktabs,graphicx}
\begin{table}[htbp]
  \centering
  \caption{Overall score (Ov.) by model (rows) and room type (columns).}
  \label{tab:roomtype_ov}
  \small
  \setlength{\tabcolsep}{3pt}
  \renewcommand{\arraystretch}{0.95}
  \resizebox{\linewidth}{!}{%
  \begin{tabular}{lcccccccccc}
    \toprule
    \textbf{Model} & \textbf{bathroom} & \textbf{bedroom} & \textbf{kitchen} & \textbf{dining room} & \textbf{living room} & \textbf{meeting room} & \textbf{home studio} & \textbf{office (D5)} & \textbf{classroom (D5)} & \textbf{library (D5)} \\
    \cmidrule(lr){2-8} \cmidrule(lr){9-11}
    \midrule
    Gemini 2.5 Pro & 65.7\% & \textbf{64.5\%} & 62.6\% & \textbf{64.2\%} & \textbf{61.0\%} & \textbf{69.1\%} & \textbf{63.0\%} & 56.6\% & 62.4\% & 70.4\% \\
    GPT-4o & \textbf{65.9\%} & 61.4\% & 60.9\% & 60.3\% & 57.5\% & 64.9\% & 60.0\% & \textbf{57.4\%} & \textbf{76.5\%} & 70.6\% \\
    Claude Sonnet 4.6 & 59.6\% & 61.4\% & 58.2\% & 60.8\% & 60.4\% & 63.9\% & 59.1\% & 57.2\% & 71.3\% & 65.5\% \\
    GPT-5.4 & 65.6\% & 63.5\% & \textbf{64.4\%} & 62.5\% & 60.2\% & 64.1\% & 59.2\% & 52.1\% & 55.4\% & 67.3\% \\
    Gemma-4-31B-IT & 64.8\% & 63.0\% & 59.4\% & 62.9\% & 60.0\% & 67.8\% & 58.8\% & 53.6\% & 58.9\% & 67.8\% \\
    Qwen3-VL-8B & 61.3\% & 59.8\% & 57.5\% & 59.3\% & 56.0\% & 60.5\% & 58.3\% & 46.1\% & 57.5\% & 67.9\% \\
    GLM-4.6V & 60.0\% & 54.7\% & 55.8\% & 55.7\% & 53.6\% & 57.2\% & 51.7\% & 48.2\% & 62.3\% & 68.9\% \\
    Qwen2.5-VL-72B-Instruct & 55.9\% & 54.4\% & 52.5\% & 52.6\% & 53.2\% & 53.0\% & 56.3\% & 49.2\% & 60.8\% & 62.2\% \\
    Qwen3-VL-235B-A22B-Instruct & 61.6\% & 58.0\% & 57.9\% & 56.1\% & 57.1\% & 56.9\% & 57.2\% & 43.7\% & 45.0\% & 60.0\% \\
    GLM-4.6V-Flash & 54.6\% & 54.0\% & 54.9\% & 54.8\% & 51.1\% & 56.2\% & 46.4\% & 54.0\% & 58.2\% & \textbf{70.9\%} \\
    Qwen3-VL-30B & 61.9\% & 58.2\% & 57.7\% & 55.9\% & 51.0\% & 58.7\% & 50.4\% & 45.9\% & 43.5\% & 60.6\% \\
    InternVL3.5-8B & 56.4\% & 48.1\% & 46.8\% & 38.5\% & 41.2\% & 42.2\% & 44.4\% & 40.0\% & 0.0\% & 58.4\% \\
    LLaVA-1.6-Mistral-7B & 56.6\% & 52.6\% & 45.9\% & 49.5\% & 39.1\% & 53.9\% & 39.5\% & 39.9\% & 55.9\% & 50.8\% \\
    \bottomrule
  \end{tabular}%
  }
\end{table}

%% file: tables/camera_viewpoint.tex
% IDEAL camera-corner alignment — preamble: \usepackage{booktabs,graphicx}
\begin{table}[t!]
  \centering
  \small
  \caption{\textbf{Performance breakdown by camera viewpoint.} We report PA, Prec., RA, GCR, GSR, and their macro-average (computed first across metrics per model, then across all models), binned by $d$: the angular difference between the camera's true yaw and the nearest cardinal direction (Appendix~\ref{app_viewpoint}). The italicized row reports the average number of in-view objects per bin; italicized values in the header denote total scenes per bin. \textbf{Bold} marks the lowest value in each metric row.}
  \vspace{.07in}
  \label{tab:corner_score_overall}
  \setlength{\tabcolsep}{6pt}
  \renewcommand{\arraystretch}{1}
  \begin{tabular}{lccccc}
    \toprule
    \textbf{Metric} & \shortstack[c]{\textbf{0\textdegree--9\textdegree}\\\scriptsize \textit{122\,scenes}} & \shortstack[c]{\textbf{9\textdegree--18\textdegree}\\\scriptsize \textit{134\,scenes}} & \shortstack[c]{\textbf{18\textdegree--27\textdegree}\\\scriptsize \textit{195\,scenes}} & \shortstack[c]{\textbf{27\textdegree--36\textdegree}\\\scriptsize \textit{245\,scenes}} & \shortstack[c]{\textbf{36\textdegree--45\textdegree}\\\scriptsize \textit{304\,scenes}} \\
    \midrule
    \textit{Avg.\ obj.\ in view} & 9.55 & 10.67 & 11.1 & 12.82 & 11.21 \\
    \midrule
    PA & 7.22\% & 6.18\% & 7.60\% & \textbf{5.20\%} & 5.22\% \\
    Prec. & 17.86\% & 16.90\% & 16.75\% & 14.12\% & \textbf{13.67\%} \\
    RA & \textbf{21.08\%} & 22.50\% & 25.83\% & 23.42\% & 23.75\% \\
    GCR & 19.77\% & 19.08\% & 17.22\% & 15.85\% & \textbf{15.10\%} \\
    GSR & 84.72\% & 85.06\% & \textbf{77.30\%} & 79.92\% & 81.77\% \\
    Average & 27.02\% & 27.96\% & 27.20\% & \textbf{25.84\%} & 26.06\% \\
    \bottomrule
  \end{tabular}
  \vspace{-.05in}
\label{tab:camera_viewpoint}
\end{table}

%% file: tables/judge_vlm_full.tex
\begin{table*}[t]
\centering
\scriptsize
\caption{Room-type aggregated Judge VLM perceptual score (mean $\pm$ Std.).}
\label{tab:judge_roomtype_appendix}
\begin{tabular}{lcccccc}
\toprule
Room Type & claude-sonnet-4-6 & gemini-2.5-pro & google\_gemma-4-31B-it & gpt-4o & gpt-5.4 & qwen2.5-vl-72b-instruct \\
\midrule
bathroom    & 2.40 $\pm$ 0.83 & \textbf{3.93 $\pm$ 0.59} & 2.93 $\pm$ 0.70 & 3.40 $\pm$ 0.51 & 3.80 $\pm$ 0.68 & 1.60 $\pm$ 0.51 \\
bedroom     & 2.80 $\pm$ 0.77 & 3.67 $\pm$ 0.49 & 2.87 $\pm$ 0.92 & 2.47 $\pm$ 0.52 & \textbf{3.80 $\pm$ 0.41} & 2.73 $\pm$ 0.88 \\
classroom   & 3.40 $\pm$ 0.60 & \textbf{3.90 $\pm$ 0.72} & 3.30 $\pm$ 0.47 & 3.25 $\pm$ 0.72 & 3.60 $\pm$ 0.50 & 1.85 $\pm$ 0.59 \\
diningroom  & 3.27 $\pm$ 0.59 & 3.47 $\pm$ 0.52 & \textbf{3.80 $\pm$ 0.68} & 2.67 $\pm$ 0.62 & 3.73 $\pm$ 0.59 & 1.80 $\pm$ 0.41 \\
homestudio  & 3.40 $\pm$ 0.83 & \textbf{3.47 $\pm$ 0.92} & 2.93 $\pm$ 0.46 & 2.60 $\pm$ 0.51 & 3.13 $\pm$ 0.92 & 2.20 $\pm$ 0.41 \\
kitchen     & 2.40 $\pm$ 0.70 & \textbf{4.10 $\pm$ 0.32} & 2.60 $\pm$ 0.52 & 2.50 $\pm$ 0.53 & 3.00 $\pm$ 0.32 & 1.10 $\pm$ 0.32 \\
library     & 3.08 $\pm$ 0.70 & 2.92 $\pm$ 0.76 & 2.88 $\pm$ 0.83 & 2.60 $\pm$ 0.82 & \textbf{3.32 $\pm$ 0.69} & 2.60 $\pm$ 0.76 \\
livingroom  & 2.53 $\pm$ 0.52 & 3.60 $\pm$ 0.51 & 3.20 $\pm$ 0.68 & 2.27 $\pm$ 0.59 & \textbf{3.67 $\pm$ 0.72} & 1.40 $\pm$ 0.51 \\
meetingroom & 3.35 $\pm$ 0.88 & 3.60 $\pm$ 0.50 & \textbf{4.05 $\pm$ 0.69} & 3.45 $\pm$ 0.51 & 3.25 $\pm$ 0.55 & 1.80 $\pm$ 0.41 \\
office      & 3.45 $\pm$ 0.60 & 3.45 $\pm$ 0.51 & 3.30 $\pm$ 0.57 & 2.40 $\pm$ 0.60 & \textbf{3.60 $\pm$ 0.50} & 1.60 $\pm$ 0.82 \\
\bottomrule
\end{tabular}
\end{table*}

%% file: tables/cross_bench.tex
\begin{table}[t]
\centering
\small
\setlength{\tabcolsep}{6pt}
\caption{Cross-benchmark ranking comparison. For each baseline benchmark, we re-compute model ranks exclusively within the intersecting model set (i.e., models evaluated on both that benchmark and IDEAL-Bench). This ensures that ranking shifts reflect relative performance changes between task paradigms rather than differences in the underlying candidate pool.}
\vspace{.05in}
\label{tab:cross_bench_rank}
\renewcommand{\arraystretch}{0.85}
\begin{tabular}{lcccc|cc}
\toprule
\multirow{2}{*}{Model} & \multicolumn{4}{c|}{QA-based Benchmarks} & \multicolumn{2}{c}{Reconstruction} \\
\cmidrule(lr){2-5} \cmidrule(lr){6-7}
 & Omni & SpatBench & SPAR-Bench & SpaCE-10 & IR3D & IDEAL \\
 & ($n{=}3$) & ($n{=}4$) & ($n{=}3$) & ($n{=}4$) & ($n{=}3$) & \\
\midrule
\multicolumn{7}{l}{\textit{Proprietary Models}} \\
Gemini-2.5-Pro       & \textbf{1} & \textbf{1} & --         & --         & 2          & \textbf{1} \\
GPT-4o               & 3          & --         & \textbf{1} & \textbf{1} & 3          & 2 \\
\midrule
\multicolumn{7}{l}{\textit{Open-source Models}} \\
Qwen2.5-VL-72B       & 2          & 3          & 2          & 4          & \textbf{1} & 3 \\
Qwen3-VL-30B-A3B     & --         & 4          & --         & --         & --         & 5 \\
Qwen3-VL-235B-A22B   & --         & 2          & --         & 2          & --         & 4 \\
LLaVA-1.6-Mistral-7B & --         & --         & 3          & --         & --         & 7 \\
InternVL3.5-8B       & --         & --         & --         & 3          & --         & 6 \\
\bottomrule
\end{tabular}
\vspace{-.1in}
\end{table}

%% file: tables/bootsrap.tex
% Requires: \usepackage{booktabs,multirow}
\begin{table}[!htbp]
\centering
\caption{IDEAL-Bench leaderboard with 95\% bootstrap confidence intervals (2000 resamples, percentile method); each cell reports \textit{point~[CI$_{lo}$, CI$_{hi}$]} in $[0,100]$. Formatting conventions (bold, underline, Ovr., Geo.) follow Table~\ref{tab:ideal_leaderboard}.}
\label{tab:bootstrap}
\resizebox{\textwidth}{!}{%
\small
\setlength{\tabcolsep}{4pt}
\renewcommand{\arraystretch}{1.15}
\begin{tabular}{lcccccccccccccc}
\toprule
\multirow{2}{*}{Model} & \multicolumn{3}{c}{Overall} & \multicolumn{2}{c}{D1: Scene Val.} & \multicolumn{2}{c}{D2: Phys. Plaus.} & \multicolumn{3}{c}{D3: Geo. Acc.} & \multicolumn{2}{c}{D4: Obj. Recog.} & \multicolumn{2}{c}{D5} \\
\cmidrule(lr){2-4} \cmidrule(lr){5-6} \cmidrule(lr){7-8} \cmidrule(lr){9-11} \cmidrule(lr){12-13} \cmidrule(lr){14-15}
 & Ovr.$\uparrow$ & Rank$\downarrow$ & Geo.$\uparrow$ & RTA$\uparrow$ & PSR$\uparrow$ & NOR$\uparrow$ & IBR$\uparrow$ & PA$\uparrow$ & Prec.$\uparrow$ & RA$\uparrow$ & RR$\uparrow$ & NHR$\uparrow$ & GCR$\uparrow$ & GSR$\uparrow$ \\
\midrule
\multicolumn{15}{l}{\textit{Proprietary Models}} \\
Gemini 2.5 Pro & \textbf{62.1 [61.6, 62.7]} & 1 & \textbf{40.1 [38.3, 41.8]} & 87.5 [85.5, 89.4] & \textbf{100.0 [100.0, 100.0]} & \underline{80.9 [79.2, 82.5]} & 34.3 [32.3, 36.4] & \underline{10.8 [9.7, 12.0]} & \textbf{25.5 [23.9, 27.0]} & \textbf{37.7 [35.7, 39.8]} & \textbf{89.9 [89.1, 90.7]} & 90.3 [89.4, 91.2] & \textbf{29.1 [26.7, 31.6]} & 97.3 [95.6, 98.8] \\
GPT-4o & \underline{60.4 [59.9, 60.9]} & 2 & \underline{35.9 [34.6, 37.4]} & 88.5 [86.4, 90.5] & \textbf{100.0 [100.0, 100.0]} & \textbf{84.6 [83.0, 86.2]} & 46.1 [44.0, 48.4] & \textbf{11.5 [10.3, 12.8]} & 23.1 [21.5, 24.6] & 20.0 [18.2, 21.8] & 72.8 [71.4, 74.1] & \underline{92.2 [91.4, 93.1]} & 25.2 [22.9, 27.7] & \textbf{100.0 [100.0, 100.0]} \\
Claude Sonnet 4.6 & 59.2 [58.6, 59.7] & 4 & 34.7 [33.1, 36.3] & 88.5 [86.3, 90.5] & \textbf{100.0 [100.0, 100.0]} & 80.6 [79.1, 82.2] & 34.6 [32.7, 36.4] & 4.9 [4.2, 5.7] & 18.6 [17.3, 19.8] & 26.8 [25.0, 28.6] & \underline{85.4 [84.3, 86.4]} & 88.1 [87.0, 89.1] & 27.8 [25.5, 30.1] & 95.7 [93.6, 97.3] \\
GPT-5.4 & 56.7 [56.0, 57.3] & 5 & 28.4 [25.9, 30.9] & \underline{89.3 [87.3, 91.2]} & \textbf{100.0 [100.0, 100.0]} & 65.0 [63.2, 66.8] & 55.3 [53.3, 57.2] & 8.5 [7.4, 9.7] & \underline{24.3 [22.8, 25.8]} & \underline{33.7 [31.8, 35.7]} & 82.2 [81.2, 83.3] & 89.9 [89.0, 90.8] & \underline{28.6 [26.4, 30.9]} & 46.7 [40.9, 52.5] \\
\midrule
\multicolumn{15}{l}{\textit{Open-source Models}} \\
Gemma-4-31B-IT & 59.7 [59.1, 60.2] & 3 & 35.7 [34.4, 37.1] & 84.4 [82.0, 86.6] & \underline{99.9 [99.7, 100.0]} & 78.3 [76.3, 80.2] & 40.9 [38.8, 43.0] & 9.8 [8.6, 10.9] & 24.1 [22.6, 25.6] & 20.8 [19.0, 22.6] & 81.8 [80.7, 83.0] & \textbf{92.6 [91.7, 93.5]} & 24.1 [22.1, 26.3] & \underline{99.9 [99.7, 100.0]} \\
Qwen3-VL-8B & 55.1 [54.5, 55.7] & 6 & 28.6 [26.6, 30.7] & 84.1 [81.8, 86.3] & \textbf{100.0 [100.0, 100.0]} & 62.2 [60.1, 64.2] & 52.5 [50.3, 54.7] & 8.2 [7.1, 9.3] & 18.5 [17.2, 19.7] & 19.6 [17.9, 21.4] & 73.0 [71.7, 74.2] & 91.1 [90.1, 92.0] & 14.4 [12.8, 16.3] & 82.2 [78.0, 86.7] \\
GLM-4.6V & 54.4 [53.9, 55.0] & 7 & 30.7 [29.3, 32.0] & 87.0 [84.9, 89.0] & \textbf{100.0 [100.0, 100.0]} & 60.9 [58.9, 63.0] & 29.6 [27.6, 31.7] & 4.4 [3.6, 5.2] & 10.1 [9.0, 11.3] & 26.4 [24.4, 28.5] & 76.8 [75.6, 78.0] & 91.2 [90.1, 92.3] & 13.3 [11.3, 15.1] & 99.3 [98.1, 100.0] \\
Qwen2.5-VL-72B-Instruct & 52.6 [52.1, 53.0] & 8 & 27.9 [26.8, 29.0] & 85.4 [83.2, 87.6] & \textbf{100.0 [100.0, 100.0]} & 64.8 [62.6, 66.8] & 26.7 [25.1, 28.4] & 2.1 [1.6, 2.6] & 7.7 [6.9, 8.6] & 18.5 [16.9, 20.2] & 70.6 [69.2, 72.0] & 91.4 [90.4, 92.3] & 11.9 [10.4, 13.6] & 99.3 [98.3, 100.0] \\
Qwen3-VL-235B-A22B-Instruct & 51.8 [51.1, 52.5] & 9 & 24.3 [22.0, 26.7] & \textbf{90.3 [88.1, 92.3]} & 76.0 [73.4, 78.7] & 58.5 [55.9, 61.1] & \underline{56.1 [53.3, 58.8]} & 7.2 [6.2, 8.3] & 21.2 [19.6, 22.9] & 22.8 [20.8, 25.0] & 76.7 [75.4, 78.0] & 90.4 [89.1, 91.6] & 16.1 [14.1, 18.0] & 54.3 [49.4, 59.4] \\
GLM-4.6V-Flash & 51.5 [50.5, 52.5] & 10 & 24.0 [21.1, 27.0] & 86.2 [83.9, 88.4] & 92.5 [90.8, 94.1] & 53.6 [51.4, 55.8] & 49.0 [46.7, 51.4] & 4.7 [3.7, 5.6] & 11.8 [10.5, 13.1] & 19.2 [17.6, 20.9] & 73.7 [72.2, 75.2] & 91.2 [90.2, 92.2] & 7.5 [6.2, 9.0] & 76.9 [67.8, 86.2] \\
Qwen3-VL-30B & 50.8 [50.2, 51.5] & 11 & 22.6 [20.5, 24.7] & 84.0 [81.6, 86.3] & 92.0 [90.4, 93.6] & 56.8 [54.7, 58.8] & 44.2 [42.2, 46.2] & 7.5 [6.6, 8.5] & 14.6 [13.4, 15.9] & 21.6 [19.8, 23.4] & 79.9 [78.7, 81.1] & 89.2 [88.0, 90.2] & 22.9 [20.6, 25.2] & 46.5 [42.1, 50.5] \\
InternVL3.5-8B & 45.9 [45.1, 46.8] & 12 & 27.0 [25.1, 29.2] & 80.8 [77.0, 84.4] & 43.7 [40.6, 46.9] & 66.9 [63.7, 70.1] & 23.0 [20.4, 25.6] & 2.6 [1.7, 3.7] & 4.9 [3.6, 6.3] & 18.4 [15.6, 21.2] & 72.0 [69.8, 74.2] & 84.2 [82.0, 86.2] & 8.9 [4.5, 14.6] & \textbf{100.0 [100.0, 100.0]} \\
LLaVA-1.6-Mistral-7B & 44.5 [43.6, 45.5] & 13 & 17.5 [14.9, 20.3] & 64.8 [61.8, 67.8] & 98.1 [97.2, 98.9] & 16.3 [14.7, 17.9] & \textbf{70.6 [68.4, 72.6]} & 2.4 [2.0, 3.0] & 4.6 [3.8, 5.4] & 19.2 [17.5, 21.0] & 69.2 [67.9, 70.6] & 82.8 [81.4, 84.2] & 8.0 [6.9, 9.1] & 53.3 [44.3, 63.1] \\
InternVL3.5-30B-A3B & \multicolumn{14}{c}{$\times$~\textit{Task Failed}} \\
Janus-Pro-7B & \multicolumn{14}{c}{$\times$~\textit{Task Failed}} \\
\bottomrule
\end{tabular}%
}
\end{table}

%% file: sections/AppE_discussion.tex
% =====================================================================
% Appendix E. Discussion
% =====================================================================

\section{Discussion}
\label{app_e}

% =====================================================================
% E1
% =====================================================================

\subsection{Limitations}
\label{app_e1}

\paragraph{Dataset scope.}
IDEAL-Scenes is restricted to single-room rectangular layouts with axis-aligned furniture; we piloted irregular floor plans but procedural mesh quality on non-rectangular footprints proved difficult to control. All assets are InfiniGen procedural meshes, covering only large layout-determining furniture (beds, desks, sofas) and omitting small objects and clutter, so the benchmark does not test fine-scale recognition or placement. The 1{,}000-scene release (100 per room type) is sufficient for stable per-room-type estimates; the released procedural pipeline supports arbitrary expansion along any of these axes.
\paragraph{Visual capture.}
Each scene is rendered by a single fixed pinhole camera without lens distortion modeling, yielding one image per scene rather than a multi-view bundle. Lighting scales with room dimensions rather than being independently sampled, so the benchmark should not be interpreted as a lighting-robustness test. Since \texttt{scene.blend} is released for every scene, downstream studies can render under alternate intrinsics, lighting conditions, or camera trajectories without regenerating geometry.
\paragraph{Evaluation design.}
The in-view category list and room type are provided in the prompt, isolating geometric reasoning from open-vocabulary detection; future work may remove these cues to test recognition jointly. Yaw is discretized to four cardinal directions, matching the empirical placement distribution but limiting evaluation of off-axis objects. The vertical position $z$
of floor-standing objects is fixed to ground truth during evaluation: a conservative choice that avoids conflating dimension errors with position errors, but one that can be tightened as model accuracy improves.

% =====================================================================
% E2
% =====================================================================

\subsection{Future Work}
\label{app_e2}

\paragraph{Dataset and complexity.}
The most direct extensions follow the axes this benchmark intentionally fixes: real-captured indoor scenes, multi-room and irregular floor plans, explicit lighting variability, and small objects or clutter with unrestricted rotation. A complementary direction is to vary difficulty along a single axis at a time within fixed room categories: sweeping lighting, occlusion, or grid density in graded steps to test whether model degradation is gradual or exhibits sharp failure thresholds.

\paragraph{Model analysis.}
Our finding that geometric regression is a universal bottleneck is a population-level statement and does not isolate the contribution of scale, training recipe, or visual-encoder architecture. A clean follow-up would fix prompt and parsing strategy and evaluate within a single model family across scales. Separately, our protocol suppresses chain-of-thought reasoning; since metric spatial judgment is plausibly a perceptual rather than deliberative skill, the gap we expose may not be the kind that longer reasoning chains can close, but verifying this by comparing the same model with and without extended reasoning is a natural next step.

%% file: sections/AppF_broader_impacts.tex
\section{Impact Statement}
\label{app:impact}
This paper introduces IDEAL-Bench and IDEAL-Scenes, a benchmark and a dataset to diagnose 3D spatial understanding in vision-language models. Potential positive impacts include advancing spatially-aware AI for robotics, embodied AI, AR/VR, and assistive technologies, and lowering barriers to reproducible spatial reasoning research. All scenes are generated procedurally via InfiniGen with no real-person imagery or scraped data. As this work presents an evaluation protocol rather than a deployable system, we identify no direct negative societal impacts.

%% file: main.bib
@String(CVPR= {IEEE Conf. Comput. Vis. Pattern Recog.})

@String(ICCV= {Int. Conf. Comput. Vis.})

@String(ECCV= {Eur. Conf. Comput. Vis.})

@String(ICLR = {Int. Conf. Learn. Represent.})

@misc{gemini,
      title={Gemini: A Family of Highly Capable Multimodal Models}, 
      author={Gemini Team},
      year={2025},
      eprint={2312.11805},
      archivePrefix={arXiv},
      primaryClass={cs.CL},
      url={https://arxiv.org/abs/2312.11805}, 
}

@misc{gemini2.5pro,
      title={Gemini 2.5: Pushing the Frontier with Advanced Reasoning, Multimodality, Long Context, and Next Generation Agentic Capabilities}, 
      author={Gheorghe Comanici and Eric Bieber and Mike Schaekermann and others},
      year={2025},
      eprint={2507.06261},
      archivePrefix={arXiv},
      primaryClass={cs.CL},
      url={https://arxiv.org/abs/2507.06261}, 
}

@article{gpt4,
  title={Gpt-4 technical report},
  author={Achiam, Josh and Adler, Steven and Agarwal, Sandhini and Ahmad, Lama and Akkaya, Ilge and Aleman, Florencia Leoni and Almeida, Diogo and Altenschmidt, Janko and Altman, Sam and Anadkat, Shyamal and others},
  journal={arXiv preprint arXiv:2303.08774},
  year={2023}
}

@article{gpt4o,
  title={Gpt-4o system card},
  author={Hurst, Aaron and Lerer, Adam and Goucher, Adam P and Perelman, Adam and Ramesh, Aditya and Clark, Aidan and Ostrow, AJ and Welihinda, Akila and Hayes, Alan and Radford, Alec and others},
  journal={arXiv preprint arXiv:2410.21276},
  year={2024}
}

@misc{gpt5,
      title={OpenAI GPT-5 System Card}, 
      author={Aaditya Singh and Adam Fry and Adam Perelman and others},
      year={2026},
      eprint={2601.03267},
      archivePrefix={arXiv},
      primaryClass={cs.CL},
      url={https://arxiv.org/abs/2601.03267}, 
}

@misc{sonnet4.6,
  title={{Claude 4.6 Sonnet System Card}},
  author={{Anthropic}},
  year={2026},
  note={Accessed: 2026-06-29}
}

@article{gemma,
  title={Gemma: Open models based on gemini research and technology},
  author={Team, Gemma and Mesnard, Thomas and Hardin, Cassidy and Dadashi, Robert and Bhupatiraju, Surya and Pathak, Shreya and Sifre, Laurent and Rivi{\`e}re, Morgane and Kale, Mihir Sanjay and Love, Juliette and others},
  journal={arXiv preprint arXiv:2403.08295},
  year={2024}
}

@article{qwen3,
  title={Qwen3-vl technical report},
  author={Qwen Team},
  journal={arXiv preprint arXiv:2511.21631},
  year={2025}
}

@misc{qwen2.5,
      title={Qwen2.5 Technical Report}, 
      author={Qwen Team},
      year={2025},
      eprint={2412.15115},
      archivePrefix={arXiv},
      primaryClass={cs.CL},
      url={https://arxiv.org/abs/2412.15115}, 
}

@misc{glm4.6,
      title={GLM-4.5V and GLM-4.1V-Thinking: Towards Versatile Multimodal Reasoning with Scalable Reinforcement Learning}, 
      author={V Team},
      year={2026},
      eprint={2507.01006},
      archivePrefix={arXiv},
      primaryClass={cs.CV},
      url={https://arxiv.org/abs/2507.01006}, 
}

@inproceedings{internvl,
  title={Internvl: Scaling up vision foundation models and aligning for generic visual-linguistic tasks},
  author={Chen, Zhe and Wu, Jiannan and Wang, Wenhai and Su, Weijie and Chen, Guo and Xing, Sen and Zhong, Muyan and Zhang, Qinglong and Zhu, Xizhou and Lu, Lewei and others},
  booktitle={Proceedings of the IEEE/CVF conference on computer vision and pattern recognition},
  pages={24185--24198},
  year={2024}
}

@inproceedings{llava,
    author      = {Liu, Haotian and Li, Chunyuan and Wu, Qingyang and Lee, Yong Jae},
    title       = {Visual Instruction Tuning},
    booktitle   = {NeurIPS},
    year        = {2023}
}

@misc{januspro,
      title={Janus-Pro: Unified Multimodal Understanding and Generation with Data and Model Scaling}, 
      author={Xiaokang Chen and Zhiyu Wu and Xingchao Liu and Zizheng Pan and Wen Liu and Zhenda Xie and Xingkai Yu and Chong Ruan},
      year={2025},
      eprint={2501.17811},
      archivePrefix={arXiv},
      primaryClass={cs.AI},
      url={https://arxiv.org/abs/2501.17811}, 
}

@inproceedings{zhou2024layout,
  title={Layout-your-3d: Controllable and precise 3d generation with 2d blueprint},
  author={Zhou, Junwei and Li, Xueting and Qi, Lu and Yang, Ming-Hsuan},
  booktitle={International Conference on Learning Representations (ICLR)},
  year={2025}
}

@misc{zhou2025coco,
      title={CoCo4D: Comprehensive and Complex 4D Scene Generation}, 
      author={Junwei Zhou and Xueting Li and Lu Qi and Ming-Hsuan Yang},
      year={2025},
      eprint={2506.19798},
      archivePrefix={arXiv},
      primaryClass={cs.CV},
      url={https://arxiv.org/abs/2506.19798}, 
}

@misc{zhou2026gena,
      title={GENA3D: Generative Amodal 3D Modeling by Bridging 2D Priors and 3D Coherence}, 
      author={Junwei Zhou and Yu-Wing Tai},
      year={2026},
      eprint={2511.21945},
      archivePrefix={arXiv},
      primaryClass={cs.CV},
      url={https://arxiv.org/abs/2511.21945}, 
}

@article{wang2025tabletopgen,
  title={TabletopGen: Instance-Level Interactive 3D Tabletop Scene Generation from Text or Single Image},
  author={Wang, Ziqian and He, Yonghao and Yang, Licheng and Zou, Wei and Ma, Hongxuan and Liu, Liu and Sui, Wei and Guo, Yuxin and Su, Hu},
  journal={arXiv preprint arXiv:2512.01204},
  year={2025}
}

@misc{ling2025scenethesislanguagevisionagentic,
      title={Scenethesis: A Language and Vision Agentic Framework for 3D Scene Generation}, 
      author={Lu Ling and Chen-Hsuan Lin and Tsung-Yi Lin and Yifan Ding and Yu Zeng and Yichen Sheng and Yunhao Ge and Ming-Yu Liu and Aniket Bera and Zhaoshuo Li},
      year={2025},
      eprint={2505.02836},
      archivePrefix={arXiv},
      primaryClass={cs.CV},
      url={https://arxiv.org/abs/2505.02836}, 
}

@InProceedings{Gu_2025_CVPR,
    author    = {Gu, Zeqi and Cui, Yin and Li, Zhaoshuo and Wei, Fangyin and Ge, Yunhao and Gu, Jinwei and Liu, Ming-Yu and Davis, Abe and Ding, Yifan},
    title     = {ArtiScene: Language-Driven Artistic 3D Scene Generation Through Image Intermediary},
    booktitle = {Proceedings of the IEEE/CVF Conference on Computer Vision and Pattern Recognition (CVPR)},
    month     = {June},
    year      = {2025},
    pages     = {2891-2901}
}

@article{feng2023layoutgpt,
  title={Layoutgpt: Compositional visual planning and generation with large language models},
  author={Feng, Weixi and Zhu, Wanrong and Fu, Tsu-jui and Jampani, Varun and Akula, Arjun and He, Xuehai and Basu, Sugato and Wang, Xin Eric and Wang, William Yang},
  journal={Advances in Neural Information Processing Systems},
  volume={36},
  pages={18225--18250},
  year={2023}
}

@InProceedings{Yang_2024_CVPR,
    author    = {Yang, Yue and Sun, Fan-Yun and Weihs, Luca and VanderBilt, Eli and Herrasti, Alvaro and Han, Winson and Wu, Jiajun and Haber, Nick and Krishna, Ranjay and Liu, Lingjie and Callison-Burch, Chris and Yatskar, Mark and Kembhavi, Aniruddha and Clark, Christopher},
    title     = {Holodeck: Language Guided Generation of 3D Embodied AI Environments},
    booktitle = {Proceedings of the IEEE/CVF Conference on Computer Vision and Pattern Recognition (CVPR)},
    month     = {June},
    year      = {2024},
    pages     = {16227-16237}
}

@misc{zhou2026perceivethenplan,
    title={Perceive-then-Plan: Layout-as-Policy for Monocular 3D Scene Layout Estimation}, 
    author={Junwei Zhou and Yu-Wing Tai},
    year={2026},
    eprint={2605.25326},
    archivePrefix={arXiv},
    primaryClass={cs.CV},
    url={https://arxiv.org/abs/2605.25326}, 
}

@article{liu2025worldcraft,
  title={Worldcraft: Photo-realistic 3d world creation and customization via llm agents},
  author={Liu, Xinhang and Tang, Chi-Keung and Tai, Yu-Wing},
  journal={arXiv preprint arXiv:2502.15601},
  year={2025}
}

@article{li2024advances,
  title={Advances in 3d generation: A survey},
  author={Li, Xiaoyu and Zhang, Qi and Kang, Di and Cheng, Weihao and Gao, Yiming and Zhang, Jingbo and Liang, Zhihao and Liao, Jing and Cao, Yan-Pei and Shan, Ying},
  journal={arXiv preprint arXiv:2401.17807},
  year={2024}
}

@article{wen20253d,
  title={3d scene generation: A survey},
  author={Wen, Beichen and Xie, Haozhe and Chen, Zhaoxi and Hong, Fangzhou and Liu, Ziwei},
  journal={arXiv preprint arXiv:2505.05474},
  year={2025}
}

@misc{liu2024comprehensivesurvey3dcontent,
      title={A Comprehensive Survey on 3D Content Generation}, 
      author={Jian Liu and Xiaoshui Huang and Tianyu Huang and Lu Chen and Yuenan Hou and Shixiang Tang and Ziwei Liu and Wanli Ouyang and Wangmeng Zuo and Junjun Jiang and Xianming Liu},
      year={2024},
      eprint={2402.01166},
      archivePrefix={arXiv},
      primaryClass={cs.CV},
      url={https://arxiv.org/abs/2402.01166}, 
}

@article{chen2025sam,
  title={Sam 3d: 3dfy anything in images},
  author={Chen, Xingyu and Chu, Fu-Jen and Gleize, Pierre and Liang, Kevin J and Sax, Alexander and Tang, Hao and Wang, Weiyao and Guo, Michelle and Hardin, Thibaut and Li, Xiang and others},
  journal={arXiv preprint arXiv:2511.16624},
  year={2025}
}

@inproceedings{xie2024physgaussian,
  title={Physgaussian: Physics-integrated 3d gaussians for generative dynamics},
  author={Xie, Tianyi and Zong, Zeshun and Qiu, Yuxing and Li, Xuan and Feng, Yutao and Yang, Yin and Jiang, Chenfanfu},
  booktitle={Proceedings of the IEEE/CVF Conference on Computer Vision and Pattern Recognition},
  pages={4389--4398},
  year={2024}
}

@article{lin2025pat3d,
  title={PAT3D: Physics-Augmented Text-to-3D Scene Generation},
  author={Lin, Guying and Huang, Kemeng and Liu, Michael and Gao, Ruihan and Chen, Hanke and Chen, Lyuhao and Lu, Beijia and Komura, Taku and Liu, Yuan and Zhu, Jun-Yan and others},
  journal={arXiv preprint arXiv:2511.21978},
  year={2025}
}

@inproceedings{omnispatial,
  title={OmniSpatial: Towards Comprehensive Spatial Reasoning Benchmark for Vision Language Models},
  author={Jia, Mengdi and Qi, Zekun and Zhang, Shaochen and Zhang, Wenyao and Yu, Xinqiang and He, Jiawei and Wang, He and Yi, Li},
  booktitle={International Conference on Learning Representations (ICLR)},
  year={2026}
}

@misc{vsibench,
      title={Thinking in Space: How Multimodal Large Language Models See, Remember, and Recall Spaces}, 
      author={Jihan Yang and Shusheng Yang and Anjali W. Gupta and Rilyn Han and Li Fei-Fei and Saining Xie},
      year={2025},
      eprint={2412.14171},
      archivePrefix={arXiv},
      primaryClass={cs.CV},
      url={https://arxiv.org/abs/2412.14171}, 
}

@misc{sparbench,
      title={From Flatland to Space: Teaching Vision-Language Models to Perceive and Reason in 3D}, 
      author={Jiahui Zhang and Yurui Chen and Yanpeng Zhou and Yueming Xu and Ze Huang and Jilin Mei and Junhui Chen and Yu-Jie Yuan and Xinyue Cai and Guowei Huang and Xingyue Quan and Hang Xu and Li Zhang},
      year={2026},
      eprint={2503.22976},
      archivePrefix={arXiv},
      primaryClass={cs.CV},
      url={https://arxiv.org/abs/2503.22976}, 
}

@misc{spatialbench,
      title={SpatialBench: Benchmarking Multimodal Large Language Models for Spatial Cognition}, 
      author={Peiran Xu and Sudong Wang and Yao Zhu and Jianing Li and Gege Qi and Yunjian Zhang},
      year={2026},
      eprint={2511.21471},
      archivePrefix={arXiv},
      primaryClass={cs.AI},
      url={https://arxiv.org/abs/2511.21471}, 
}

@misc{spatialrgpt,
      title={SpatialRGPT: Grounded Spatial Reasoning in Vision Language Models}, 
      author={An-Chieh Cheng and Hongxu Yin and Yang Fu and Qiushan Guo and Ruihan Yang and Jan Kautz and Xiaolong Wang and Sifei Liu},
      year={2024},
      eprint={2406.01584},
      archivePrefix={arXiv},
      primaryClass={cs.CV},
      url={https://arxiv.org/abs/2406.01584}, 
}

@misc{spatialvlm,
      title={SpatialVLM: Endowing Vision-Language Models with Spatial Reasoning Capabilities}, 
      author={Boyuan Chen and Zhuo Xu and Sean Kirmani and Brian Ichter and Danny Driess and Pete Florence and Dorsa Sadigh and Leonidas Guibas and Fei Xia},
      year={2024},
      eprint={2401.12168},
      archivePrefix={arXiv},
      primaryClass={cs.CV},
      url={https://arxiv.org/abs/2401.12168}, 
}

@article{windecker2025navitrace,
  title={Navitrace: Evaluating embodied navigation of vision-language models},
  author={Windecker, Tim and Patel, Manthan and Reuss, Moritz and Schwarzkopf, Richard and Cadena, Cesar and Lioutikov, Rudolf and Hutter, Marco and Frey, Jonas},
  journal={arXiv preprint arXiv:2510.26909},
  year={2025}
}

@article{ir3dbench,
  title={Ir3d-bench: Evaluating vision-language model scene understanding as agentic inverse rendering},
  author={Liu, Parker and Li, Chenxin and Li, Zhengxin and Wu, Yipeng and Li, Wuyang and Yang, Zhiqin and Zhang, Zhenyuan and Lin, Yunlong and Han, Sirui and Feng, Brandon Y},
  journal={arXiv preprint arXiv:2506.23329},
  year={2025}
}

@inproceedings{scanrefer,
  title={ScanRefer: 3D Object Localization in RGB-D Scans Using Natural Language},
  author={Chen, Dave Zhenyu and Chang, Angel X. and Nie{\ss}ner, Matthias},
  booktitle={European Conference on Computer Vision (ECCV)},
  year={2020}
}

@inproceedings{referit3d,
  title={ReferIt3D: Neural Listeners for Fine-Grained 3D Object Identification 
         in Real-World Scenes},
  author={Achlioptas, Panos and Abdelreheem, Ahmed and Xia, Fei and 
          Elhoseiny, Mohamed and Guibas, Leonidas J.},
  booktitle={European Conference on Computer Vision (ECCV)},
  year={2020}
}

@misc{space10,
      title={SpaCE-10: A Comprehensive Benchmark for Multimodal Large Language Models in Compositional Spatial Intelligence}, 
      author={Ziyang Gong and Wenhao Li and Oliver Ma and Songyuan Li and Zhaokai Wang and Songyuan Li and Jiayi Ji and Xue Yang and Gen Luo and Junchi Yan and Rongrong Ji},
      year={2025},
      eprint={2506.07966},
      archivePrefix={arXiv},
      primaryClass={cs.CV},
      url={https://arxiv.org/abs/2506.07966}, 
}

@inproceedings{space3d,
  title={{Space3D-Bench: Spatial 3D Question Answering Benchmark}},
  author={Szymanska, Emilia and Dusmanu, Mihai and Buurlage, Jan-Willem and Rad, Mahdi and Pollefeys, Marc},
  booktitle={European Conference on Computer Vision (ECCV) Workshops},
  year={2024}
}

@misc{infinigen,
      title={Infinigen Indoors: Photorealistic Indoor Scenes using Procedural Generation}, 
      author={Alexander Raistrick and Lingjie Mei and Karhan Kayan and David Yan and Yiming Zuo and Beining Han and Hongyu Wen and Meenal Parakh and Stamatis Alexandropoulos and Lahav Lipson and Zeyu Ma and Jia Deng},
      year={2024},
      eprint={2406.11824},
      archivePrefix={arXiv},
      primaryClass={cs.CV},
      url={https://arxiv.org/abs/2406.11824}, 
}

@misc{arkit,
      title={ARKitScenes: A Diverse Real-World Dataset For 3D Indoor Scene Understanding Using Mobile RGB-D Data}, 
      author={Gilad Baruch and Zhuoyuan Chen and Afshin Dehghan and Tal Dimry and Yuri Feigin and Peter Fu and Thomas Gebauer and Brandon Joffe and Daniel Kurz and Arik Schwartz and Elad Shulman},
      year={2022},
      eprint={2111.08897},
      archivePrefix={arXiv},
      primaryClass={cs.CV},
      url={https://arxiv.org/abs/2111.08897}, 
}

@misc{scannet,
      title={ScanNet: Richly-annotated 3D Reconstructions of Indoor Scenes}, 
      author={Angela Dai and Angel X. Chang and Manolis Savva and Maciej Halber and Thomas Funkhouser and Matthias Nießner},
      year={2017},
      eprint={1702.04405},
      archivePrefix={arXiv},
      primaryClass={cs.CV},
      url={https://arxiv.org/abs/1702.04405}, 
}

@misc{hypersim,
      title={Hypersim: A Photorealistic Synthetic Dataset for Holistic Indoor Scene Understanding}, 
      author={Mike Roberts and Jason Ramapuram and Anurag Ranjan and Atulit Kumar and Miguel Angel Bautista and Nathan Paczan and Russ Webb and Joshua M. Susskind},
      year={2021},
      eprint={2011.02523},
      archivePrefix={arXiv},
      primaryClass={cs.CV},
      url={https://arxiv.org/abs/2011.02523}, 
}

@misc{ase,
      title={SceneScript: Reconstructing Scenes With An Autoregressive Structured Language Model}, 
      author={Armen Avetisyan and Christopher Xie and Henry Howard-Jenkins and Tsun-Yi Yang and Samir Aroudj and Suvam Patra and Fuyang Zhang and Duncan Frost and Luke Holland and Campbell Orme and Jakob Engel and Edward Miller and Richard Newcombe and Vasileios Balntas},
      year={2024},
      eprint={2403.13064},
      archivePrefix={arXiv},
      primaryClass={cs.CV},
      url={https://arxiv.org/abs/2403.13064}, 
}

@misc{3dfront,
      title={3D-FRONT: 3D Furnished Rooms with layOuts and semaNTics}, 
      author={Huan Fu and Bowen Cai and Lin Gao and Lingxiao Zhang and Jiaming Wang Cao Li and Zengqi Xun and Chengyue Sun and Rongfei Jia and Binqiang Zhao and Hao Zhang},
      year={2021},
      eprint={2011.09127},
      archivePrefix={arXiv},
      primaryClass={cs.CV},
      url={https://arxiv.org/abs/2011.09127}, 
}

@misc{matterport3d,
      title={Matterport3D: Learning from RGB-D Data in Indoor Environments}, 
      author={Angel Chang and Angela Dai and Thomas Funkhouser and Maciej Halber and Matthias Nießner and Manolis Savva and Shuran Song and Andy Zeng and Yinda Zhang},
      year={2017},
      eprint={1709.06158},
      archivePrefix={arXiv},
      primaryClass={cs.CV},
      url={https://arxiv.org/abs/1709.06158}, 
}

@misc{scenenetrgbd,
      title={SceneNet RGB-D: 5M Photorealistic Images of Synthetic Indoor Trajectories with Ground Truth}, 
      author={John McCormac and Ankur Handa and Stefan Leutenegger and Andrew J. Davison},
      year={2017},
      eprint={1612.05079},
      archivePrefix={arXiv},
      primaryClass={cs.CV},
      url={https://arxiv.org/abs/1612.05079}, 
}

@misc{interiornet,
      title={InteriorNet: Mega-scale Multi-sensor Photo-realistic Indoor Scenes Dataset}, 
      author={Wenbin Li and Sajad Saeedi and John McCormac and Ronald Clark and Dimos Tzoumanikas and Qing Ye and Yuzhong Huang and Rui Tang and Stefan Leutenegger},
      year={2018},
      eprint={1809.00716},
      archivePrefix={arXiv},
      primaryClass={cs.CV},
      url={https://arxiv.org/abs/1809.00716}, 
}

@misc{structure3d,
      title={Structured3D: A Large Photo-realistic Dataset for Structured 3D Modeling}, 
      author={Jia Zheng and Junfei Zhang and Jing Li and Rui Tang and Shenghua Gao and Zihan Zhou},
      year={2020},
      eprint={1908.00222},
      archivePrefix={arXiv},
      primaryClass={cs.CV},
      url={https://arxiv.org/abs/1908.00222}, 
}

@misc{m3dlayout,
      title={M3DLayout: A Multi-Source Dataset of 3D Indoor Layouts and Structured Descriptions for 3D Generation}, 
      author={Yiheng Zhang and Zhuojiang Cai and Mingdao Wang and Meitong Guo and Tianxiao Li and Li Lin and Yuwang Wang},
      year={2026},
      eprint={2509.23728},
      archivePrefix={arXiv},
      primaryClass={cs.CV},
      url={https://arxiv.org/abs/2509.23728}, 
}

@inproceedings{sun3d,
  title     = {{SUN3D}: A Database of Big Spaces Reconstructed Using {SfM} and Object Labels},
  author    = {Xiao, Jianxiong and Owens, Andrew and Torralba, Antonio},
  booktitle = {Proceedings of the IEEE International Conference on Computer Vision (ICCV)},
  pages     = {1625--1632},
  year      = {2013},
  doi       = {10.1109/ICCV.2013.458}
}

@inproceedings{sunrgbd,
  title     = {{SUN RGB-D}: A {RGB-D} Scene Understanding Benchmark Suite},
  author    = {Song, Shuran and Lichtenberg, Samuel P. and Xiao, Jianxiong},
  booktitle = {Proceedings of the IEEE Conference on Computer Vision and Pattern Recognition (CVPR)},
  pages     = {567--576},
  year      = {2015},
  doi       = {10.1109/CVPR.2015.7298655}
}

@inproceedings{scenenn,
  title     = {{SceneNN}: A Scene Meshes Dataset with {aNNotations}},
  author    = {Hua, Binh-Son and Pham, Quang-Hieu and Nguyen, Duc Thanh and Tran, Minh-Khoi and Yu, Lap-Fai and Yeung, Sai-Kit},
  booktitle = {Proceedings of the International Conference on 3D Vision (3DV)},
  pages     = {92--101},
  year      = {2016},
  doi       = {10.1109/3DV.2016.18}
}

@misc{spatialreasoner,
      title={SpatialReasoner: Towards Explicit and Generalizable 3D Spatial Reasoning}, 
      author={Wufei Ma and Yu-Cheng Chou and Qihao Liu and Xingrui Wang and Celso de Melo and Jianwen Xie and Alan Yuille},
      year={2025},
      eprint={2504.20024},
      archivePrefix={arXiv},
      primaryClass={cs.CV},
      url={https://arxiv.org/abs/2504.20024}, 
}

@misc{ma,
      title={Do 3D Large Language Models Really Understand 3D Spatial Relationships?}, 
      author={Xianzheng Ma and Tao Sun and Shuai Chen and Yash Bhalgat and Jindong Gu and Angel X Chang and Iro Armeni and Iro Laina and Songyou Peng and Victor Adrian Prisacariu},
      year={2026},
      eprint={2603.23523},
      archivePrefix={arXiv},
      primaryClass={cs.CL},
      url={https://arxiv.org/abs/2603.23523}, 
}

@misc{zhang,
      title={Do Vision-Language Models Represent Space and How? Evaluating Spatial Frame of Reference Under Ambiguities}, 
      author={Zheyuan Zhang and Fengyuan Hu and Jayjun Lee and Freda Shi and Parisa Kordjamshidi and Joyce Chai and Ziqiao Ma},
      year={2025},
      eprint={2410.17385},
      archivePrefix={arXiv},
      primaryClass={cs.CL},
      url={https://arxiv.org/abs/2410.17385}, 
}

@misc{zha,
      title={How to Enable LLM with 3D Capacity? A Survey of Spatial Reasoning in LLM}, 
      author={Jirong Zha and Yuxuan Fan and Xiao Yang and Chen Gao and Xinlei Chen},
      year={2025},
      eprint={2504.05786},
      archivePrefix={arXiv},
      primaryClass={cs.CV},
      url={https://arxiv.org/abs/2504.05786}, 
}

@misc{wang,
      title={What Makes for Good Visual Tokenizers for Large Language Models?}, 
      author={Guangzhi Wang and Yixiao Ge and Xiaohan Ding and Mohan Kankanhalli and Ying Shan},
      year={2023},
      eprint={2305.12223},
      archivePrefix={arXiv},
      primaryClass={cs.CV},
      url={https://arxiv.org/abs/2305.12223}, 
}

@article{liu,
  title={{Spatial Intelligence in Vision-Language Models: A Comprehensive Survey}},
  author={Liu, Disheng and Liang, Tuo and Hu, Zhe and Peng, Jierui and Lu, Yiren and Xu, Yi and Fu, Yun and Yin, Yu},
  year={2025},
  journal={TechRxiv preprint},
  doi={10.36227/techrxiv.176231405.57942913},
}
